\documentclass{article}

\usepackage[preprint, nonatbib]{neurips_2025}

\usepackage[%
  natbib=true,
  url=true,
  eprint=false,
  hyperref=true,
  style=numeric-comp,
  maxnames=1,
  minnames=1,
  maxbibnames=5,
  minbibnames=4,
  giveninits=true,
  uniquename=init,]{biblatex}
\bibliography{bibliography}
\AtEveryBibitem{\clearname{editor}} 

\usepackage[utf8]{inputenc} 
\usepackage[T1]{fontenc}    
\usepackage{hyperref}       
\usepackage{url}            
\usepackage{booktabs}       
\usepackage{amsfonts}       
\usepackage{amsmath}        
\usepackage{nicefrac}       
\usepackage{microtype}      
\usepackage{xcolor}         

\usepackage{graphicx}
\usepackage{svg}
\usepackage{subcaption} 
\usepackage{svg}
\usepackage{wrapfig}
\usepackage{siunitx}
\usepackage{enumitem} 
\usepackage[capitalize,noabbrev]{cleveref}

\DeclareMathOperator*{\argmin}{\arg\!\min}

\title{P3D: Scalable Neural Surrogates for High-Resolution 3D Physics Simulations with Global Context}

\newcommand{\affiliation}[2]{\normalsize\rm\textsuperscript{#1}#2}

\author{
  Benjamin Holzschuh\thanks{Correspondence to \texttt{benjamin.holzschuh@tum.de}} 
  \And
  Georg Kohl 
  \And  
  Florian Redinger 
  \And Nils Thuerey \And \\
  \\ [-24pt]
  \affiliation{}{\small{Technical University of Munich}} \\
}

\begin{document}

\maketitle

\begin{abstract}
  We present a scalable framework for learning deterministic and probabilistic neural surrogates for high-resolution 3D physics simulations. We introduce a hybrid CNN-Transformer backbone architecture targeted for 3D physics simulations, which significantly outperforms existing architectures in terms of speed and accuracy. Our proposed network can be pretrained on small patches of the simulation domain, which can be fused to obtain a global solution, optionally guided via a fast and scalable sequence-to-sequence model to include long-range dependencies.  
  This setup allows for training large-scale models with reduced memory and compute requirements for high-resolution datasets. We evaluate our backbone architecture against a large set of baseline methods with the objective to simultaneously learn the dynamics of 14 different types of PDEs in 3D. We demonstrate how to scale our model to high-resolution isotropic turbulence with spatial resolutions of up to $512^3$. Finally, we demonstrate the versatility of our network by training it as a diffusion model to produce probabilistic samples of highly turbulent 3D channel flows across varying Reynolds numbers, accurately capturing the underlying flow statistics.
\end{abstract}

\section{Introduction}

Training neural networks on high-resolution data substantially increases the required GPU memory and compute costs. 
Scaling models and their input dimensions typically requires substantial engineering effort, posing a major barrier to the widespread and cost-effective adoption of machine learning across application domains.
Scientific machine learning and engineering are especially affected due to the multi-scale nature of relevant phenomena whose modeling often requires specialized and highly computationally demanding numerical solutions. In this paper, we focus on learning surrogate models for simulations focusing on fluid dynamics that have downstream applications in fields such as aerospace~\cite{arranz2024building},
climate science~\cite{bodnar2024aurora}, energy systems~\cite{degrave2022magnetic}, and biomedical engineering~\cite{morris2016computational}. 
Machine learning models inherently compete with existing solvers, which are often employed to create the reference targets for learning. As such, they need to either significantly outperform the corresponding solvers while maintaining an acceptable level of accuracy~\cite{kochov_fluids_2021,pestourie2023physics}, or yield solutions where traditional solvers fall short, for example working with noisy~\cite{franz2023nglobt} or only partial input data~\cite{shu2023physics}, or by providing uncertainty estimates~\cite{jacobsen2023cocogen}. 
A large fraction of papers in this area address learning problems for either low-dimensional or comparatively 
smooth data in 2D. In this paper, we deliberately focus on high-resolution 3D phenomena, covering a wide variety of different types of PDEs.  

Specifically, we introduce a scalable framework for surrogate modeling in 3D. We first present \textit{P3D}, a hybrid CNN-Transformer backbone, and demonstrate how an efficient combination of convolutions for local features with fast processing of high spatial resolution with multiple channels and windowed attention mechanisms for high-dimensional token representations at higher levels significantly improves the scalability and accuracy compared to state-of-the-art baselines. We evaluate the P3D backbone architecture through an extensive comparison with existing architectures for modeling physics simulations in 3D, training on 14 different types of dynamics simultaneously. We then train P3D as a surrogate model for isotropic turbulence in 3D at resolution $512^3$ and demonstrate how our network trained on crops of $128^3$ can be scaled to the entire domain, while achieving high accuracy with temporally stable autoregressive rollouts.

A significant difficulty when modeling large systems is how to aggregate and distribute information globally across the network. 

We propose to link the bottleneck layers of P3D with a sequence-to-sequence model, called \emph{context model}, for an efficient global processing and information aggregation similar to highly optimized self-attention mechanisms in LLM layers like HyperAttention~\cite{han2023hyperattention}  or linear alternatives like state space models \cite{gu2023mamba}. Additionally, we propose a direct mechanism to let global information flow back to higher processing levels by injecting information into adaptive instance normalization layers.  

\begin{figure}[t]
    \centering
    \includegraphics[width=\linewidth]{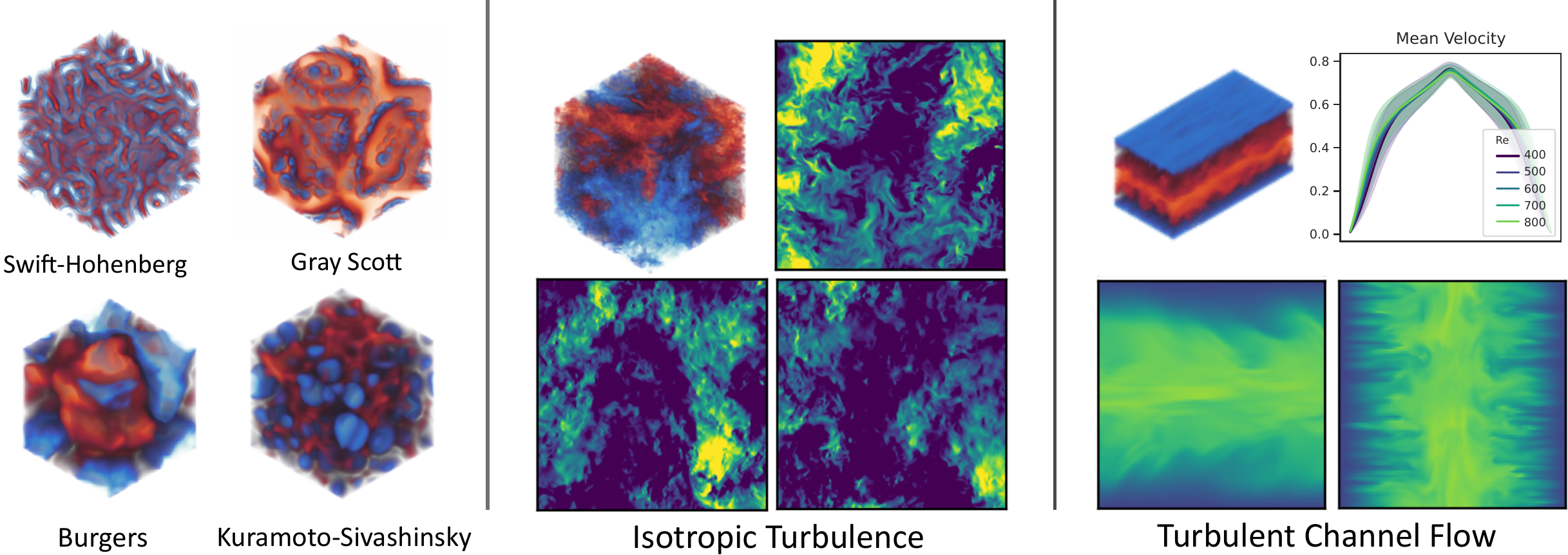}
    \caption{Experiments: we train P3D on 14 different PDE dynamics simultaneously and verify its high efficiency and performance in a large benchmark comparison (left). We scale P3D to a simulation of forced isotorpic turbulence at resolution $512^3$ training only on crops of the simulation domain at size $128^3$ (middle). We train P3D as a diffusion model for a turbulent channel flow, assembling the simulation domain from smaller crops that are linked via a global context model (right). }
    \label{fig:overview_experiments}
\end{figure}

In our final experiment, we train P3D as a diffusion model to learn the distribution of velocity and pressure fields of a turbulent channel flow on a non-equidistant grid. This setup requires access to global information like the relative position to the walls, and cannot be addressed by learning local representations alone. We verify that velocity profiles from the generated samples closely match the ground truth. High distributional accuracy is obtained even when the solution fields are constructed with smaller blocks, which are only coordinated through the information flow from the context model. See \cref{fig:overview_experiments} for an overview of the experiments. To summarize our contributions:

\vspace{-3pt}
\begin{itemize}[itemsep=0pt]
    \item We introduce \textit{P3D}, a hybrid CNN-Transformer architecture for autoregressive prediction of high-resolution physics simulations in 3D, combining CNNs for fast learning of local features and windowed self-attention for deep representation learning. 
    \item We demonstrate the efficiency and versatility of P3D in three experiments: (1) learning multiple types of simulations at the same time, (2) scaling P3D trained on crops of $128^3$ to a high-resolution simulation of isotropic turbulence at $512^3$, and, (3) generating probabilistic samples from P3D trained 
    via flow matching for the velocity and pressure fields of a turbulent channel flow, closely matching the ground truth flow statistics. 
    \item We propose mechanisms for efficient global information processing, including linking bottleneck layers with a sequence-to-sequence context model and injecting global information into adaptive instance normalization layers via region tokens.
    \item We evaluate different setups for finetuning, which enable a more fine-grained control of precomputation and gradient backpropagation through encoder and decoder blocks to reduce VRAM and compute requirements.
\end{itemize}

\vspace{-5pt}
\section{Related Work}

\textbf{PDEs and Machine Learning}
Machine learning has sparked much resonance in accelerating and improving numerical PDE solvers as well as fully replacing them. Approaches that are combined with existing PDE solvers can replace components of the solver \cite{bar-sinai2019_Learning}, 
learn closure models \cite{Duraisamy_2019, Sirignano_2023_closures} or 
learn corrections to a fast solver on a coarse grid \cite{um2021solverintheloop, kochov_fluids_2021, dresdner2023learning}. Other directions target problems that are more difficult to address with numerical solvers, such as inverse problems \cite{raissi2019physics, bruna2022neural, DBLP:conf/nips/HolzschuhVT23}, or uncertainty quantification~\cite{xiao2019quantification,liu2024uncertainty}. In this work, we among others  target training our model architecture as a diffusion model for probabilistic predictions.
Leveraging diffusion models for autoregressive prediction and inverse problems has been explored by several works in the past \cite{lippe2023pde,kohl2024benchmark, shu2023physics, shehata2025improved}, albeit limited to data in 2D. 

\textbf{Pretraining and 3D Surrogates}
There has been an increased interest in training foundation models, which are pretrained on multiple PDE dynamics simultaneously \cite{subramanian2023foundation, yang2023context, mccabe2023multiple}. These pretrained models can be used either for zero-shot-predictions or for finetuning when adapting to new dynamics, which allows for improved results with fewer training samples. 
Important previous works have demonstrated learning PDEs in 3D,
e.g., for learning large-eddy simulations \cite{peng2023linear, li2023long, li2024transformer, jiang2025implicit}, and for elastic wave propagation \cite{lehmann20243d}. Smoke buoyancy in 3D was targeted as a test case by Li et al. \cite{li2023scalable}, while axial vision transformers with 2D-to-3D weight transfers were proposed by McCabe et al.~\cite{mccabe2023multiple}. Notably, most previous work targets resolutions of up to $64^3$, an exception being probabilistic experiments at a resolution of $128^3$ \cite{molinaro2024gencfd}, and surrogate training with up to $128\!\times\!128\!\times\!256$ \cite{ohana2024_TheWell}. This motivates our contributions for scalable architectures, as surrogate models for truly high-resolution 3D physics simulations are of paramount interest in different scientific areas, such as magnetohydrodynamics and compressible turbulence in astrophysics \cite{ohana2024_TheWell, burkhart2020catalogue}.

\textbf{Efficient Transformer Architectures}
Transformers have become a dominant backbone architecture in deep learning due to their high computational efficiency and their ability to model long-range causal relationships \cite{vaswani2017attention, devlin2018bert}. Transformers have also become a popular competitor to CNNs in vision and understanding tasks \cite{dosovitskiy2020image, rodrigo2024comprehensive} and have recently been adopted for learning surrogate models for physics simulations \cite{mccabe2023multiple, wu2024transolver, alkin2024universal}. A major computational difficulty is the quadratic complexity of the global self-attention mechanism. Different approaches have been proposed to address this limitation. For example, to restrict the computation of the attention operation to a local window~\cite{liu2021Swin}, or to compute the attention only across the data axes~ \cite{ho2019axial}. Others use hierarchical attention with carrier tokens for fast aggregation of information~\cite{hatamizadehfastervit}.
Alternatives directly modify the attention operator to achieve linear or near-linear complexity \cite{Cao2021transformer, DBLP:journals/corr/abs-2006-04768}. 

\section{Method}

\paragraph{Problem formulation} Let $\Omega_S$ denote a spatial domain with $n$ physical quantities $u(x,t) : \Omega_S \times [0,T] \rightarrow \mathbb{R}^n$ that are discretized in time and space and described by the temporal sequence $[\mathbf{u}_0, \mathbf{u}_{\Delta t}, ..., \mathbf{u}_T]$. We consider all additional information about the sequence such as the type of PDE or hyperparameters of the simulator to be encoded in an $m$-dimensional conditioning vector $\mathbf{c} \in \mathbb{R}^m$. We assume the availability of many such sequences as training data, representing the temporal evolution of different types of PDEs with varying initial conditions or coefficients. We denote our proposed network architecture by $\mathcal{M}_\theta$ with weights $\theta$.

We address two main tasks in this paper. The first is \textit{autoregressive prediction}: For a given sequence of $P$ preceding states $[\mathbf{u}_{t-P\Delta t}, ..., \mathbf{u}_{t-\Delta t}]$, denoted by $\mathbf{u}_{\mathrm{in}}$, our target is to predict the next state $\mathbf{u}_t := \mathbf{u}_{\mathrm{out}}$. 
The second task is to train a \textit{probabilistic sampler} 
to draw samples from a distribution of states representing solutions for a PDE as specified by 
the parameter vector $\mathbf{c}$. In this case $\mathbf{u}_\mathrm{in} = \emptyset$. 

\subsection{Backbone Architecture: P3D Transformer}

\begin{figure}
    \centering
    \includegraphics[width=.95\linewidth]{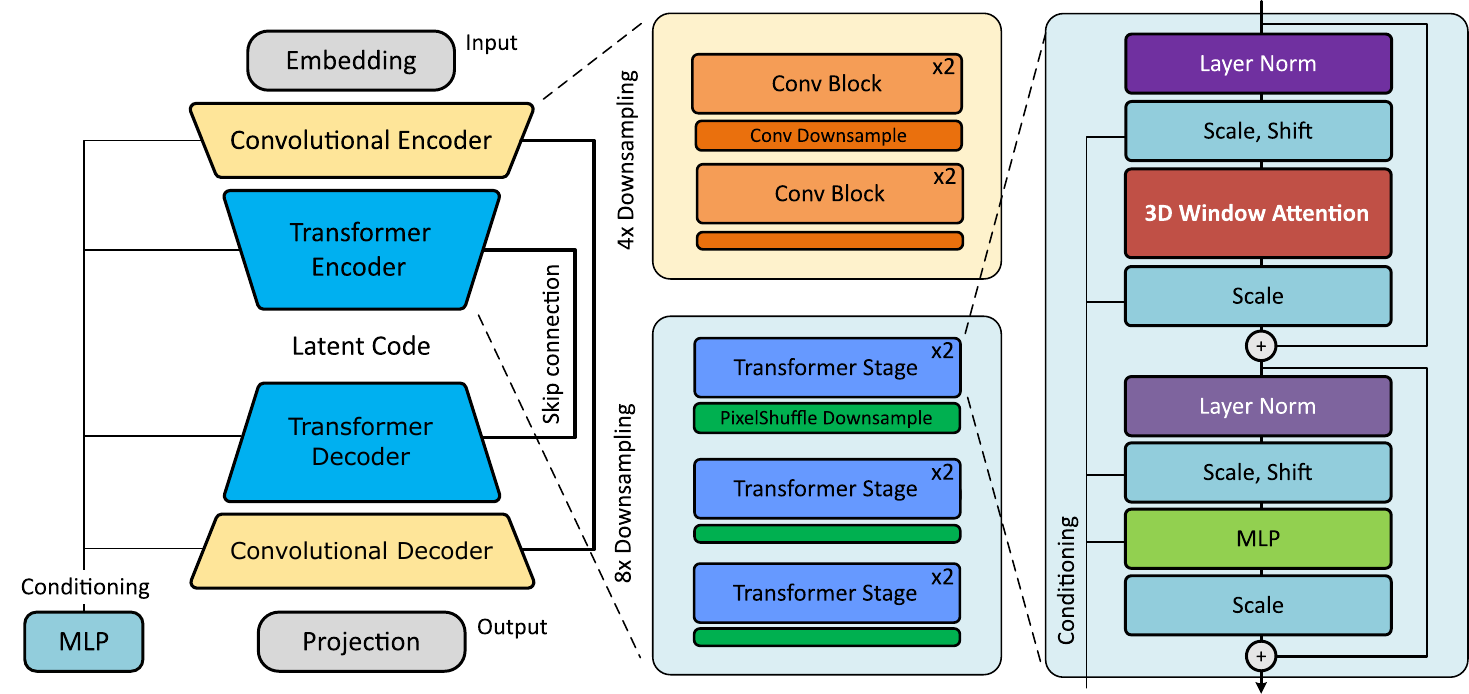}
    \caption{Overview of P3D. Convolutional blocks for local features processing are combined with transformers for deep representation learning, yielding a U-shaped multi-scale architecture. The transformer backbone combines windowed attention and conditioning via adaptive instance normalization, which are modified and optimized for 3D.}
    \label{fig:overview_architecture}
\end{figure}

The key components of the proposed P3D hybrid CNN-Transformer architecture are the hierarchical U-shape structure with the hybrid encoder- and decoder-pair based on convolutional and Transformer blocks. A visual overview is given in \cref{fig:overview_architecture}. 
In the following, we highlight the main components of the architecture and explain how they support its central goal to enable the efficient  inference of high-resolution 3D solutions.

\paragraph{Hybrid encoder/decoder} We rely on convolutional en- and decoders. Fully transformer-based architectures that work in the pixel space for 2D data and images like ViTs rely on a patchification operation to transform patches of size $p \times p$ into tokens. A corresponding approach in 3D would transform patches of size $p^3$ into a single token, significantly increasing the amount of information encoded in each token. To balance both the number of tokens for the transformer as well as the information density of each token, we learn local features via the convolutional encoder to obtain an optimized compressed representation. 
\paragraph{Attention and positional encoding} 
The self-attention operation used by transformers has quadratic complexity in the number of tokens. For 3D data, this becomes a major computational issue as the number of tokens grows cubically with the spatial discretization, leading to computational blow-up as the domain size increases. The central building block of our transformer encoder is the windowed multi-head self-attention ~\cite{liu2021Swin}, which only computes self-attention within a local region. For computing the attention scores between the tokens, we use the log-spaced relative positions of tokens inside the same window.

The architecture of the transformer encoder block 
combines the Swin Transformer~\cite{liu2021Swin} and the Diffusion Transformer \cite[DiT]{PeeblesDIT} into a 3D variant. 
It shares similarities with PDE-Transformer \cite{holzschuh25pde}, but significant alterations are: (1) the patchification is replaced by a large convolutional encoder and decoder, (2) we removed shifting of windows for computing the windowed attention. We did not see a noticeable drop in performance and decided to optimize to drop the shifting of windows for improved computational efficiency. 
The convolutional encoder/decoder follow the design of modern UNet blocks, using adaptive instance normalization and group normalization. We give a detailed description of the architecture in appendix B. 

\subsection{Global Context}

P3D intentionally does not use any absolute positional embeddings as well as no operations aggregating and distributing information globally. Thus, it has to rely on learning local features and dynamics within the perceptual field. This promotes translation-equivariance, which is an important inductive bias for surrogate modeling of PDEs. At the same time, global information and long-range dependencies often play a crucial role to obtain correct solutions. 
Our general strategy for learning large-scale simulations is to pretrain models on smaller crops of the simulation and then scale the trained networks to larger inputs. However, this does not allow for modeling long-range dependencies. To address this shortcoming, we link the bottleneck layers of the U-shape architecture with a sequence model.  

\begin{wrapfigure}{r}{0.55\textwidth}
  \centering
  \includegraphics[width=0.54\textwidth]{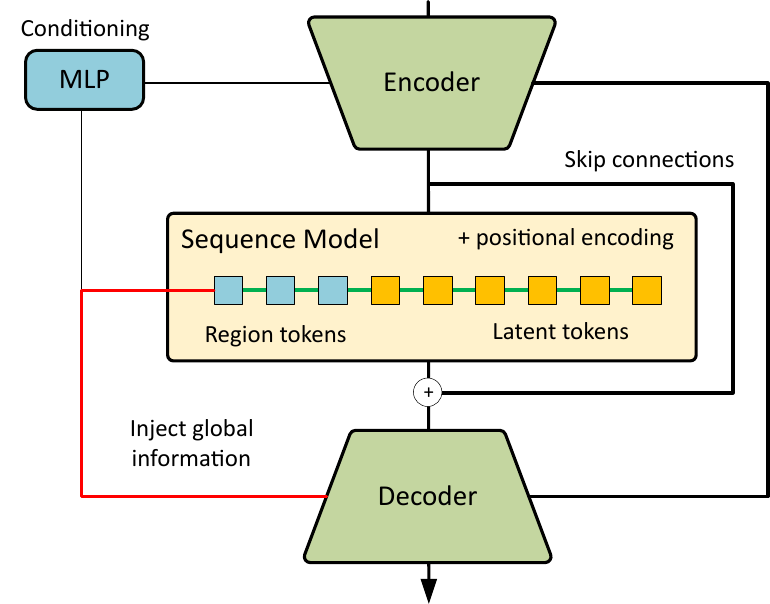}
  \caption{Global context via a sequence model. The bottleneck layers are connected to the sequence model, which embeds the bottleneck representation as latent tokens. Region tokens are used to inject global information directly into the decoder.} 
  \label{fig:context_network}
  \vspace{19pt}
  \centering
  \includegraphics[width=0.48\textwidth]{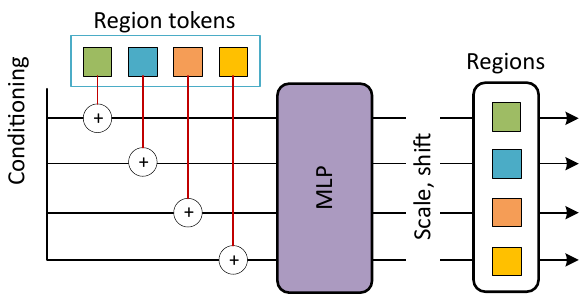}
  \caption{Conditioning via the region tokens. The input domain is partitioned into regions/crops, each of which has a corresponding messenger token. The regions are modulated individually via learnable scale and shift vectors based on the conditioning and region token. } \vspace{-10pt}
  \label{fig:region_token}
\end{wrapfigure}

\paragraph{Token embeddings} The bottleneck layers consist of tokens, which are embedded into latent tokens via a linear layer. P3D compresses a crop of size $32^3$ into a single latent token. Then, a frequency-based positional embedding vector is added to each latent token, similar to \cite{dosovitskiy2020image}. In addition, we partition the domain into \emph{regions} and we match the size of regions with the size of the domain crops P3D was pretrained on. For each region, we include a corresponding so-called \emph{region token} in the sequence of latent tokens, similar to the classification token in ViTs. Each region token is initialized via a learnable embedding layer and we add a frequency-based positional embedding vector. The purpose of region tokens is to serve as a more direct feedback mechanism to the decoder, which we describe in the next paragraph.
Our implementation uses $n=6$ layers with Hyper Attention \cite{han2023hyperattention}. \cref{fig:context_network} provides an overview of this setup. In principle, any efficient sequence model can be used. After the sequence of region and latent tokens is processed, the latent tokens are added to the input of the decoder via a skip connection.

\paragraph{Region tokens} The region tokens are retained and used as a more direct mechanism to let information flow through the decoder network. 
Region tokens are initialized as a learnable embedding vector with frequency-based positional encoding and are processed in the sequence model together with the latent tokens. 
Each region token corresponds to a crop that is processed independently of other crops by the encoder and decoder blocks. We use scale and shift operations to condition the decoder block of each region on the region token.  Within each decoder block, for each adaptive instance normalization layer with scale or shift operations, we transform the region token via an MLP layer to a region embedding vector which is added to the embedding vector of the conditioning $\mathbf{c}$. Each region gets modulated differently based on the region token. This is visualized in \cref{fig:region_token}. 

\paragraph{Saving memory}

During finetuning of the context network layer, it is not necessary to backpropagate gradients through the entire architecture. As working with 3D data requires large amounts of GPU memory, reducing the required memory is important. 
Instead, we can partition the input into individual crops and process them independently by the encoder. The context model then aggregates global information and distributes it again. Finally, the output is assembled from the individual patches. We introduce different setups for memory efficient training in \cref{sec:supdiff} and \cref{fig:scaling_larger_domains}.

\subsection{Training Methodologies}\label{sec:supdiff}

\vspace{-2pt}
\paragraph{Supervised training} 

For tasks that have a deterministic solution, such as training a surrogate model for a numerical solver, the P3D Transformer can be trained in a supervised manner using mean squared error (MSE) loss, enabling fast, single-step inference. In this case the network is directly optimized with the MSE 
\begin{equation}
    \mathcal{L}_S =
    \mathbb{E}
    \left[ || \mathcal{M}_\Theta (\mathbf{u}_\mathrm{in}, \mathbf{c}) - \mathbf{u}_\mathrm{out}||_2^2 \right].
\end{equation}

\vspace{-3pt}
\paragraph{Diffusion training} 
To allow for sampling from the full posterior distribution rather than approximating an averaged outcome, diffusion training is preferable for probabilistic solutions.
For such cases, we employ the flow matching formulation of diffusion models \citep{lipman2023, DBLP:conf/iclr/LiuG023, NEURIPS2020}. Given the input $\mathbf{u}_\mathrm{in}$ and a conditioning vector $\mathbf{c}$, samples $\mathbf{x}_0$ drawn from a noise distribution $p_0 = \mathcal{N}(0,I)$ are transformed into samples $\mathbf{x}_1$ from the posterior distribution $p_1$ by solving an ordinary differential equation (ODE) of the form $d\mathbf{x}_t = v(\mathbf{x_t}, t) ; dt$.
Then the model $\mathcal{M}_\Theta$ learns the velocity field $v$ by regressing a vector field that defines a probabilistic trajectory from $p_0$ to $p_1$. Samples along this trajectory are produced through the forward process 
\begin{equation}
\mathbf{x}_t = t \: \mathbf{u}_\mathrm{out} + [1-(1-\sigma_\mathrm{min})t] \: \epsilon \end{equation} 
for $t \in [0,1]$ with $\epsilon \sim \mathcal{N}(0,I)$ and a time threshold of $\sigma_\mathrm{min}=10^{-4}$. 
The velocity $v$ can be regressed by training via \begin{equation}
    \mathcal{L}_\mathrm{FM} = \mathbb{E}\left[||\mathcal{M}_\Theta (\mathbf{u}_\mathrm{in},\mathbf{x}_t,\mathbf{c},t) - \mathbf{u}_\mathrm{out} + (1-\sigma_\mathrm{min}) \epsilon||_2^2\right].
\end{equation}
After training, samples can be drawn from the posterior conditioned on $\mathbf{u}_\mathrm{in}$ and $\mathbf{c}$, by sampling $\mathbf{x}_0 \sim \mathcal{N}(0,I)$ and integrating the corresponding ODE $d\mathbf{x}_t = \mathcal{M}(\mathbf{u}_\mathrm{in},\mathbf{x}_t, \mathbf{c},t) \: dt$ over the time interval $t=0$ to $t=1$. We typically employ explicit Euler steps with a suitable, chosen step size $\Delta t$. 

\subsection{Scaling Output Domains} \label{sec:domain_scaling} 
\begin{figure}
    \centering
    \includegraphics[width=\linewidth]{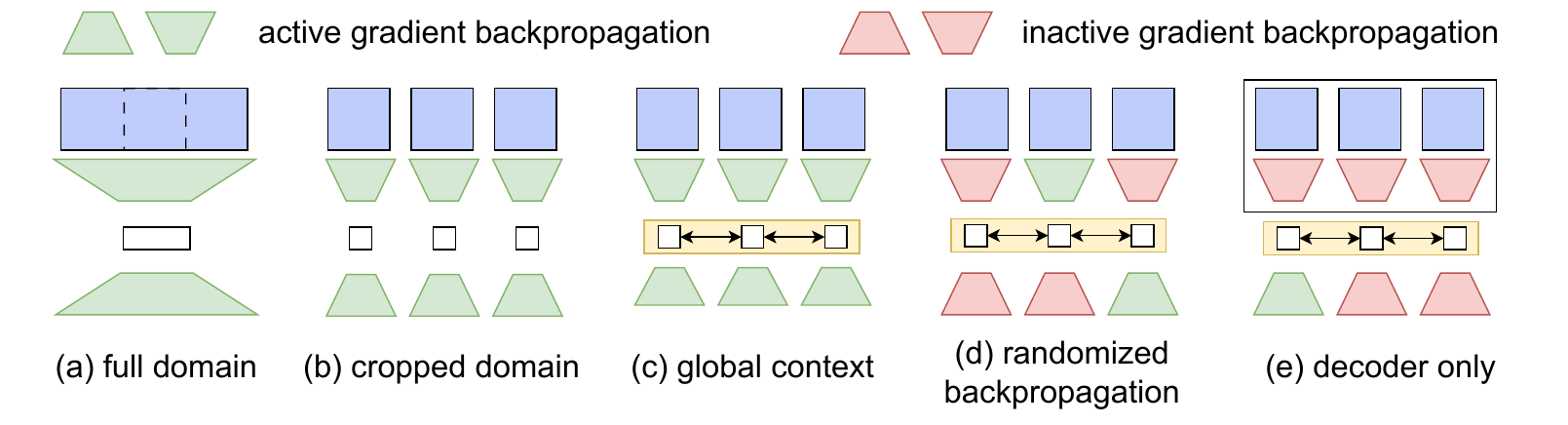}
    \caption{Different training and inference setups. (a) shows training on the full domain and (b) on domain crops. (c) includes the context network for global information processing, which can also be trained by randomly disabling gradient backpropagation for a percentage of the encoders and decoders, see (d). In (e) the latent codes from a pretrained encoder can be precomputed and only the context network and decoder are trained.}
    \label{fig:scaling_larger_domains}
\end{figure}

We consider different setups for training and inference, see \cref{fig:scaling_larger_domains}, which include training on the full domain (a), on crops (b) and different training variants in combination with the context network (c) to (e). While it is preferable to train on small crops due to compute requirements, for inference, we generally want to process the full domain. There are two naive strategies: (1) we scale to the full domain via the translation equivariance of the P3D architecture, i.e., we combine the domain crops and process them as a single input, and, (2) we encode and decode each crop of the full domain independently and combine the network outputs. We tag a model that is trained on crops of resolution $x^3$ and which internally runs inference on resolution $y^3$ by <$x$|$y$>. For example, a network trained on crops of size $64^3$ that is scaled via strategy (1) to resolution $128^3$ is tagged <$64|128$>, while the same network scaled via strategy (2) is tagged <$64$|$64$>. For strategy (2), if we use the context network for communication between the latent codes, we use the tag <X$x$|X$y$>.

\section{Experiments}

We evaluate our proposed hybrid CNN-Transformer architecture P3D,  
as well as different scaling approaches, in three experiments. In the first experiment, \cref{sec:multiple_pdes}, we focus on learning a set of different PDEs at the same time, a task popular for training foundation models \cite{mccabe2023multiple, herde2024poseidon, holzschuh25pde}. Based on the given input, the network has to identify the dynamics and predict the temporal evolution of the system. In the second experiment in \cref{sec:isotropic_turbulence}, we scale P3D to a large-scale simulation of isotropic turbulence with a spatial resolution of $512^3$. In the final experiment, we train P3D as a generative model for the equilibrium regime of turbulent channel flows with varying Reynolds number. 
Due to the long initial transient phase, the equilibrium states are very costly to compute for traditional solvers.
Directly producing a large number of samples from the turbulent equilibrium bypasses the warm-up period~\cite{lienen2023zero,lino2025dgn}, 
and as such provides an interesting and challenging scenario that is relevant for real-world applications. 
We will demonstrate below that P3D can be trained to faithfully recover the correct distributions at high spatial resolution.

Our model has 3 different configurations: $S$, $B$ and $L$ that determine the embedding dimension $d$ (32, 64 and 128 respectively) of the first layer. The corresponding models have $11$M to ca. $180$M parameters. We denote the configurations with P3D-$S$ for our model with the $S$ config, changed accordingly for $B$ and $L$.

\subsection{Jointly Learning Multiple PDEs} \label{sec:multiple_pdes}

\begin{wraptable}{r}{0.55\textwidth}
\centering
\vspace{-13pt}
\caption{Comparison of the normalized RMSE ($\times 10^{-2}$) on the test dataset. The reported value is the average over all $14$ types of PDEs.}
\label{tab:results}
\begin{tabular}{l c c c}
\toprule 
& \multicolumn{3}{c}{Crop size} \\ 
\cmidrule(lr){2-4} 
Method & $32^3$ & $64^3$ & $128^3$ \\
\midrule 
TFNO & 8.46 & 8.37 & - \\
FactFormer & 6.24 & 4.62 & - \\
UNet$_\mathrm{ConvNext}$ & 8.59 & 7.09 & - \\
UNet$_\mathrm{GenCFD}$ & 7.61 & 8.04 & 8.27 \\
AViT & 20.9 & 25.0 & 15.1 \\
Swin3D & 7.92 & 7.04 & 5.03 \\
AFNO & 9.95 & 4.98 & 4.79 \\
\midrule
P3D-S & 6.27 & 3.76 & 3.33 \\
P3D-B & 4.69 & 3.03 & 2.52 \\
P3D-L & \bf 4.13 & \bf 2.49 & \bf 2.08 \\ 
\bottomrule
\end{tabular}
\end{wraptable}

Our dataset for this task comprises $14$ different types of PDEs, e.g., Burger's equation, Kuramoto-Shivashinsky, Gray-Scott, Swift-Hohenberg and many others.  The dataset is based on APEBench \cite{koehler2024_ApeBench}, and a full description of each PDE with visualizations can be found in \cref{app:datasets} in the appendix. 
For all PDEs except Gray-Scott, we consider $60$ different simulations with varying initial conditions and PDE-specific parameters such as viscosity, domain extent or diffusivity. For Gray-Scott, we include $10$ simulations for each of its hyperparameters. 
Each spatially periodic simulation contains $20$ snapshots discretized at resolution $384^3$. 
We evaluate and benchmark models on random crops of the simulation domain of size $128^3, 64^3$ and $32^3$. 
Even with full information about simulation hyperparameters and the type of PDE, the behavior is not fully deterministic as quantities beyond the cropped regions can influence the solution inside it, e.g., for transport into the region.
Simulations have a different number of physical channels and we zero-pad data with fewer channels than the number of maximum channels $N_C = 3$. For training, we use a fixed learning rate of $2.0 \cdot 10^{-4}$ for all models with the AdamW optimizer with weight decay $10^{-15}$ and batch size $256$ in bf16-mixed precision. The only input to the networks is the observed state $\mathbf{u}_\mathrm{in}$. Note that the type of PDE and simulation hyperparameters are unknown to the models. 

\paragraph{Training on cropped data} Cropping affects the boundary condition. It yields a time-dependent boundary condition on the cropped data, which is not known by the model. This can be seen as an extension and more difficult variant of the multi-physics training \cite{subramanian2023foundation, yang2023context, mccabe2023multiple}, where in addition to not knowing the PDE or simulation hyperparameters, the model has to estimate the boundary conditions in a data-driven manner. Mathematically, the model $\mathcal{M}_\theta$ is trained to regress 
\begin{align}
\argmin_\Theta \: \mathbb{E}_{(s_t,s_{t+\Delta t}, \mathbf{c}) \in \mathcal{D}_\mathrm{train}} [ \mathbb{E}_{(s_{t}^{\mathrm{crop}}, s_{t+\Delta t}^{\mathrm{crop}})}[ || \mathcal{M}_\Theta(s_{t}^{\mathrm{crop}}, \mathbf{c}) - s_{t+\Delta t}^{\mathrm{crop}}||_2^2]], 
\end{align}
where we sample $(s_t,s_{t+\Delta t}, \mathbf{c}) \in \mathcal{D}_\mathrm{train}$ from the training dataset and apply a random cropping to obtain $(s_{t}^{\mathrm{crop}}, s_{t+\Delta t}^{\mathrm{crop}})$. The input $\mathbf{u}_\mathrm{in}$ corresponds to $s_{t}^{\mathrm{crop}}$ and $\mathbf{u}_\mathrm{out}$ to $s_{t+\Delta t}^{\mathrm{crop}}$.
The mapping $s_{t}^{\mathrm{crop}} \mapsto s_{t+\Delta t}^{\mathrm{crop}}$ is not deterministic since the boundary conditions are not prescribed. The model $\mathcal{M}_\Theta$ has to learn a prediction that minimizes the prediction error w.r.t. all possible simulation states that are outside the cropped domain, i.e., the optimal prediction $s^*$ for $s_{t+\Delta t}^{\mathrm{crop}}$ minimizes
\begin{align}
   s^* = \argmin_s \: \mathbb{E}_{(\hat{s}_{t}, \hat{s}_{t+\Delta t}) \sim \mathcal{D}_\mathrm{train}} \left[ || \mathrm{crop}(\hat{s}_{t+\Delta t}) - s ||_2^2\:|\:\mathrm{crop}(\hat{s}_{t}) = s_{t}^{\mathrm{crop}}\right],
\end{align}
\begin{minipage}{\textwidth} where we use the crop operation $\mathrm{crop}(\cdot)$ that was used for $s_{t}^{\mathrm{crop}}$. Hence, the model has to implicitly learn the time-dependent boundary condition. The performance depends on how well the model is able to extrapolate the dynamics outside the cropped input for a short prediction horizon.
\end{minipage}

\paragraph{Baseline methods} We consider a wide range of SOTA baseline architectures for evaluation. Specifically, we include Swin3D, our own implementation of the SwinV2 architecture \cite{liu2021Swin} extended to 3D, AViT, an axial vision transformer~\cite{mccabe2023multiple}, Adaptive FNOs \cite[AFNO]{guibas21_afno}, which uses token mixing via attention in the Fourier domain, Tucker-Factorized FNOs \cite[TFNO]{kossaifi2023multi}, and FactFormer \cite{li2024scalable},  a transformer-based model that uses an axial factorization of kernel integrals. Additionally, we consider two different convolutional UNet architectures, one  based on ConvNeXt blocks \cite{ohana2024_TheWell}, as well as a large, generative UNet \cite{molinaro2024gencfd}. 
We train all models for $1000$ epochs on four H100 GPUs.
\begin{figure}[t] 
    \centering 
    \includegraphics[width=.36\linewidth]{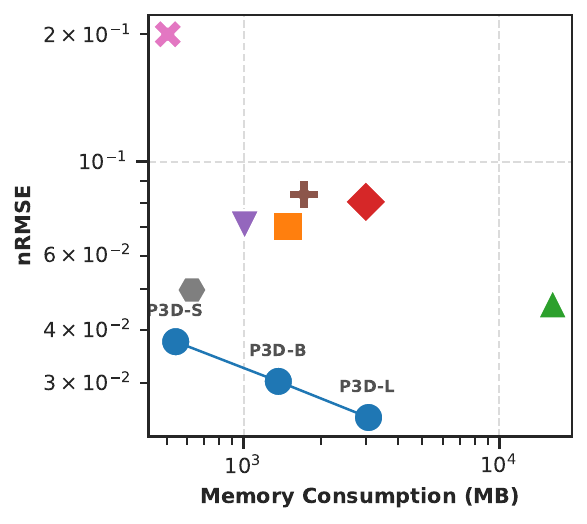}
    \includegraphics[width=.54\linewidth]{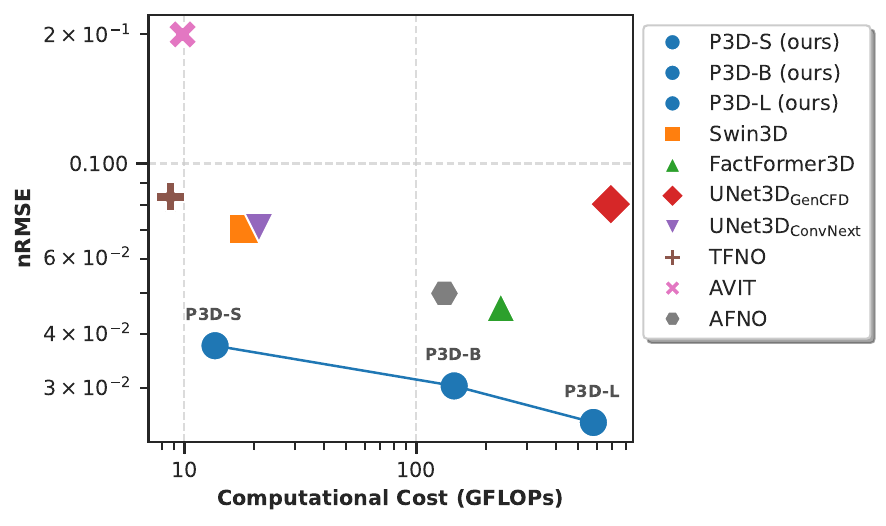}
    \caption{Comparison of model accuracy vs. (left) memory usage during backpropagation and (right) computational costs for inference for jointly learning different types of PDEs with crops of size $64^3$ for P3D and baselines.}
    \label{fig:comparison_baselines} 
\end{figure}
\paragraph{Results}  
Overall, as the crop size increases, the observed simulation domains becomes larger, increasing the amount of information that becomes available to the network. At the same time, regions near the border become smaller, thus decreasing the relative weight of areas that are more difficult to predict than the center. We see an according decrease in the normalized RMSE across all methods, as shown in \cref{tab:results}. See \cref{app:datasets} in the appendix for detailed evaluations of each PDE type. We don't consider any inference strategy for upscaling in this experiment.
Our P3D networks perform best across all crop sizes. Performance significantly improves when scaling the model size from $S$ to $L$.
\paragraph{Memory and compute} For scaling an architecture to high-resolution 3D simulations, the memory requirements as well as the inference speed are essential. Transformers architectures have been shown to achieve improved accuracy as the number of parameters and floating point operations increase \cite{holzschuh25pde}, therefore comparing different architectures needs to take both factors into account. In \cref{fig:comparison_baselines}, we compare the MSE at patch size $64^3$ against (1) the computational cost for inference measured in GFLOPs, and (2) the VRAM consumption in MB for a backward pass with batch size 1.
The P3D networks provide the best tradeoff between accuracy and computational cost/memory requirements.
\subsection{Isotropic Turbulence} \label{sec:isotropic_turbulence}
The goal of the next experiment is to scale P3D to a high-resolution simulation involving complex dynamics.
For this, we consider forced isotopic turbulence simulated via direct numerical simulation (DNS) at resolution $1024^3$ provided by the John Hopkins Turbulence Database \cite{perlman2007_JHTDB}. The dataset
is downsampled from the original resolution to $512^3$ with a total of $500$ snapshots, which are saved after reaching a statistical stationary state.
The data is split into test and training sets, where the first $420$ snapshots are used for training and the last $80$ snapshots for testing. The flow field resolution of 
$512^3$ means that every model call of P3D generates over 400 million data points.
\begin{figure}[t]
    \centering
    \includegraphics[width=\linewidth]{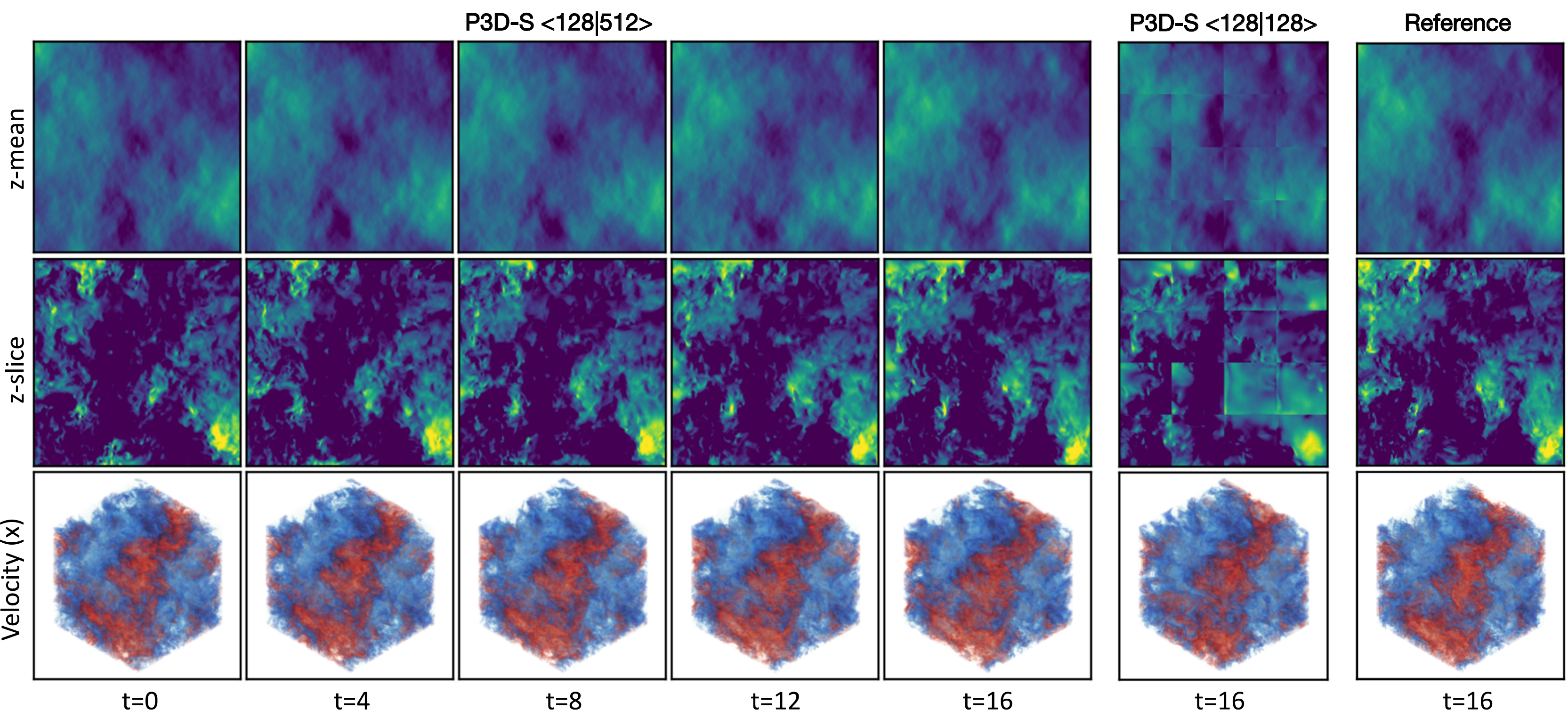}
    \caption{Forced isotropic turbulence. Prediction on the test set at resolution $512^3$ with an autoregressive rollout of $16$ steps. The model is pretrained on patches of size $128^3$, without finetuning on $512^3$. P3D-$S$ <128|512> encodes and decodes all region crops simultaneously, whereas P3D-$S$ <128|128> processes crops of size $128^3$ independently. In the latter case, boundaries between crop regions become apparent for longer rollouts.}
    \label{fig:turbulence_vis}
\end{figure}

\paragraph{RMSE and spectral error for crops $128^3$} We first evaluate the performance of P3D compared to baselines from the previous experiment trained on domain crops of size $128^3$. We consider an evaluation of autoregressive rollouts from $1$ to $15$ steps. We evaluate the RMSE and a spectral error based on the enstrophy graph, which we compute based on the vorticity that is derived from the velocity fields of the data, see \cref{app:isotropic_turbulence} in the appendix for details. We train all models for $4000$ epochs using the same setup as in \cref{sec:multiple_pdes}, but reduce the batch size to $32$. We show the results in \cref{tab:isotropic_turbulence_eval}. P3D performs best and achieves the best RMSE and spectral error. Continuing training for more epochs further improves results, showing that P3D does not overfit. As more autoregressive rollout steps are used, differences to other models regarding the RMSE become less pronounced, which is expected, as the prediction starts to deviate from the reference due to the uncertainty from the boundary of the crop regions. However, the spectral error remains significantly better.

\begin{table}[tb]
\caption{Performance comparison on the test set with crop size $128^3$ for RMSE ($\times 10^{-2}$) and the L2 enstrophy graph error ($\times 10^{2}$) at different autoregressive rollout steps.}
\vspace{5pt}
\label{tab:isotropic_turbulence_eval}
\centering
\begin{tabular}{lccccccccl}
\toprule
\textbf{Model} & \multicolumn{4}{c}{\textbf{RMSE}} & \multicolumn{4}{c}{\textbf{L2 Enstrophy}} & \textbf{epochs} \\
\cmidrule(lr){2-5} \cmidrule(lr){6-9}
& \textbf{1} & \textbf{5} & \textbf{10} & \textbf{15} & \textbf{1} & \textbf{5} & \textbf{10} & \textbf{15} & \\
\midrule
UNet$_\mathrm{GenCFD}$ & 5.48 & 25.42 & 48.60 & 67.72 & 4.25 & 14.5 & 22.7 & 140 & 4000 \\
Swin3D & 4.44 & 15.42 & 25.83 & 34.57 & 11.7 & 103 & 169 & 201 & 4000 \\
AViT & 9.45 & 19.57 & 30.00 & 37.77 & 26.70 & 49.1 & 84.6 & 112 & 4000 \\
AFNO & 3.69 & 13.33 & 23.52 & 29.80 & 7.69 & 88.7 & 158 & 190 & 4000 \\
\midrule
P3D-S & 2.81 & 9.87 & 20.50 & 28.25 & 2.15 & 8.23 & 21.6 & 31.9 & 4000 \\
& 2.17 & 8.99 & \bf 19.40 & \bf 27.40 & 1.29 & 6.68 & 16.0 & 25.3 & 20000 \\
\midrule
P3D-B & 2.04 & 8.79 & 20.23 & 31.52 & 0.72 & 1.39 & \bf 3.38 & 19.20 & 4000 \\
& \bf 1.54 & \bf 8.11 & 21.09 & 44.92 & \bf 0.21 & \bf 0.71 & 3.49 & \bf 14.70 & 20000 \\
\bottomrule
\end{tabular}
\end{table}

\begin{wraptable}{r}{0.6\textwidth} 
\centering
\vspace{-12pt}
\caption{Forced Isotropic Turbulence. RMSE ($\times 10^{-2}$) on the test set for P3D trained on crop size $128^3$ for different scaling strategies.}
\label{tab:isotropic_turbulence}
\begin{tabular}{l c c c} 
\toprule
& \multicolumn{3}{c}{Scaling strategy} \\
\cmidrule(lr){2-4} 
Method & <$128$|$128$> & <$128$|$256$> & <$128|512$> \\
\midrule 
P3D-S & 1.90 & 1.68 & 1.60 \\
P3D-B & 1.38 & 1.24 & - \\
\bottomrule 
\end{tabular}
\vspace{-10pt}
\end{wraptable}

\paragraph{Scaling P3D to $512^3$} We report results for scaling the P3D models trained on crops of size $128^3$ to the full domain $512^3$. The model weights for this evaluation are the EMA weights at epoch $2000$. Training for $2000$ epochs took 11h 48m and 20h 25m respectively on $4$ A100 GPUs. We compare the performance of P3D when scaling the network to larger domain sizes using the two naive strategies introduced in \cref{sec:domain_scaling}. P3D <128|128> processes blocks independently using the original training resolution. Thus the domain $512^3$ is split into $64$ blocks of size $128^3$ that are processed independent of each other. P3D <128|512> processes the full domain, leveraging the translation-equivariance of the architecture.
\paragraph{Results}
We give an evaluation of the RMSE in  \cref{tab:isotropic_turbulence}. Increasing the domain size during inference consistently improves the RMSE. With increasing domain size, there are relatively fewer areas close to the boundary, thus the uncertainty of turbulent motions is reduced and networks are able to provide more accurate predictions. Similar to previous results, larger networks improve performance as well. Note that the RMSE is different between \cref{tab:isotropic_turbulence_eval,tab:isotropic_turbulence}, since we do not consider longer rollouts in \cref{tab:isotropic_turbulence} and thus the test set is different. In this experiment, due to the isotropic and homogeneous nature of the simulation, we achieve good results without requiring global information. 

\begin{figure}[t]
    \centering
    \includegraphics[width=\linewidth]{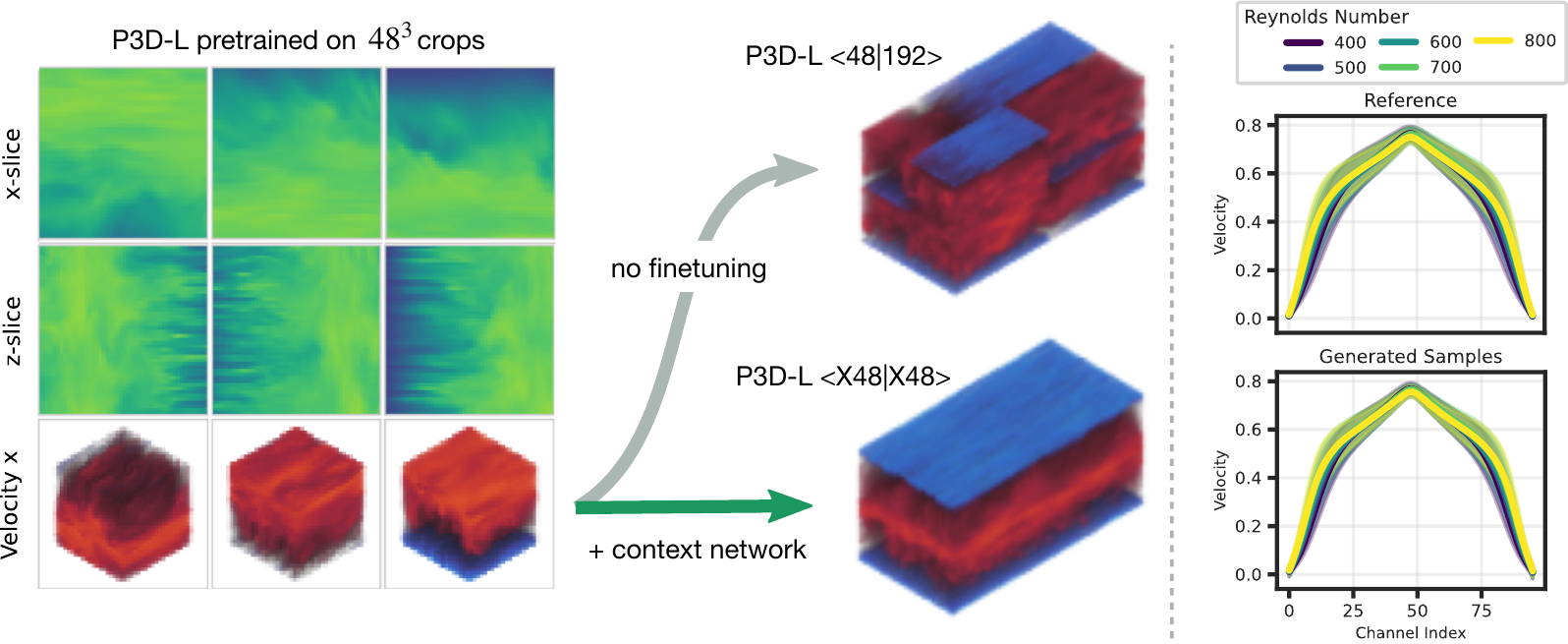}
    \caption{Turbulent channel flow. We pretrain PDE-$L$ on crops of size $48^3$. For P3D-$L$ <48|192> scaled to the full domain $96^2 \times 192$, the relative positions of crops to the wall is critical. Without absolute positional encodings and global information passing, the predicted samples are incorrect. Finetuning with the global context network on the full domain solves this issue. 
    The flow statistics (right) show a close match between the time-resolved DNS reference and  the independent samples produced by P3D-$L$ <X48|X48> conditioned with  varying Reynolds numbers.
    }
    \label{fig:tcf}
\end{figure}

\subsection{Turbulent Channel Flow}
For the last experiment we train P3D as a generative model for learning to sample from a turbulent channel flow simulation: 
it contains a periodic channel with no-slip boundaries at $\pm$y and is driven by a dynamic forcing to prevent a loss of energy.
The seemingly simply geometry represents a well-studied and relevant scenario that is highly challenging as it requires fine
spatial and temporal discretizations with correspondingly long simulation times to provide converged turbulence statistics \cite{hoyas2008}.
Due to the very costly initial transient phase of these simulations, it is especially attractive to phrase the turbulent channel flow (TCF)
problem as a probabilistic learning problem, where states from the relevant equilibrium phase can be sampled directly, i.e.
without resolving the initial warm-up phase.

\newcommand{\spacetweak}{\vspace{-5pt}}

\spacetweak{}
\paragraph{Dataset} We generate a dataset comprising 20 simulations with Reynolds numbers $\mathrm{Re}$ within the interval $[400, 800]$ spaced equidistantly. After the initial-warmup phase, we simulate $ETT=20$ eddy turnover cycles, which we save in $200$ snapshots with $\Delta t = 0.1$. The computational grid comprises $96 \times 96 \times 192$ spatially adaptive cells with a finer discretization near the wall. The data contains channels for the velocity in X/Y/Z direction as well as pressure. As the grid points in the computational grid are not equidistant, they are remapped into physical space for analysis and visualization. However, we train P3D directly with computational grid data.

\begin{table}[htbp]
\caption{Turbulent channel flow quantitative statistical evaluation. L2 distance between the first two moments of the velocity profiles along the channel flow direction. For the finetuned models, we report the mean and standard deviation over $5$ finetuning trainings initialized with different seeds.}
\label{tab:performance_comparison}
\vspace{5pt}
\centering
\begin{tabular}{lllll}
\toprule
\textbf{Model} & \textbf{Mean L2} ($10^{-5}$) & \textbf{Std. Dev. L2} ($10^{-5}$) & \textbf{VRAM} (GB) & \textbf{epochs} \\
\midrule 
UNet$_\mathrm{GenCFD}$ full domain & 132.38 & 17.66 & 17.45 & 400 \\
AFNO full domain & 28.73 & 1849.3 & 3.45 & 400 \\
P3D-$L$ full domain & 3.02 & 13.20 & 14.9 & 400 \\
P3D-$L$ <48|192> & 5862.81 & 233.77 & 2.82 & 2000 \\
\midrule
finetune w/o region tokens & 4541 ± 495 & 2026 ± 267 & 15.85 & 20 \\
finetune & 23.6 ± 21.4 & 40.4 ± 49.4 & 15.85 & 20 \\
finetune, decoder only & 941 ± 484 & 1170 ± 392 & 6.04 & 20 \\
finetune, decoder only & 97.7 ± 102.6 & 131 ± 149 & 6.04 & 100 \\
finetune, decoder only & 16.8 ± 5.0 & 24.1 ± 17.2 & 6.04 & 500 \\
\bottomrule
\end{tabular}
\end{table}

\spacetweak{}
\paragraph{Global context for P3D-$L$ trained on $48^3$} We train P3D-$L$ as a diffusion model following \cref{sec:supdiff}, on the full domain and crops of size $48^3$.
Pretraining on crops requires significantly less compute and VRAM and the network converges much faster. After pretraining with patches, we assemble the full domain. As there are no absolute positional encodings and patches are processed individually, the per-patch processing cannot correlate the patch content when denoising. 
As such, they are not aware of the position of the wall relative to the position of the patch, and the global structure of the assembled flow is incorrect. 
When finetuning with the context network <X48|X48>, information about the relative positions of the patches can be distributed, and the model is able to produce physically meaningful global solutions. 

\paragraph{Statistical evaluation}
Let $\mathbf{x}=(x_1,x_2,x_3)$ denote the spatial coordinates and $u(\mathbf{x}, t, \mathrm{Re})$ denote the velocity of the flow direction X. The reference simulations reach an equilibrium phase after the initial transient phase of the warmup. Therefore, for the reference simulations, the moments $u_m(\mathbf{x}, t, \mathrm{Re})$ should be the same for all $t$. Additionally, the setup combining periodic boundary conditions and no-slip boundaries for the wall implies that the statistics \emph{only} depend on the distance to the wall on the flow axis $x_1$, i.e., $u_m((x_1,x_2,x_3), t)$ is the same for all $x_2$ and $x_3$ inside the domain. Thus it is reasonable to consider $u_m(x_1, \mathrm{Re})$ and calculate the moments by sampling over $x_2,x_3, t$. Since the baseline methods cannot be conditioned on the Reynolds number $\mathrm{Re}$, we also compute the velocity profile over $\mathrm{Re}$. The corresponding moments of the velocity profile $u_m(x_1)$ for $m=1,2$ are used to compute metrics based on the L2 distance between moments of the velocity profile graph $x_1 \mapsto u_m(x_1)$ of the reference and predicted samples. The resulting distances for different baselines and the first two moments are shown in \cref{tab:performance_comparison}. We additionally train two baselines, AFNO and UNet$_\mathrm{GenCFD}$ with identical training setups. 
The distribution of the mean velocity along the flow direction provides a meaningful statistical metric to evaluate accuracy. 
The reference statistics are computed with the time-resolved DNS solutions. 

\paragraph{Moments of the flow field} We included comparisons of the first three moments (mean, variance, skewness) $u_m$ of the flow direction (velocity in x-direction) averaging over $x_1$ as well. We report the standard deviation when estimating the moments of the reference based on randomly drawing 20 simulations states of the equilibrium phase per Reynolds number as done for the velocity profiles to properly assess how close the generated samples should match the reference. For the finetuned P3D-$L$ <X48|X48>, we picked the best model out of the five finetuned models. See  \cref{tab:flow_moments}. Overall, P3D-$L$ trained on the full domain and finetuned achieve the best results.

\begin{figure}[tp]
\caption{First three moments of the velocity field in the flow direction for the reference, P3D-$L$ and the two baseline methods.}
\label{tab:flow_moments}
\centering
\begin{tabular}{lccc}
\toprule
\textbf{Moment} & \textbf{Mean} & \textbf{Variance} & \textbf{Skewness} \\
\midrule
Reference & 0.5034$\pm$0.0007 & 0.0511$\pm$0.0001 & -0.776$\pm$0.007 \\
\midrule
UNet$_\mathrm{GenCFD}$ full domain & 0.5002 & 0.0532 & -0.723 \\
AFNO full domain & 0.5040 & 0.0930 & -0.361 \\
\midrule
P3D-$L$ full domain & 0.5009 & 0.0513 & -0.789 \\
P3D-$L$ <X48|X48> & 0.5044 & 0.0510 & -0.802 \\
\bottomrule
\end{tabular}
\vspace{-10pt}
\end{figure}

\paragraph{Ablation on finetuning}

We evaluate the effects of finetuning the P3D models trained on the cropped domain $48^3$, see \cref{tab:performance_comparison}. The model \emph{finetune} corresponds to the P3D-$L$ <X48|X48> model where we use different training setups, as shown in \cref{fig:scaling_larger_domains}. For finetuning with and without region tokens, we use the setup \ref{fig:scaling_larger_domains}c, which backpropagates gradients through all crops and network modules. Finetuning takes ca. 65 minutes on 4 H100 GPUs and requires more memory than training on the full domain, due to the additional context network, but only requires few epochs for achieving good results. For the decoder only finetuning, we only backpropagate gradients through the decoder and context network, and also only backpropagate through 10\% randomly selected decoder blocks, which corresponds to setup \ref{fig:scaling_larger_domains}e. We do not precompute the bottleneck representations from the frozen encoders, but this could be done to further reduce the VRAM and compute requirements. Finetuning for this case and 20 epochs takes ca. 43 minutes. Overall, this setup achieves a significant reduction in VRAM, but requires more training epochs. The generated velocity profiles for P3D-$L$ <X48|X48>, shown on the right side of \cref{fig:tcf}, qualitatively show the high accuracy of the P3D-$L$ model over the full range of Reynolds numbers. See also \cref{fig-app: tcf methods comparison,fig-app: tcf samples full domain,fig-app: tcf samples context,fig-app: tcf samples cropped upscaled} in the appendix for visualizations. While finetuning matches the statistics well, there are still visible discontinuities between generated crop regions, which leaves further room for improvement.

\begin{wrapfigure}{r}{0.5\linewidth}
\vspace{-10pt}
\caption{Speedup and Timings}
\label{tab:speedup_timings}
\centering
\begin{tabular}{lcc}
\toprule
\textbf{Method} & \textbf{samples/s} & \textbf{Speedup} \\
\midrule
P3D-L & 0.17 & 144x \\
AFNO & 1.53 & 1246x \\
UNet$_\mathrm{GenCFD}$ & 0.10 & 81x \\
\midrule
DNS (GPU) & 0.0022 & - \\
\bottomrule
\end{tabular}
\vspace{-10pt}
\end{wrapfigure}

\paragraph{Benchmarking speedups}

For this experiment, we can also provide speedups compared to our GPU-based solver that we used to generate the data set. It is important to note that the warmup phase of the DNS takes on average 2 hours and 24 minutes, which has to be included as the generating solver cannot skip this phase.

Timings are obtained on a L40 GPU with 100 steps for inference. For the calculation of the speedup, we assume 20 samples are sufficient to compute the statistics for each Reynolds number and we can select every 10th sample of the DNS to avoid high autocorrelation of samples. While AFNO has a higher speedup, the statistics are not sufficiently accurate and thus cannot be used.

\section{Conclusion} We have presented \textit{P3D}: an efficient hybrid CNN-Transformer architecture for learning surrogates for high-resolution 3D physics simulations. We demonstrated the strong advantages of P3D over a comprehensive list of baselines for simultaneously learning different types of PDEs, showing improved accuracy and stable training while at the same time being faster and more memory efficient than the strongest competitors. 

We scaled P3D to a high-resolution isotropic turbulence simulation by pretraining on smaller crops from the domain, and demonstrated its capabilities as a probabilistic generative model. The P3D model accurately predicts distributions of high-resolution velocity and pressure fields for a turbulent channel flow with varying Reynolds numbers, demonstrating how to include global information and coordinated pretrained networks via a global context model. Our architecture establishes the foundation for scaling scientific foundation models to very high resolutions, unlocking their potential to deliver real-world impact across scientific domains.

\section*{Achnowledgements}

{ 
This work was supported by the ERC Consolidator Grant \textit{SpaTe} (CoG-2019-863850). 

The authors also gratefully acknowledge the support of the Erlangen National High Performance Computing Center (NHR@FAU) of the Friedrich-Alexander-Universität Erlangen-Nürnberg (FAU) under the BayernKI project \textit{B278BB}. BayernKI funding is provided by Bavarian state authorities. NHR funding is provided by federal and Bavarian state authorities. NHR@FAU hardware is partially funded by the German Research Foundation (DFG) – 440719683. }

\clearpage 

\printbibliography



\newcommand{\dAdv}{\texttt{adv}}
\newcommand{\dAdvDiff}{\texttt{adv-diff}}
\newcommand{\dBurgers}{\texttt{burgers}}
\newcommand{\dDecayTurb}{\texttt{decay-turb}}
\newcommand{\dDiff}{\texttt{diff}}
\newcommand{\dDisp}{\texttt{disp}}
\newcommand{\dFisher}{\texttt{fisher}}
\newcommand{\dGS}{\texttt{gs}}
\newcommand{\dGSalpha}{\texttt{gs-alpha}}
\newcommand{\dGSbeta}{\texttt{gs-beta}}
\newcommand{\dGSgamma}{\texttt{gs-gamma}}
\newcommand{\dGSdelta}{\texttt{gs-delta}}
\newcommand{\dGSepsilon}{\texttt{gs-epsilon}}
\newcommand{\dGStheta}{\texttt{gs-theta}}
\newcommand{\dGSiota}{\texttt{gs-iota}}
\newcommand{\dGSkappa}{\texttt{gs-kappa}}
\newcommand{\dHyp}{\texttt{hyp}}
\newcommand{\dKDV}{\texttt{kdv}}
\newcommand{\dKolmFlow}{\texttt{kolm-flow}}
\newcommand{\dKS}{\texttt{ks}}
\newcommand{\dSH}{\texttt{sh}}
\newcommand{\dChannel}{\texttt{channel}}
\newcommand{\dIsoTurb}{\texttt{iso-turb}}
\newcommand{\dMHD}{\texttt{mhd}}
\newcommand{\dTransBL}{\texttt{trans-bl}}
\newcommand{\dActMat}{\texttt{act-mat}}
\newcommand{\dHelmStair}{\texttt{helm-stair}}
\newcommand{\dRayBen}{\texttt{ray-ben}}
\newcommand{\dShearFlow}{\texttt{shear-flow}}
\newcommand{\dTurbRad}{\texttt{turb-rad}}

\clearpage

\begin{refsection}

\appendix 

\section{Network Architectures}

We provide additional details on the backbone architecture of P3D, supplementing the information in the main paper.

\paragraph{Embedding of time, class labels and physical parameters} We combine all three types of conditionings within a combined embedding layer. Time for flow matching/diffusion and physical parameters are implemented via timestep embeddings. Class labels are implemented via label embeddings. The embedding vectors of all three types are added and used as the joint embedding. The embedding dimension for each in the convolutional encoder/decoder is 64. In our experiments, class labels are not used.

\paragraph{Convolutional encoder} The convolution encoder first embeds the input using a Conv3D layer (kernel size $3$, padding 1) with filters that correspond to the embedding dimensions of the configuration. This is followed by downsampling layers implemented via Conv3D layers (kernel size $3$, padding 1, stride 2). Intermediate states before each downsampling operations are saved for residual connections. Encoder blocks and consecutive downsampling are applied twice. 
For each layer, the corresponding number of filters is shown in \cref{tab-app: architecture parameters}.
Encoder blocks are repeated twice. 
Each encoder block consists of GroupNormalization layers, followed by GELU activations, Conv3D layer (kernel size 3, padding 1), GroupNormalization, modulation via shift and scale operations depending on the conditioning, GELU and an additional Conv3D layer (kernel size 3, padding 1). The input and output of each encoder block are connected via skip connections. The shift and scale vectors are learned via linear layers from the embedding vectors of the convolutional encoder/decoder. 

\paragraph{Convolutional decoder}
The design of the convolutional decoder mirrors the convolutional encoder in a U-shape architecture with residual connections. Upsampling layers are implemented via a combination of Conv3D layers to increase the number of filters and PixelShuffle3D layers. For an input number of channels $C_\mathrm{in}$ and a target number of channels in the upsampled output $C_\mathrm{out}$, the Conv3D operation first expands the number of channels $C_\mathrm{in} \times H \times W \times D \to 8 \, C_\mathrm{out} \times H \times W \times D$ and PixelShuffle3D spatially rearranges the pixels $8 \, C_\mathrm{out} \times H \times W \times D \to C_\mathrm{out} \times 2H \times 2W \times 2D$.

\begin{table*}[ht]
  \caption{Different configurations $S$, $B$ and $L$ of P3D. Table shows the total number of weights, the number of filters within the convolutional encoder/decoder and the number of groups for GroupNormalization layers.}
  \label{tab-app: architecture parameters} %
  \centering
  \small
  \vspace{0.1cm}
  \begin{tabular}{l l l l}
 \toprule
 \textbf{Configuration} & Number of parameters & Embedding dimensions & Number of groups \\
 \midrule
S & 11.2M & [32, 32, 64] & 16 \\
B & 46.2M & [64, 128, 128] & 32 \\
L & 181M &  [128, 256, 256] & 32 \\
 \bottomrule
 \end{tabular}
\end{table*}

\paragraph{Performance comparison of baseline architectures}

Below, we summarize the numbers of the different architectures used in the experiments for an input of size $128^3$ with a single channel, see \cref{tab:performance_comparison_architecture}.

\begin{table}[h]
\caption{Performance comparison of architectures.}
\label{tab:performance_comparison_architecture}
\centering
\begin{tabular}{lcccc}
\toprule
\textbf{Model} & \textbf{Params} & \textbf{GFLOPS} & \textbf{Memory} & \textbf{Throughput} \\
\midrule
Swin3D & 50.3M & 144.8 & 2.9GB & 9.86it/s \\
FactFormer & 5.0M & - & 20.4GB & 0.82it/s \\
UNet$_\mathrm{GenCFD}$ & 100.0M & 5519.8 & 4.8GB & 2.03it/s \\
UNet$_\mathrm{ConvNext}$ & 9.2M & 167.8 & 2.4GB & 11.9it/s \\
TFNO & 75.6M & 69.7 & 5.1GB & 4.7it/s \\
AViT & 60.0M & 71.5 & 0.3GB & 188.3it/s \\
AFNO & 64.1M & 1058.2 & 0.4GB & 31.7it/s \\
\midrule
P3D-$S$ & 11.2M & 108.5 & 0.9GB & 35.2it/s \\
P3D-$B$ & 46.1M & 1165.3 & 2.1GB & 10.0it/s \\
P3D-$L$ & 181.2M & 4638.1 & 4.6GB & 3.7it/s \\
\bottomrule
\end{tabular}
\end{table}
Regarding the number of parameters, GFLOPs, memory and throughput, the three configurations $S$, $B$ and $L$ of P3D are well positioned compared to the baselines we chose. Timings were obtained on a 
RTX A5000 GPU with batch size $1$. Importantly, across all tasks, P3D outperforms the baselines in terms of accuracy. We fixed the training setup for all architectures, using learning rate/optimizer/batch size that are common for training large transformer or UNet models. Due to the number and scale of the different models, we do not perform any hyperparameter tuning for individual models.

\clearpage

\section{Experiment 1: Jointly Learning PDEs} \label{app:datasets}

The datasets used in this experiment were carefully selected for a diverse list of partial differential equations (PDEs) with a focus on high spatial resolution. This is combined with variations of the physical parameters of the PDEs and different initial conditions, creating diverse dynamics across the different simulations for each type of PDE.
\\ 
\\
The dataset encompasses linear, reaction-diffusion, and nonlinear PDEs. We utilized the \textit{Exponax} solver, as detailed by Koehler et al. (2024) in the APEBench benchmark. The solver utilizes Exponential Time Differencing Runge-Kutta (ETDRK) methods. We intentionally opted not to use the APEBench dataset directly from the original authors. This decision was driven by our goal to create datasets with enhanced resolution and greater diversity in the underlying physical behaviors, rather than relying solely on variations in initial conditions as done in APEBench. 
It is a characteristic of the ETDRK methods that they operate within the Fourier domain. Consequently, their application is limited to scenarios with periodic domains and cannot accommodate complex boundary conditions. 

\begin{table*}[ht]
  \caption{Summary of datasets produced for the joint PDE learning task, 
  covering linear, reaction-diffusion, and nonlinear PDEs. The table details the dimensions of each dataset: number of simulations (s), time steps (t), fields/channels (f), and spatial dimensions (x, y, z). Beyond the explicitly varied quantities listed for each dataset, the initial conditions for every simulation (s) are also distinct.}
  \label{tab-app: dataset_overview_rephrased} 
  \centering
  \small
  \vspace{0.1cm}
  \begin{tabular}{l | c c c c c c c c}
 \toprule
 \textbf{Dataset} & s & t & f & x & y & z & Varied Quantities across s & Test Set \\
 \midrule
  \dHyp{} & 60 & 30 & 1 & 384 & 384 & 384 & hyper-diffusivity & $s\in[50,60[$\\
  \cmidrule{2-9}
  \dFisher{} & 60 & 30 & 1 & 384 & 384 & 384 & diffusivity, reactivity & $s\in[50,60[$ \\
  \dSH{} & 60 & 30 & 1 & 384 & 384 & 384 & reactivity, critical number & $s \in [50,60[$ \\
  \dGSalpha{} & 10 & 30 & 2 & 320 & 320 & 320 & initial conditions only & separate: s=3, t=100 \\
  \dGSbeta{} & 10 & 30 & 2 & 320 & 320 & 320 & initial conditions only & separate: s=3, t=100 \\
  \dGSgamma{} & 10 & 30 & 2 & 320 & 320 & 320 & initial conditions only & separate: s=3, t=100 \\
  \dGSdelta{} & 10 & 30 & 2 & 320 & 320 & 320 & initial conditions only & $s\in[8,10[$ \\
  \dGSepsilon{} & 10 & 30 & 2 & 320 & 320 & 320 & initial conditions only & separate: s=3, t=100 \\
  \dGStheta{} & 10 & 30 & 2 & 320 & 320 & 320 & initial conditions only & $s\in[8,10[$ \\
  \dGSiota{} & 10 & 30 & 2 & 320 & 320 & 320 & initial conditions only & $s\in[8,10[$\\
  \dGSkappa{} & 10 & 30 & 2 & 320 & 320 & 320 & initial conditions only & $s\in[8,10[$\\
  \cmidrule{2-9}
  \dBurgers{} & 60 & 30 & 2 & 384 & 384 & 384 & viscosity & $s\in[50,60[$\\
  \dKDV{} & 60 & 30 & 2 & 384 & 384 & 384 & domain extent, viscosity & $s\in[50,60[$\\
  \dKS{} & 60 & 30 & 1 & 384 & 384 & 384 & domain extent & separate: s=5, t=200\\
 \bottomrule
 \end{tabular}
\end{table*}

\subsection{Data Generation Setup}

A key aspect of the simulations, in addition to parameter variation, is the use of randomized initial conditions. The standard approach for constructing these conditions involves randomly selecting one of three initialization methods, each providing a unique spectral energy distribution. The first method, a random truncated Fourier series initializer, involves summing multiple Fourier series up to a cutoff frequency, chosen as a uniformly random integer between 2 and 10 (exclusive of 11). The second, the Gaussian random field initializer, produces a power-law spectrum in Fourier space, where energy diminishes polynomially with the wavenumber; its exponent is uniformly randomly selected from [2.3,3.6[. The third method, the diffused noise initializer, generates a tensor of values from normally distributed white noise, subsequently diffusing it. This results in a spectrum that decays exponentially quadratically, with an intensity rate uniformly random from [0.00005,0.01[. After generation, all initializers ensure the initial conditions' values are normalized to a maximum absolute value of one. For vector quantities, the randomly chosen initializer is applied independently to each component.

\subsection{PDE Types}

We make use of Exponential Time Differencing Runge-Kutta (ETDRK) methods to efficiently simulate different PDEs via Exponax.
While the chosen linear PDEs are simple and analytically solvable, the underlying dynamics are essential for more complicated PDEs. These linear PDEs can be understood as representing a scalar attribute, such as density. Unless stated otherwise, sampling from intervals is consistently performed using a uniform random distribution.
In the following, we describe a range of different three-dimensional PDE problems that are employed in our experiments ranging from linear, over reaction-diffusion, to non-linear PDEs. The class of non-linear PDEs is particularly challenging, as these cases more closely resemble real-world problems.

\paragraph{Hyper-Diffusion (\dHyp{})} behaves similarly to diffusion, where density dissipates inside a periodic domain due to the effects of hyper-diffusion. Unlike diffusion, hyper-diffusion does not treat all wavelengths equally, analogous to the relation between dispersion and advection. Higher frequency components are damped even more aggressively compared to normal diffusion, leading to visually stronger blur effect of the density field over time.

\begin{itemize}[itemsep=0pt]
    \item Dimensionality: $s=60$, $t=30$, $f=1$, $x=384$, $y=384$, $z=384$
    \item Initial Conditions: random truncated Fourier / Gaussian random field / diffused noise
    \item Boundary Conditions: periodic
    \item Time Step of Stored Data: 0.01
    \item Spatial Domain Size of Simulation: $[0, 1] \times [0, 1]$
    \item Fields: density
    \item Varied Parameters: hyper-diffusivity $\in [0.00005, 0.0005[$
    \item Validation Set: random $15\%$ split of all sequences from $s \in [0,50[$
    \item Test Set: all sequences from $s \in [50,60[$ 
\end{itemize}

\begin{figure*}[ht]
    \centering
    \includegraphics[width=0.99\textwidth]{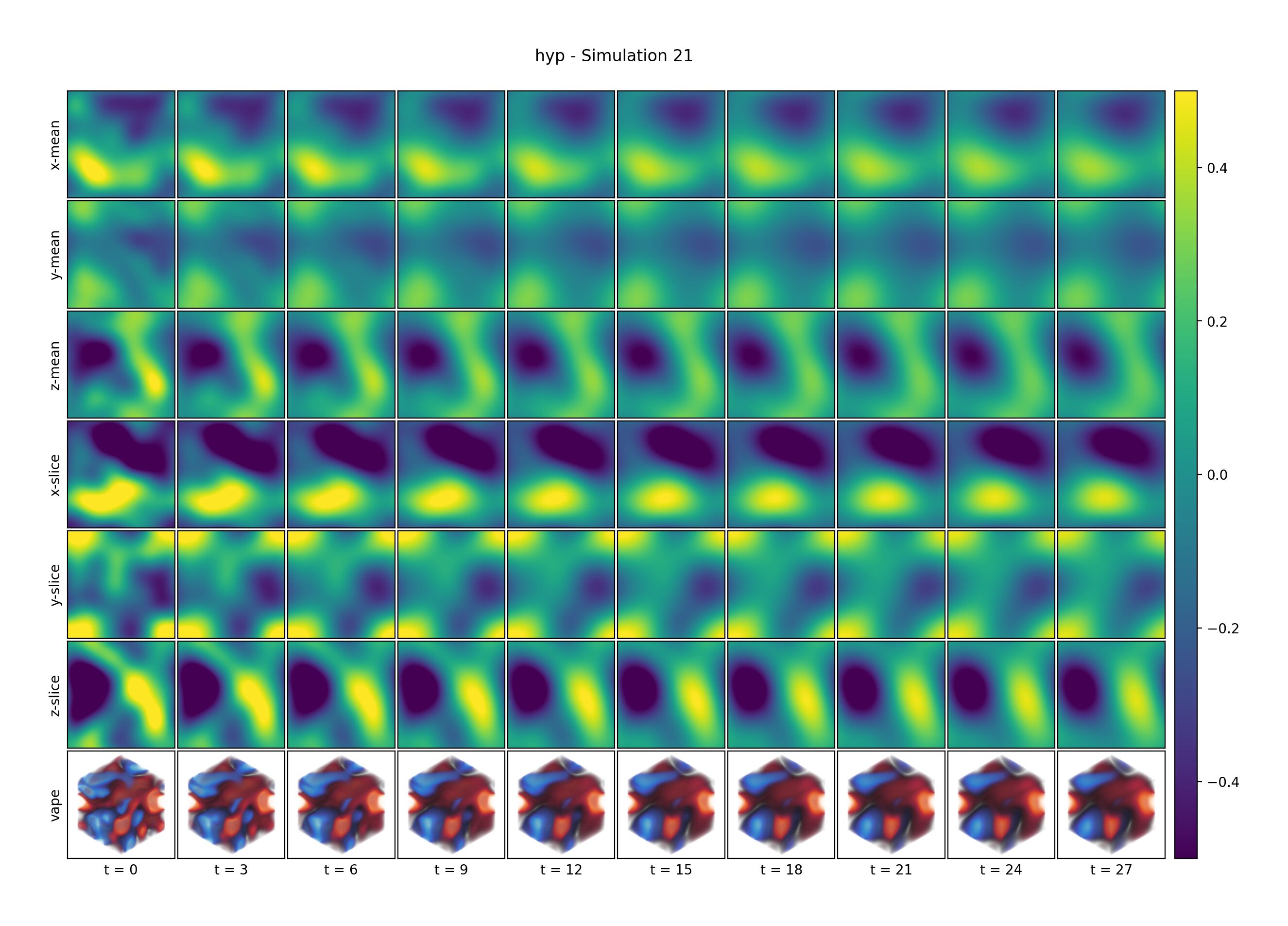}
    \caption{Random example simulation from \dHyp{}.}
    \label{fig-app: dataset linear hyp}
\end{figure*}

\clearpage

\begin{figure*}[ht]
    \centering
    \includegraphics[width=0.99\textwidth]{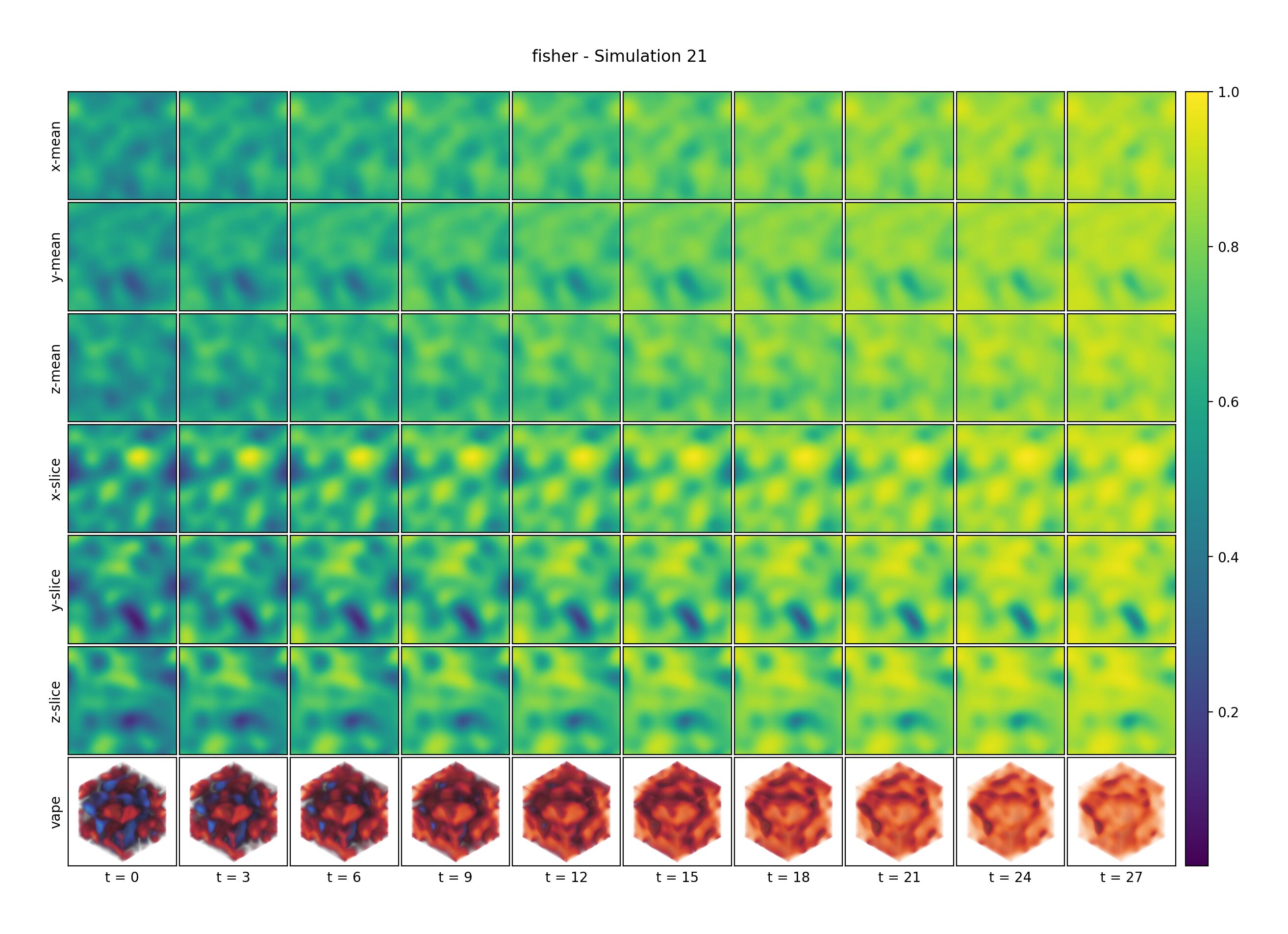}
    \caption{Random example simulation from \dFisher{}.}
    \label{fig-app: dataset reaction fisher}
\end{figure*}

\paragraph{Fisher-KPP (\dFisher{})} 
is a foundational reaction-diffusion PDE. These systems are used to model biological or geological processes, often resulting in pattern formation. This equation details how the concentration of a substance varies over time and space, influenced by a reaction process governed by a reactivity parameter, and its dispersal through diffusion, which is defined by a diffusivity parameter. Its applications extend to various domains, including wave propagation, population dynamics, ecology, and plasma physics. \Cref{fig-app: dataset reaction fisher} shows example visualizations from \dFisher{}.

\begin{itemize}[itemsep=0pt]
    \item Dimensionality: $s=60$, $t=30$, $f=1$, $x=384$, $y=384$, $z=384$
    \item Initial Conditions: random truncated Fourier / Gaussian random field / diffused noise (with clamping to $[0,1]$)
    \item Boundary Conditions: periodic
    \item Time Step of Stored Data: 0.005
    \item Spatial Domain Size of Simulation: $[0, 1] \times [0, 1]$
    \item Fields: concentration
    \item Varied Parameters: diffusivity $\in [0.0001, 0.02[$ and reactivity $\in [5,15[$
    \item Validation Set: random $15\%$ split of all sequences from $s \in [0,50[$
    \item Test Set: all sequences from $s \in [50,60[$ 
\end{itemize}

\clearpage

\begin{figure*}[ht]
    \centering
    \includegraphics[width=0.99\textwidth]{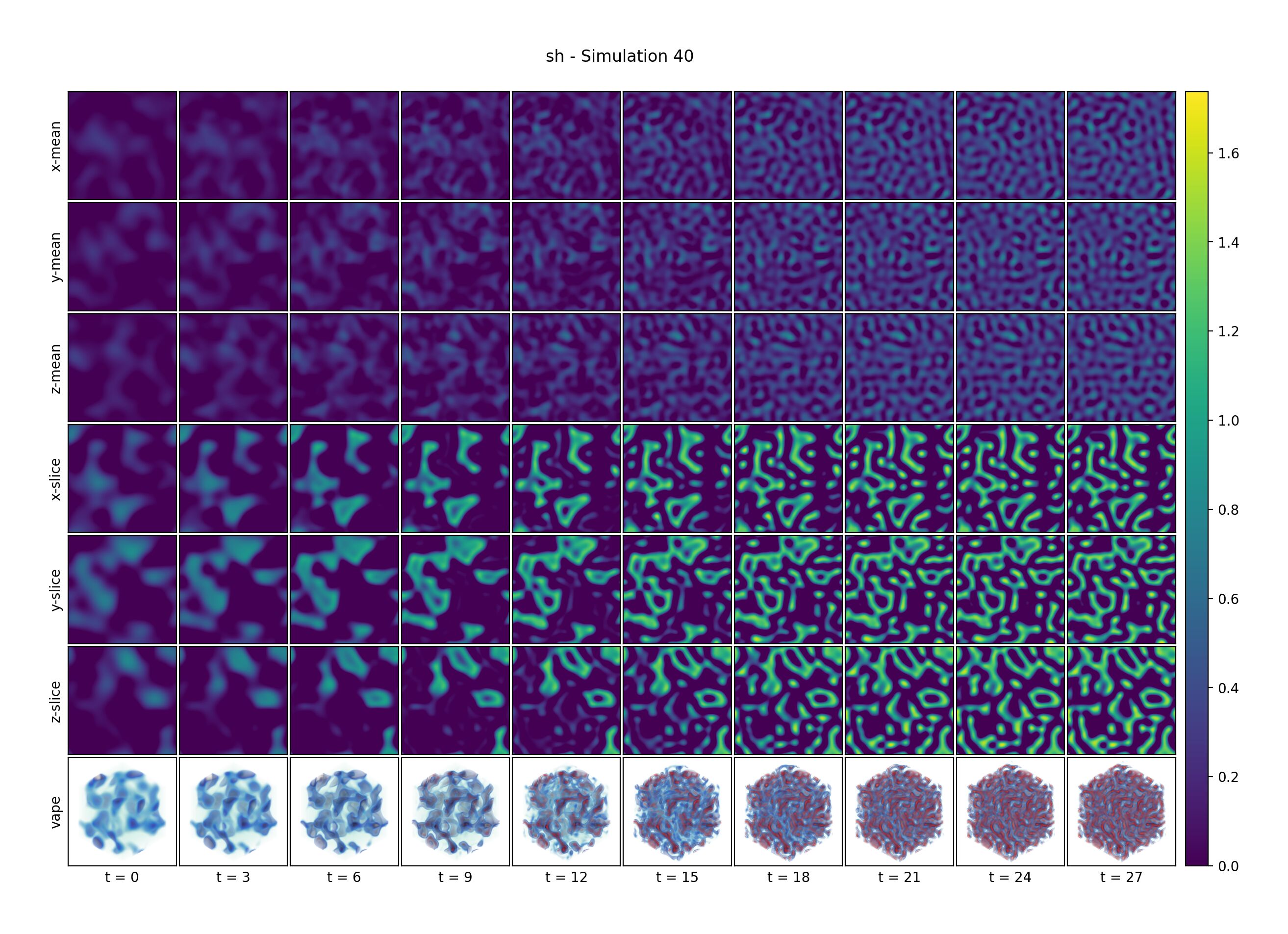}
    \caption{Random example simulation from \dSH{}.}
    \label{fig-app: dataset reaction sh}
\end{figure*}

\paragraph{Swift-Hohenberg (\dSH{})} is known for depicting various pattern formation processes. This equation can be applied to illustrate the structure of wrinkles in curved elastic bilayer materials. A prime example is the formation of human fingerprints, where tensions between skin layers generate their unique wrinkling. \Cref{fig-app: dataset reaction sh} shows example visualizations from \dSH{}.

\begin{itemize}[itemsep=0pt]
    \item Dimensionality: $s=60$, $t=30$, $f=1$, $x=384$, $y=384$, $z=384$
    \item Initial Conditions: random truncated Fourier / Gaussian random field / diffused noise
    \item Boundary Conditions: periodic
    \item Time Step of Stored Data: 0.5 (with 5 substeps for the simulation)
    \item Spatial Domain Size of Simulation: $[0, 20\pi] \times [0, 20\pi]$
    \item Fields: concentration
    \item Varied Parameters: reactivity $\in [0.4,1[$ and critical number $\in [0.8,1.2[$
    \item Validation Set: random $15\%$ split of all sequences from $s \in [0,50[$
    \item Test Set: all sequences from $s \in [50,60[$ 
\end{itemize}

\clearpage

\begin{figure*}[ht]
    \centering
    \includegraphics[width=0.85\textwidth]{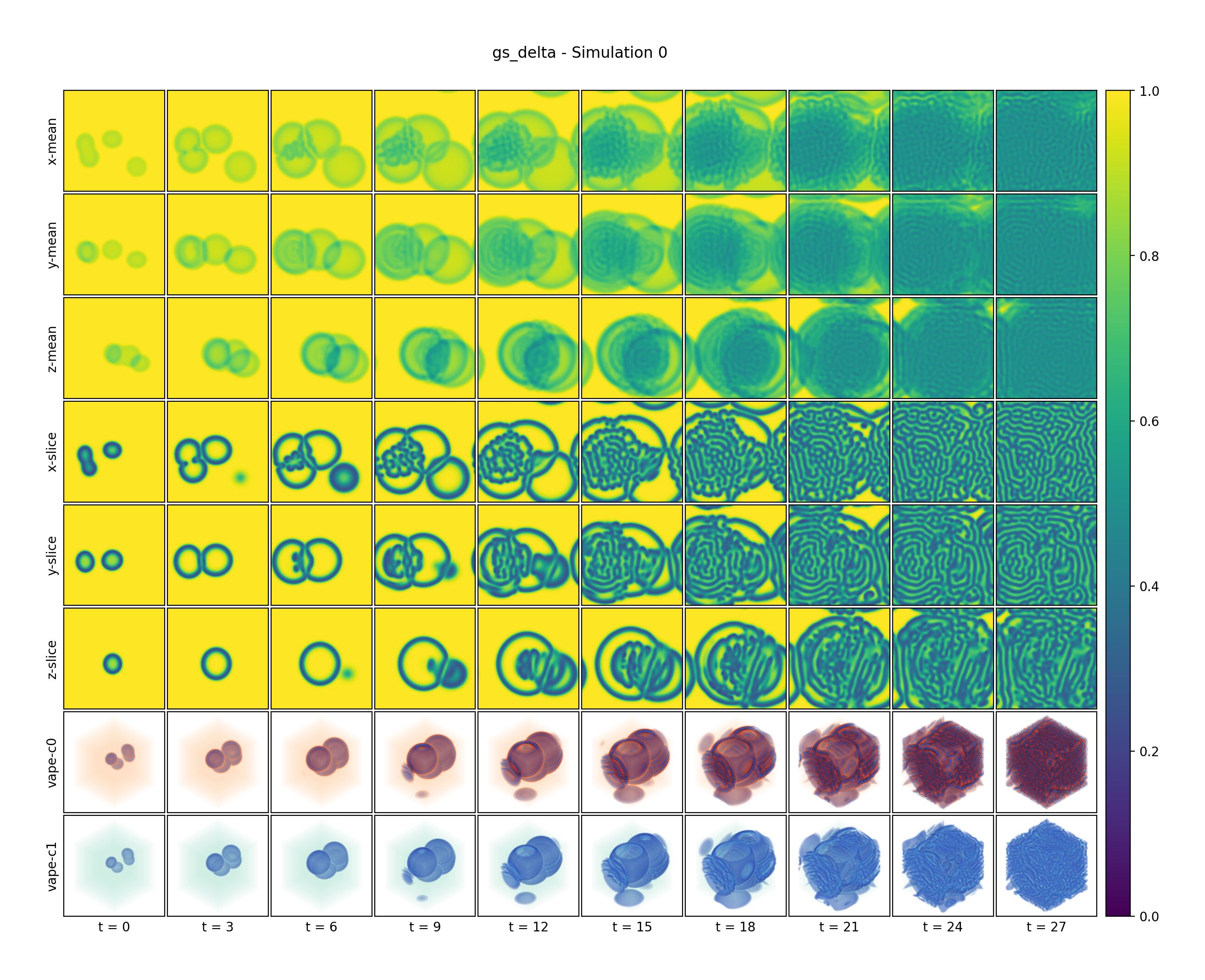}
    \includegraphics[width=0.85\textwidth]{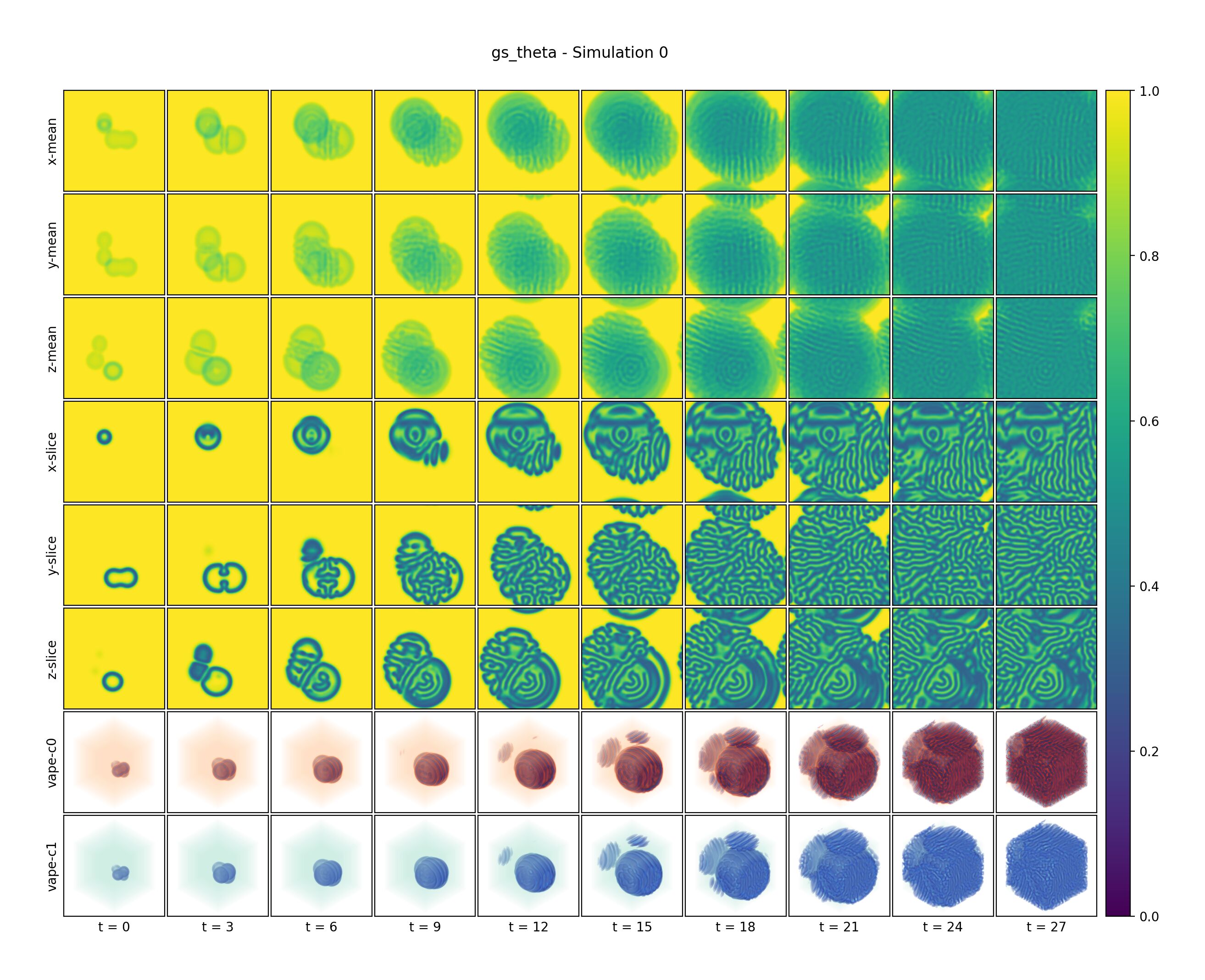}
    \caption{Random example simulations from steady configurations of the Gray-Scott model of a reaction-diffusion system: \dGSdelta{}, \dGStheta{}.} 
    \label{fig-app: dataset gs steady}
\end{figure*}

\begin{figure*}[ht]
    \centering
    \includegraphics[width=0.85\textwidth]{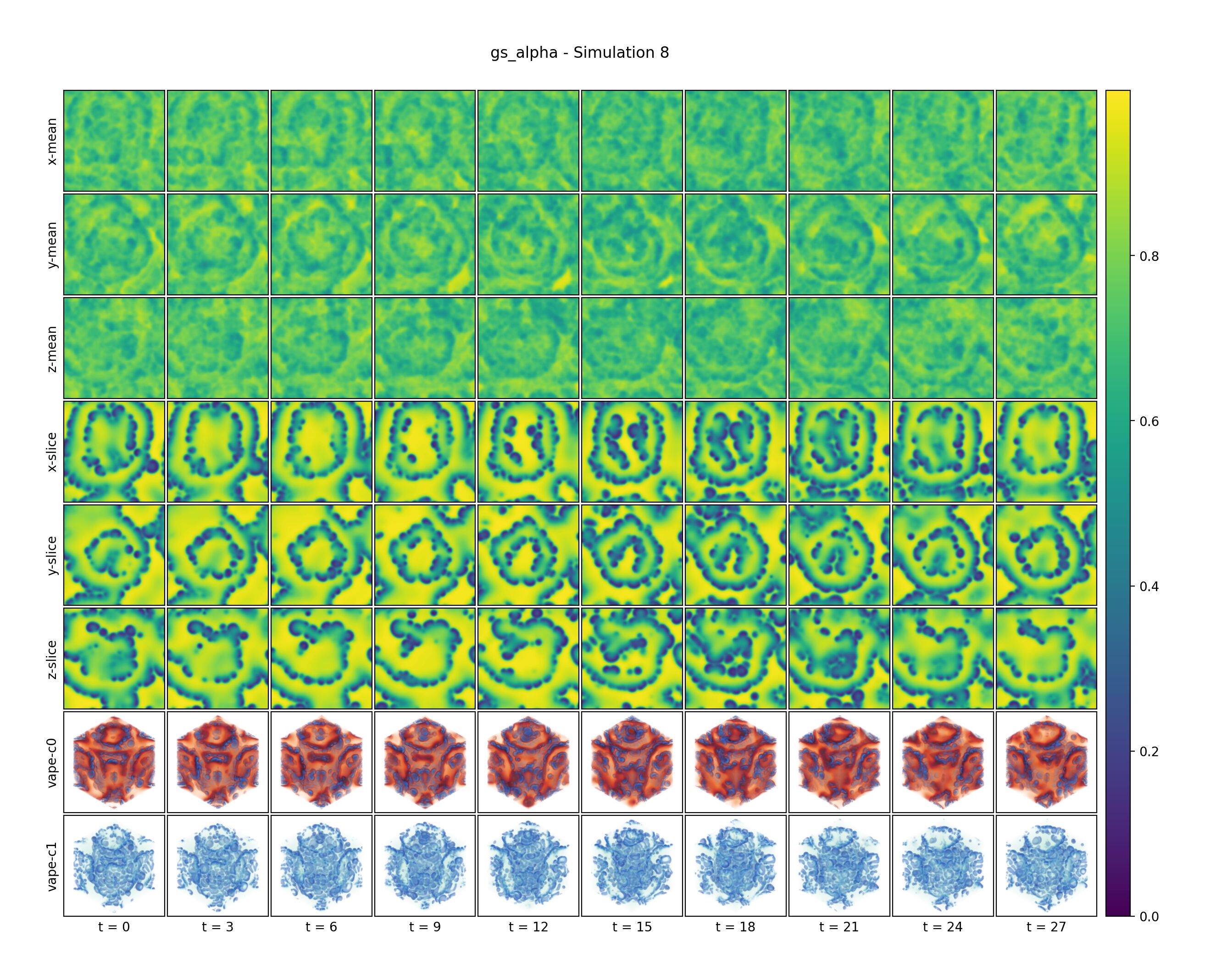}
    \includegraphics[width=0.85\textwidth]{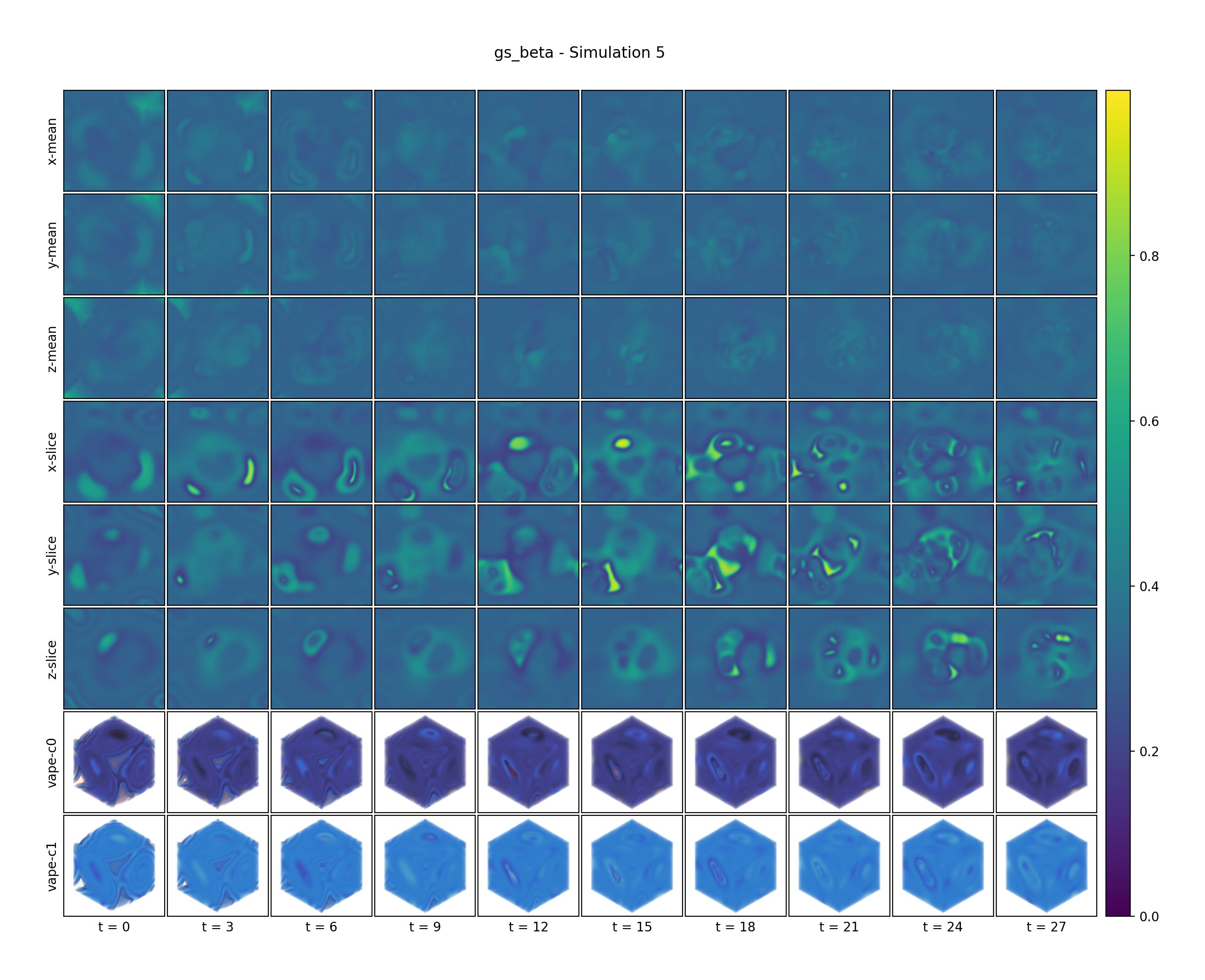}
    \caption{Random example simulations from unsteady configurations of the Gray-Scott model of a reaction-diffusion system: \dGSalpha{}, \dGSbeta{}. }
    \label{fig-app: dataset gs unsteady}
\end{figure*}

\paragraph{Gray-Scott (\dGS{})} illustrates the dynamic interplay of two reacting and diffusing chemical substances. Substance $s_a$ with concentration $c_a$ is depleted through reaction but resupplied based on a defined feed rate. Meanwhile, substance $s_b$, the reaction's product with concentration $c 
_b$, is eliminated from the domain at a given kill rate. The balance between these two rates profoundly influences the simulation outcomes, leading to diverse stable or evolving patterns. 
We simulate several cases: four with temporally steady configurations, which result in a state that does not substantially change anymore (\dGSdelta{}, \dGStheta{}, \dGSiota{}, and \dGSkappa{}), and four temporally unsteady configurations, which continuously evolve over time (\dGSalpha{}, \dGSbeta{}, \dGSgamma{}, and \dGSepsilon{}). For the unsteady case, separate test sets with longer temporal rollouts are created. \Cref{fig-app: dataset gs steady} shows example visualizations from the steady configurations, and \cref{fig-app: dataset gs unsteady} from the unsteady configurations and corresponding test sets. For further details, see \cite{pearson1993_Complex}.
\\
\\
For all simulations, the diffusivity of the substances is fixed to $d_a=0.00002$ and $d_b=0.00001$. In addition to that, datasets are initialized with a Gaussian blob initializer. The initializer creates four Gaussian blobs at random positions and variances in the center 60\% (20\% for \dGSkappa{}) of the domain, where the initialization of $c_a$ is the complement of $c_b$, i.e. $c_a = 1 - c_b$.

\begin{description}
    \item[Steady Configurations (\dGSdelta{}, \dGStheta{}, \dGSiota{}, and \dGSkappa{}):] 
\end{description}
\vspace{-0.6cm}
\begin{itemize}[itemsep=0pt]
    \item Dimensionality: $s=10$, $t=30$, $f=2$, $x=320$, $y=320$, $z=320$ (per configuration)
    \item Initial Conditions: random Gaussian blobs
    \item Boundary Conditions: periodic
    \item Time Step of Simulation: 1.0 (all configurations)
    \item Time Step of Stored Data:
    \begin{itemize}[itemsep=0pt]
        \item \dGSdelta{}: 130.0
        \item \dGStheta{}: 200.0
        \item \dGSiota{}: 240.0
        \item \dGSkappa{}: 300.0
    \end{itemize}
    \item Number of Warmup Steps (discarded, in time step of data storage): 
    \begin{itemize}[itemsep=0pt]
        \item \dGSdelta{}: 0
        \item \dGStheta{}: 0
        \item \dGSiota{}: 0
        \item \dGSkappa{}: 15
    \end{itemize}
    \item Spatial Domain Size of Simulation: $[0, 2.5] \times [0, 2.5]$
    \item Fields: concentration $c_a$, concentration $c_b$
    \item Varied Parameters: feed rate and kill rate determined by configuration (i.e., initial conditions only within configuration)
    \begin{itemize}[itemsep=0pt]
        \item \dGSdelta{}: feed rate: 0.028, kill rate: 0.056
        \item \dGStheta{}: feed rate: 0.040, kill rate: 0.060
        \item \dGSiota{}: feed rate: 0.050, kill rate: 0.0605
        \item \dGSkappa{}: feed rate: 0.052, kill rate: 0.063
    \end{itemize}
    \item Validation Set: random $15\%$ split of all sequences from $s \in [0,8[$
    \item Test Set: all sequences from $s \in [8,10[$ 
\end{itemize}

\begin{description}
    \item[Unsteady Configurations (\dGSalpha{}, \dGSbeta{}, \dGSgamma{}, and \dGSepsilon{}):]
\end{description}
\vspace{-0.6cm}
\begin{itemize}
    \item Dimensionality: $s=10$, $t=30$, $f=2$, $x=320$, $y=320$, $z=320$ (per configuration)
    \item Initial Conditions: random Gaussian blobs
    \item Boundary Conditions: periodic
    \item Time Step of Simulation: 1.0 (all configurations)
    \item Time Step of Stored Data: 
    \begin{itemize}[itemsep=0pt]
        \item \dGSalpha{}: 30.0
        \item \dGSbeta{}: 30.0
        \item \dGSgamma{}: 75.0
        \item \dGSepsilon{}: 15.0
    \end{itemize}
    
    \item Number of Warmup Steps (discarded, in time step of data storage): 
    \begin{itemize}
        \item \dGSalpha{}: 75
        \item \dGSbeta{}: 50
        \item \dGSgamma{}: 70
        \item \dGSepsilon{}: 300
    \end{itemize}

    \item Spatial Domain Size of Simulation: $[0, 2.5] \times [0, 2.5]$
    \item Fields: concentration $c_a$, concentration $c_b$
    \item Varied Parameters: feed rate and kill rate determined by configuration (i.e., initial conditions only within configuration)
    \begin{itemize}
        \item \dGSalpha{}: feed rate: 0.008, kill rate: 0.046
        \item \dGSbeta{}: feed rate: 0.020, kill rate: 0.046
        \item \dGSgamma{}: feed rate: 0.024, kill rate: 0.056
        \item \dGSepsilon{}: feed rate: 0.020, kill rate: 0.056
    \end{itemize}
    
    \item Validation Set: random $15\%$ split of all sequences from $s \in [0,10[$
    \item Test Set: separate simulations with $s=30$, $t=100$, $f=2$, $x=320$, $y=320$, $z=320$ (per configuration)
\end{itemize}

\clearpage

\begin{figure*}[ht]
    \centering
    \includegraphics[width=0.99\textwidth]{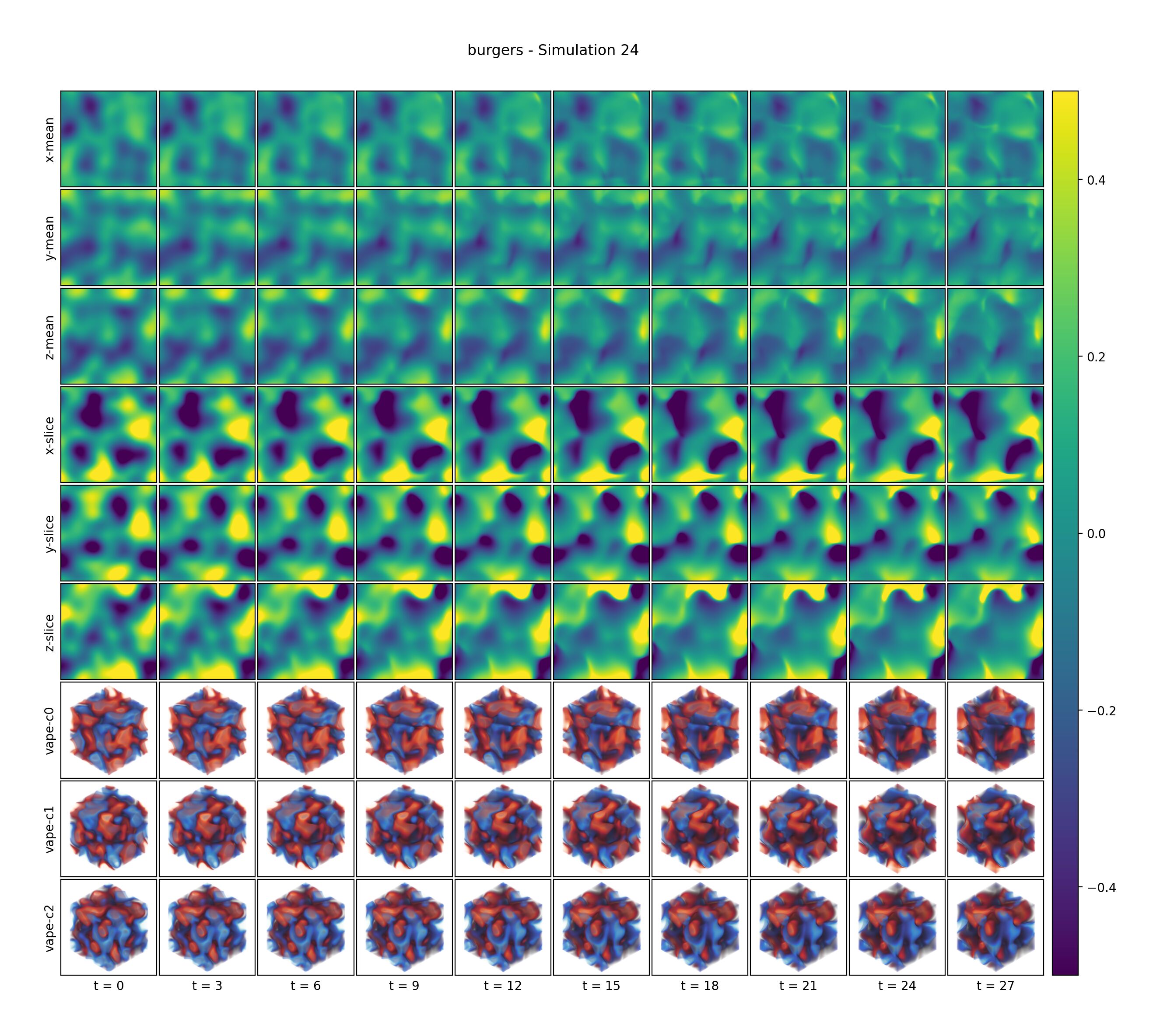}
    \caption{Random example simulation from \dBurgers{}.}
    \label{fig-app: dataset nonlinear burgers}
\end{figure*}

\paragraph{Burgers (\dBurgers{})}
bears resemblance to an advection-diffusion problem. Instead of modeling the transport of a scalar density, this equation describes how a flow field itself evolves due to the combined effects of advection and diffusion. This process can result in the formation of abrupt discontinuities, often referred to as shock waves, which present a significant challenge for accurate simulation. Burgers' equation also finds utility in fields such as nonlinear acoustics and the modeling of traffic flow.
\Cref{fig-app: dataset nonlinear burgers} shows example visualizations from \dBurgers{}.

\begin{itemize}[itemsep=0pt]
    \item Dimensionality: $s=60$, $t=30$, $f=2$, $x=384$, $y=384$, $z=384$
    \item Initial Conditions: random truncated Fourier / Gaussian random field / diffused noise
    \item Boundary Conditions: periodic
    \item Time Step of Stored Data: 0.01 (with 50 substeps for the simulation)
    \item Spatial Domain Size of Simulation: $[0, 1] \times [0, 1]$
    \item Fields: velocity ($x, y$)
    \item Varied Parameters: viscosity $\in [0.001,0.005[$
    \item Validation Set: random $15\%$ split of all sequences from $s \in [0,50[$
    \item Test Set: all sequences from $s \in [50,60[$ 
\end{itemize}

\clearpage

\begin{figure*}[ht]
    \centering
    \includegraphics[width=0.99\textwidth]{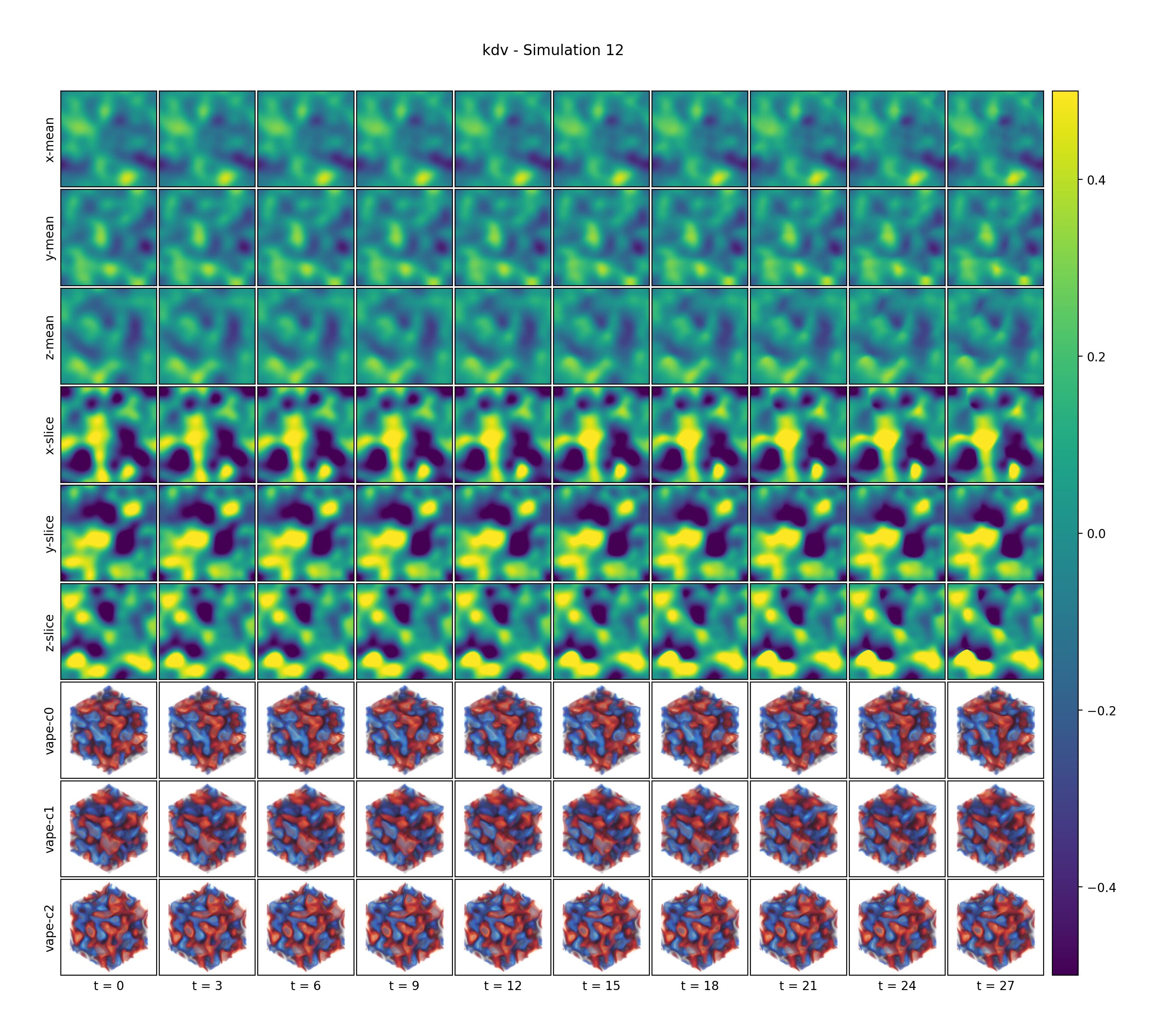}
    \caption{Random example simulation from \dKDV{}.}
    \label{fig-app: dataset nonlinear kdv}
\end{figure*}

\paragraph{Korteweg-de-Vries (\dKDV{})} 
presents simulations of the Korteweg-de-Vries equation within a periodic domain. This equation models dispersive, non-dissipative wave propagation and is a classic example of an integrable PDE. It poses a challenge because energy is transferred to high spatial frequencies, resulting in distinct, moving soliton waves that maintain their shape and propagation speed. Throughout these simulations, the convection coefficient remains constant at $-6$, and the dispersivity coefficient is consistently $1$.
\Cref{fig-app: dataset nonlinear kdv} shows example visualizations from \dKDV{}.

\begin{itemize}[itemsep=0pt]
    \item Dimensionality: $s=60$, $t=30$, $f=2$, $x=384$, $y=384$, $z=384$
    \item Initial Conditions: random truncated Fourier / Gaussian random field / diffused noise
    \item Boundary Conditions: periodic
    \item Time Step of Stored Data: 0.05 (with 10 substeps for the simulation)
    \item Spatial Domain Size of Simulation: varied per simulation
    \item Fields: velocity ($x, y$)
    \item Varied Parameters: domain extent $\in [30,120[$ identically for $x, y, z$, i.e. a square domain, and viscosity $\in [0.1,0.25[$ 
    \item Validation Set: random $15\%$ split of all sequences from $s \in [0,50[$
    \item Test Set: all sequences from $s \in [50,60[$ 
\end{itemize}

\clearpage

\begin{figure*}[ht]
    \centering
    \includegraphics[width=0.99\textwidth]{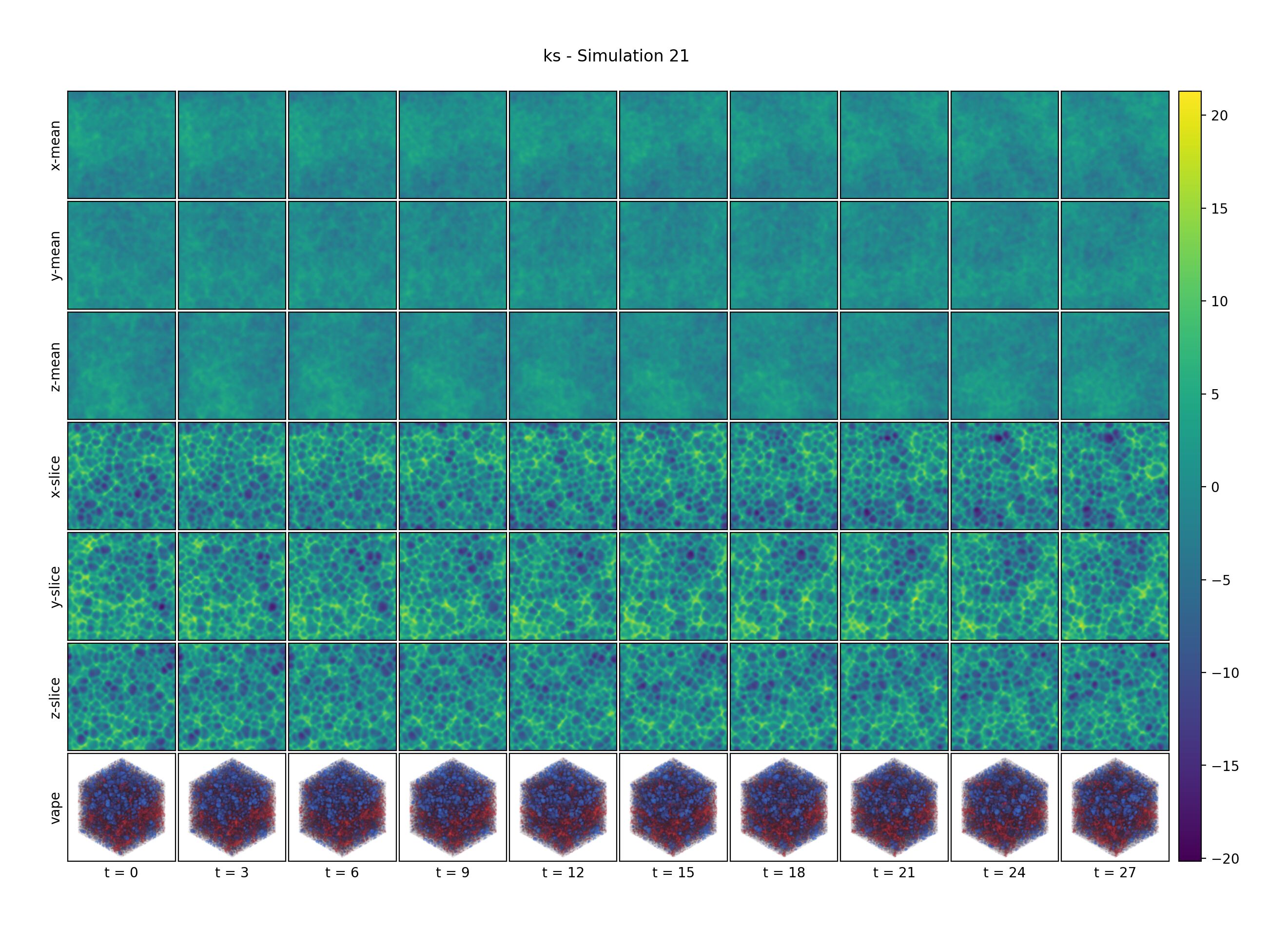}
    \caption{Random example simulation from \dKS{}.}
    \label{fig-app: dataset nonlinear ks}
\end{figure*}

\paragraph{Kuramoto-Sivashinsky (\dKS{})} 
models thermo-diffusive flame instabilities in combustion and also finds use in reaction-diffusion systems on a periodic domain. It's notable for its chaotic behavior, where even slightly different initial conditions can lead to wildly divergent temporal trajectories over time. 
The initial transient phase of the simulations is not included in the dataset. \cref{fig-app: dataset nonlinear ks} shows example visualizations from \dKS{}.

\begin{itemize}[itemsep=0pt]
    \item Dimensionality: $s=60$, $t=30$, $f=1$, $x=384$, $y=384$, $z=384$
    \item Initial Conditions: random truncated Fourier / Gaussian random field / diffused noise
    \item Boundary Conditions: periodic
    \item Time Step of Stored Data: 0.2 (with 2 substeps for the simulation)
    \item Number of Warmup Steps (discarded, in time step of data storage): 200
    \item Spatial Domain Size of Simulation: varied per simulation
    \item Fields: density
    \item Varied Parameters: domain extent $\in [10,130[$ identically for $x, y$, i.e. a square domain
    \item Validation Set: random $15\%$ split of all sequences from $s \in [0,600[$
    \item Test Set: separate simulations with $s=50$, $t=200$, $f=1$,  $x=384$, $y=384$, $z=384$
\end{itemize}

\clearpage

\subsection{Visualizations of Predictions}

\begin{figure}[h]
    \centering
    \includegraphics[width=0.49\linewidth]{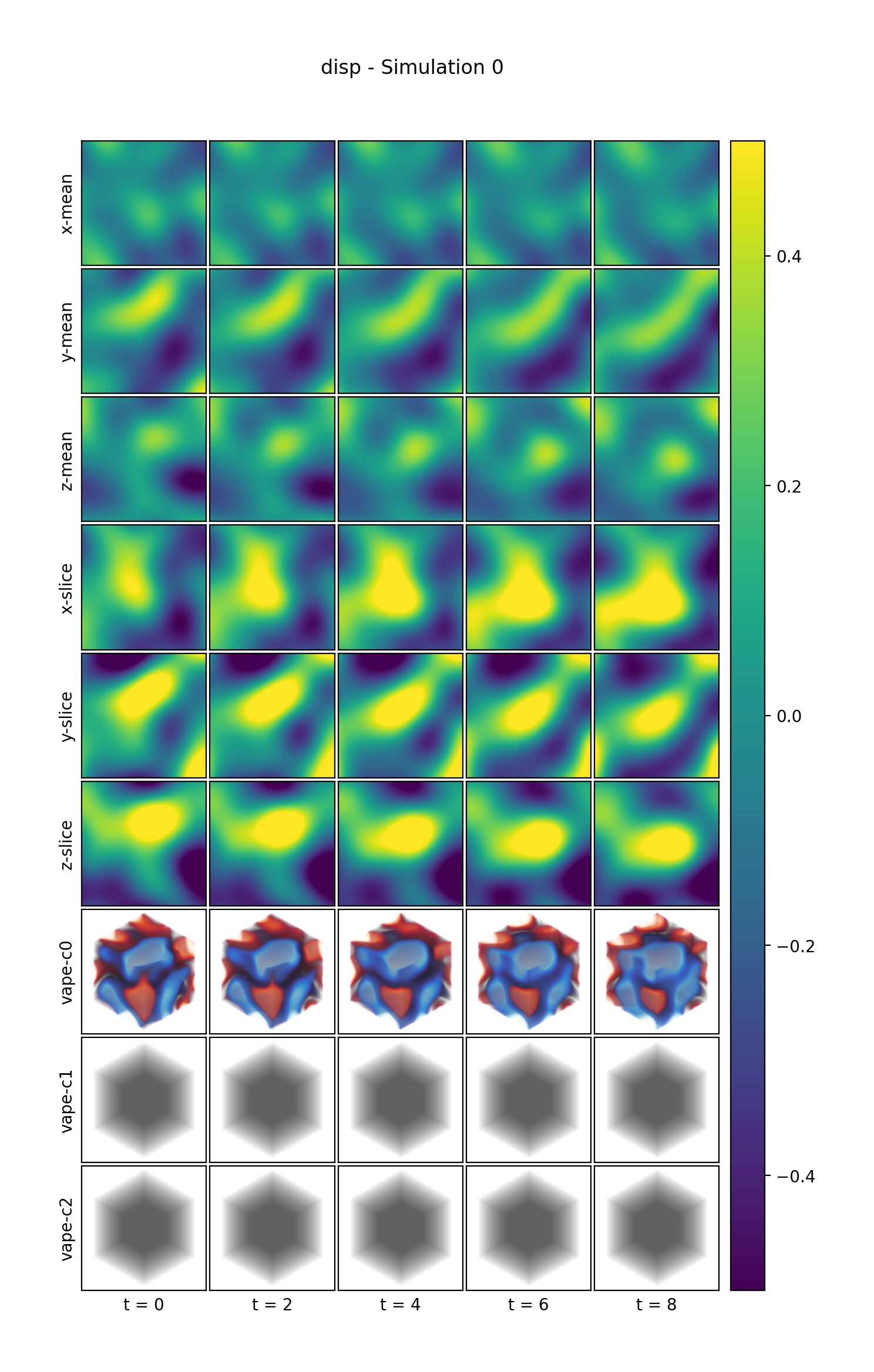}
    \includegraphics[width=0.49\linewidth]{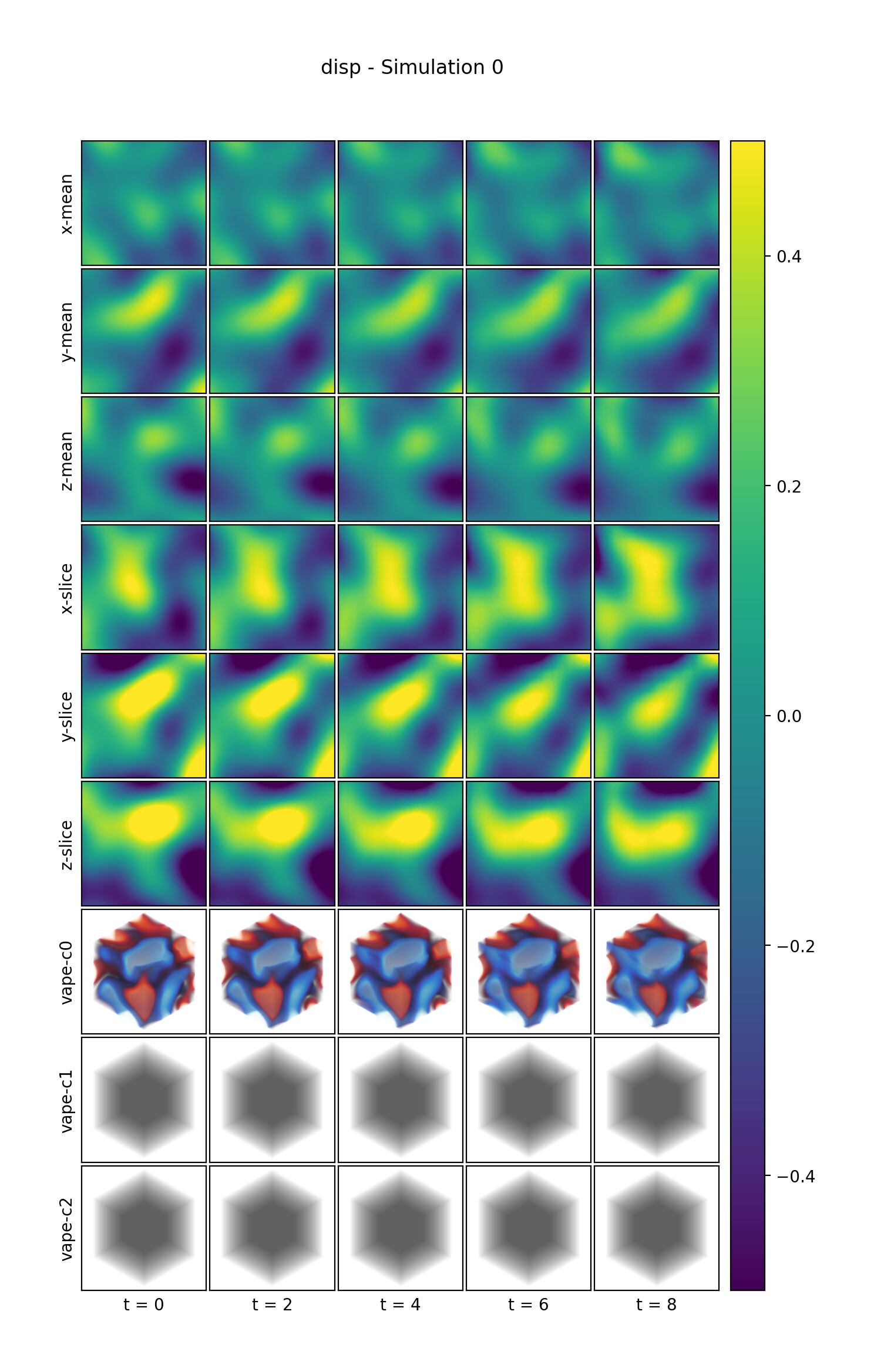}
    \caption{\dHyp{}. Reference (left) and autoregressive prediction for $t=8$ steps  with P3D-S <128|320> (right) on the test set at resolution $320^3$.}
    \label{fig:disp prediction}
\end{figure}

\begin{figure}[h]
    \centering
    \includegraphics[width=0.49\linewidth]{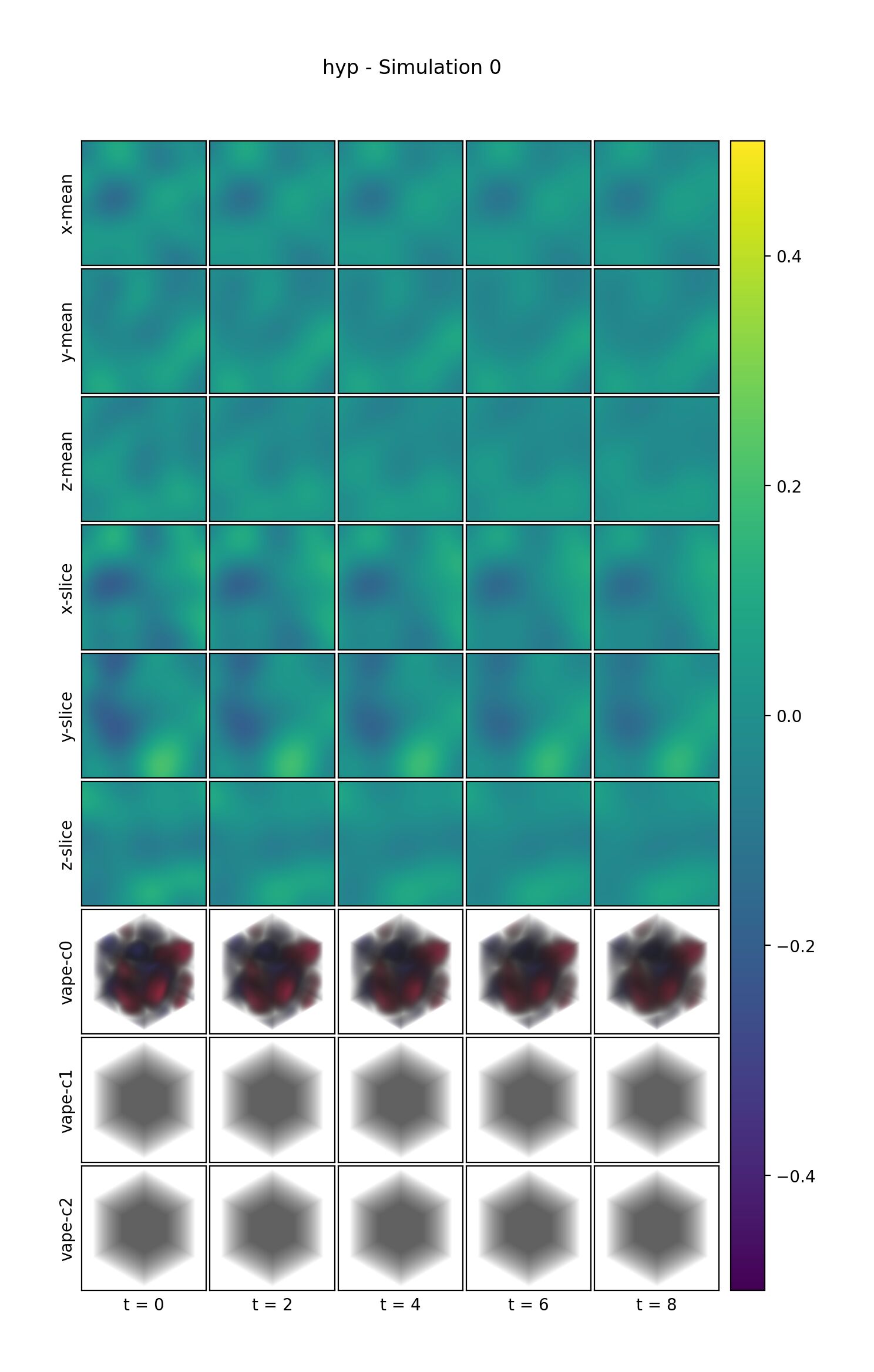}
    \includegraphics[width=0.49\linewidth]{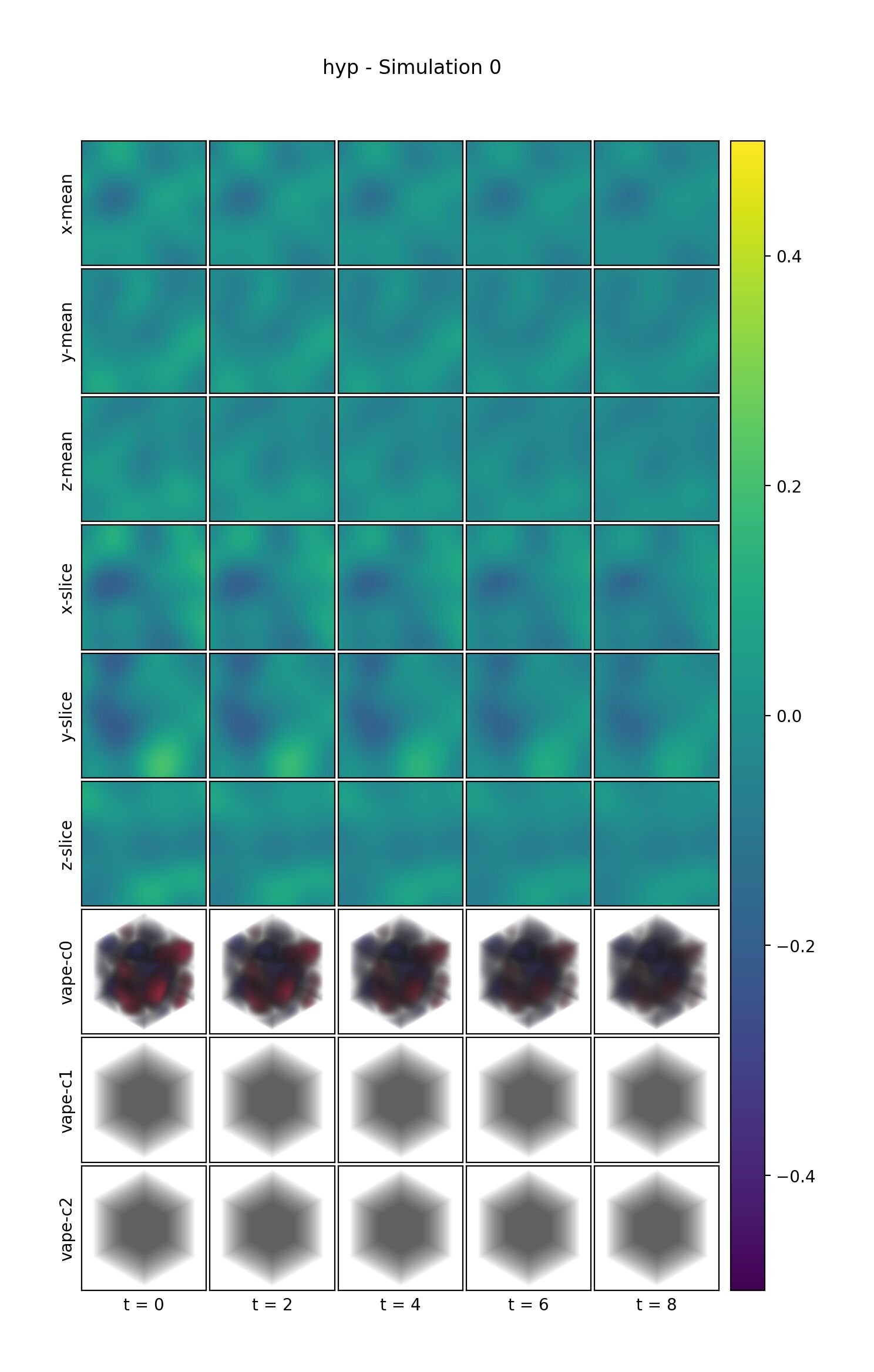}
    \caption{\dDisp{}. Reference (left) and autoregressive prediction for $t=8$ steps  with P3D-S (right) on the test set at resolution $320^3$.}
    \label{fig:hyp prediction}
\end{figure}

\begin{figure}[h]
    \centering
    \includegraphics[width=0.49\linewidth]{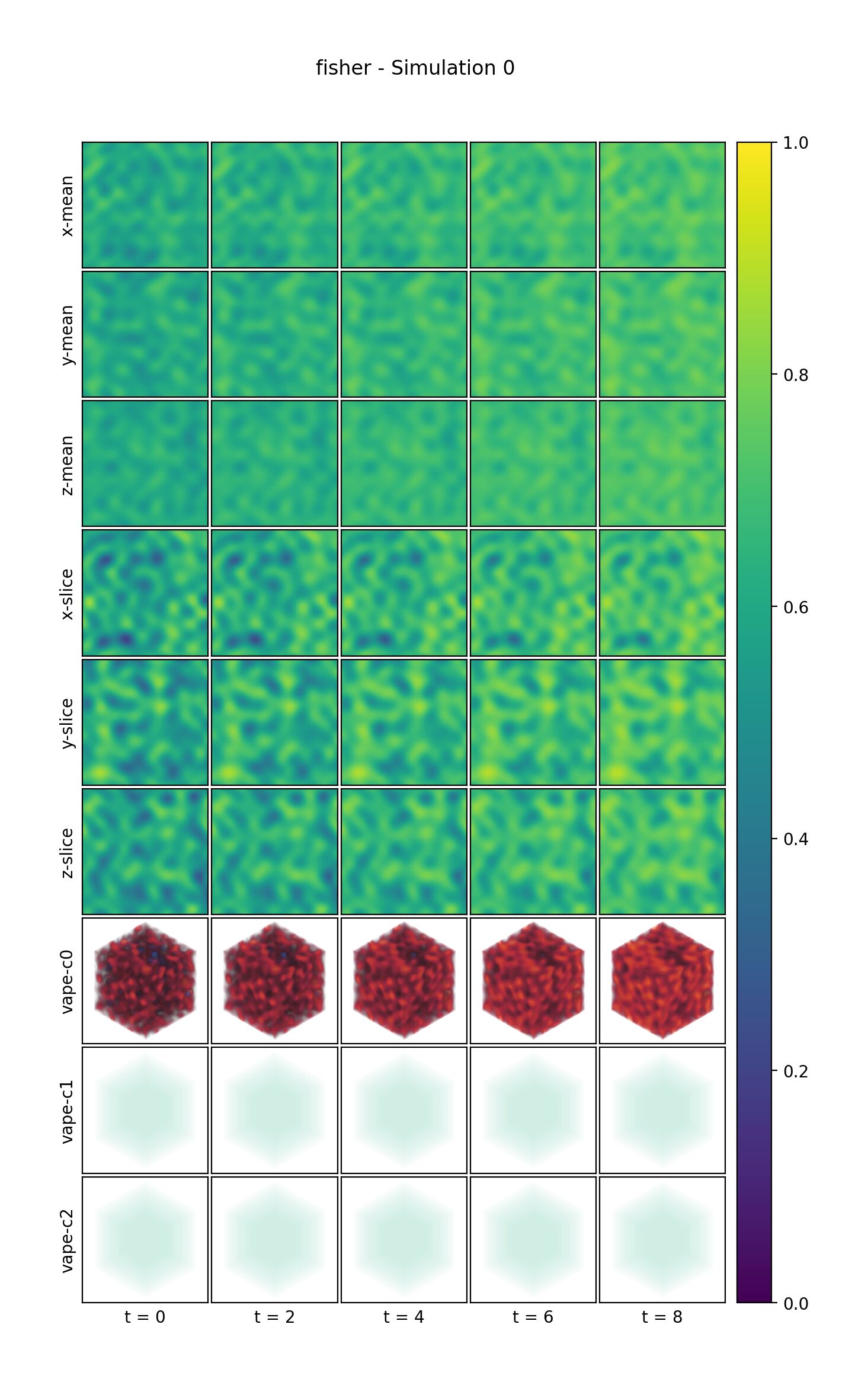}
    \includegraphics[width=0.49\linewidth]{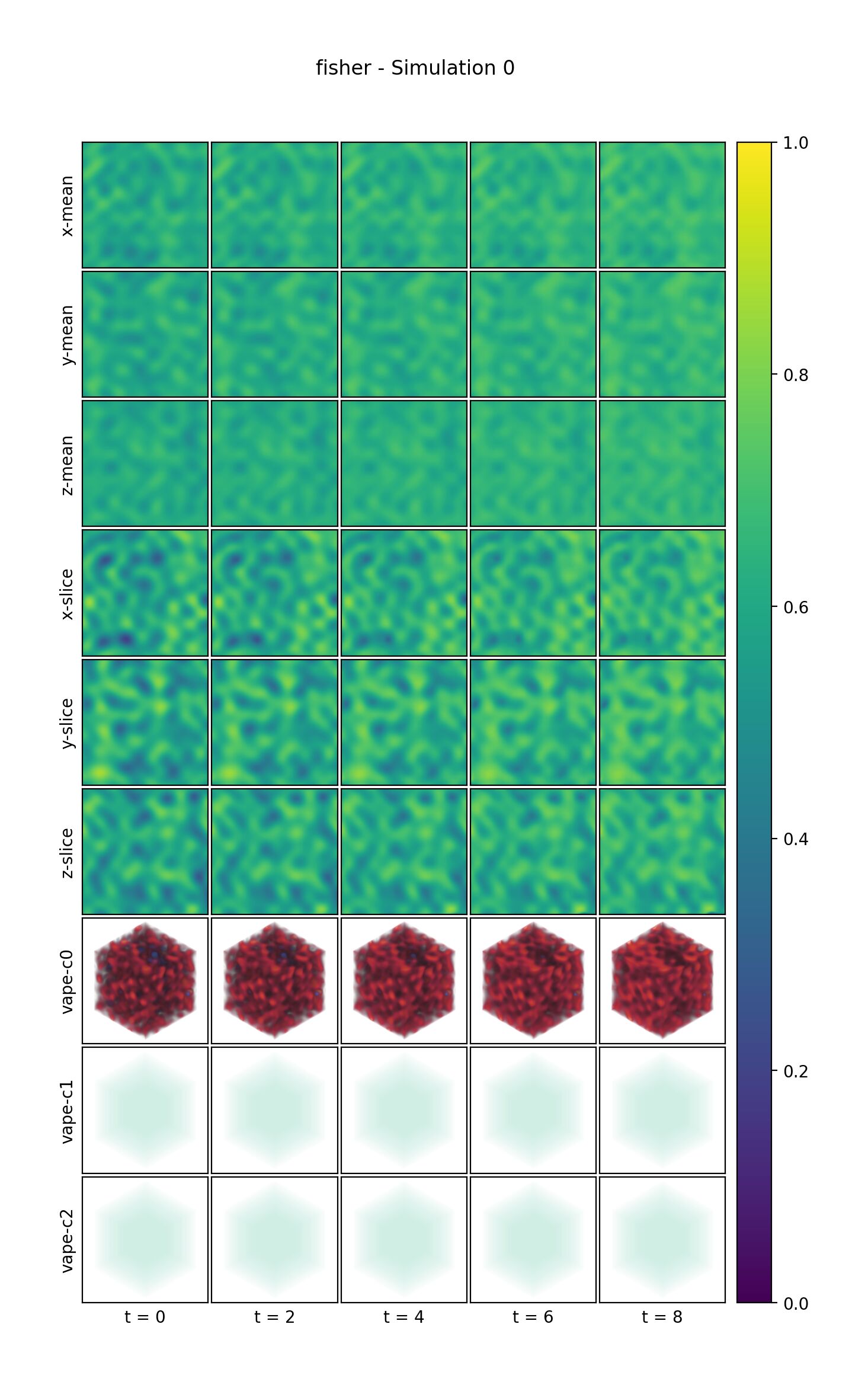}
    \caption{\dFisher{}. Reference (left) and autoregressive prediction for $t=8$ steps  with P3D-S <128|320> (right) on the test set at resolution $320^3$.}
    \label{fig:fisher prediction}
\end{figure}

\begin{figure}[h]
    \centering
    \includegraphics[width=0.49\linewidth]{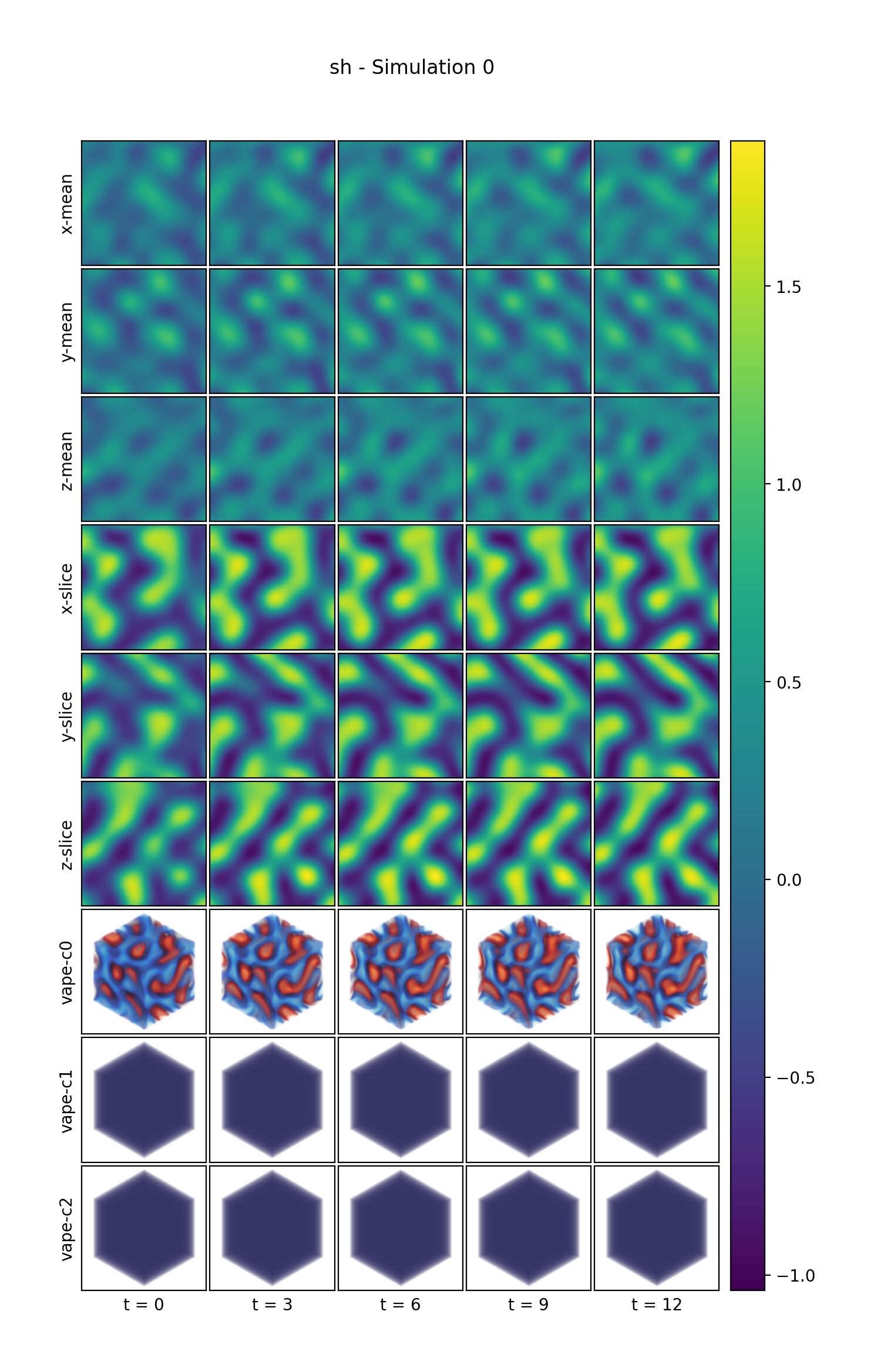}
    \includegraphics[width=0.49\linewidth]{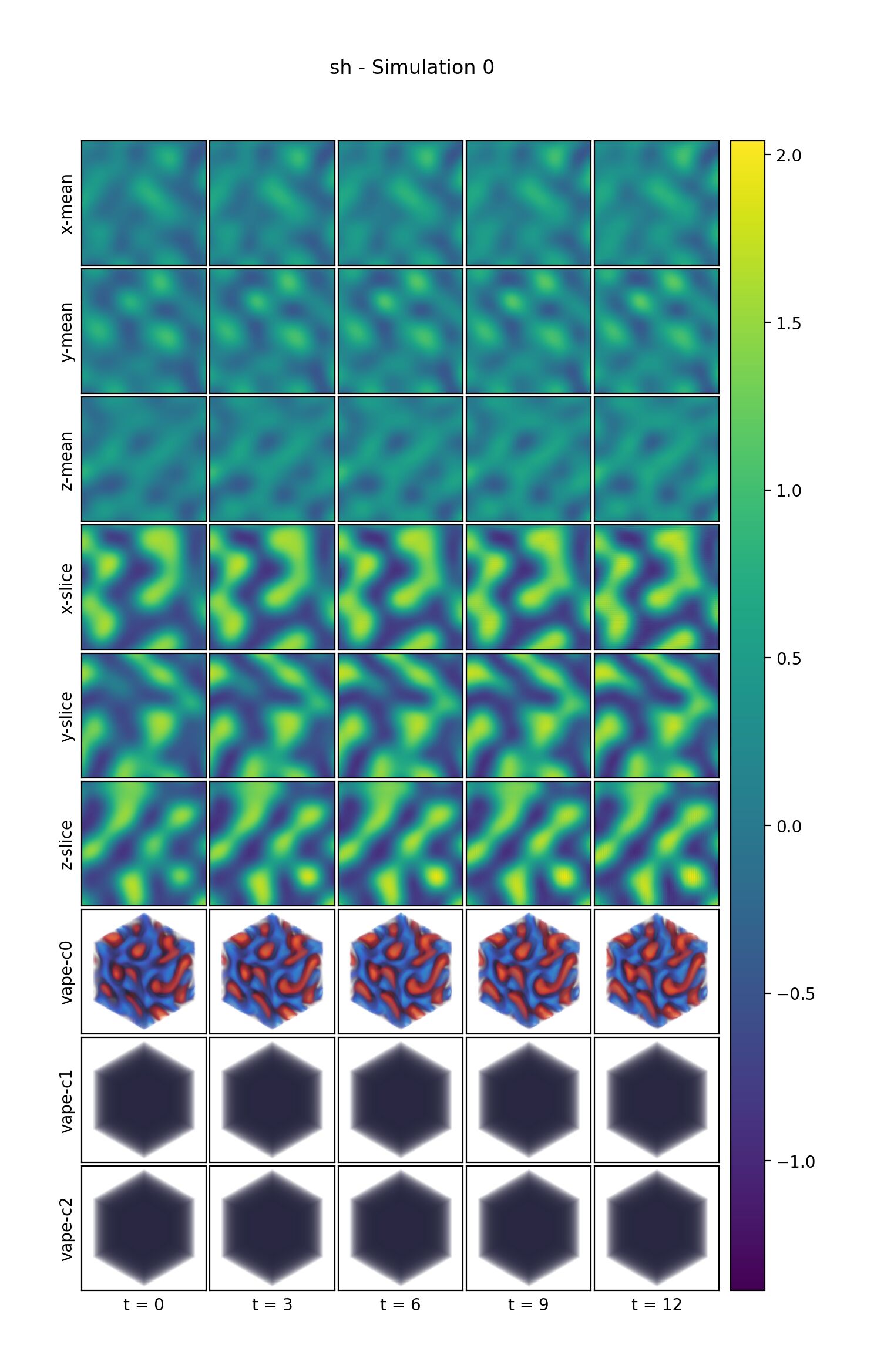}
    \caption{\dSH{}. Reference (left) and autoregressive prediction for $t=8$ steps  with P3D-S <128|320> (right) on the test set at resolution $320^3$.}
    \label{fig:sh prediction}
\end{figure}

\begin{figure}
    \centering
    \includegraphics[width=0.49\linewidth]{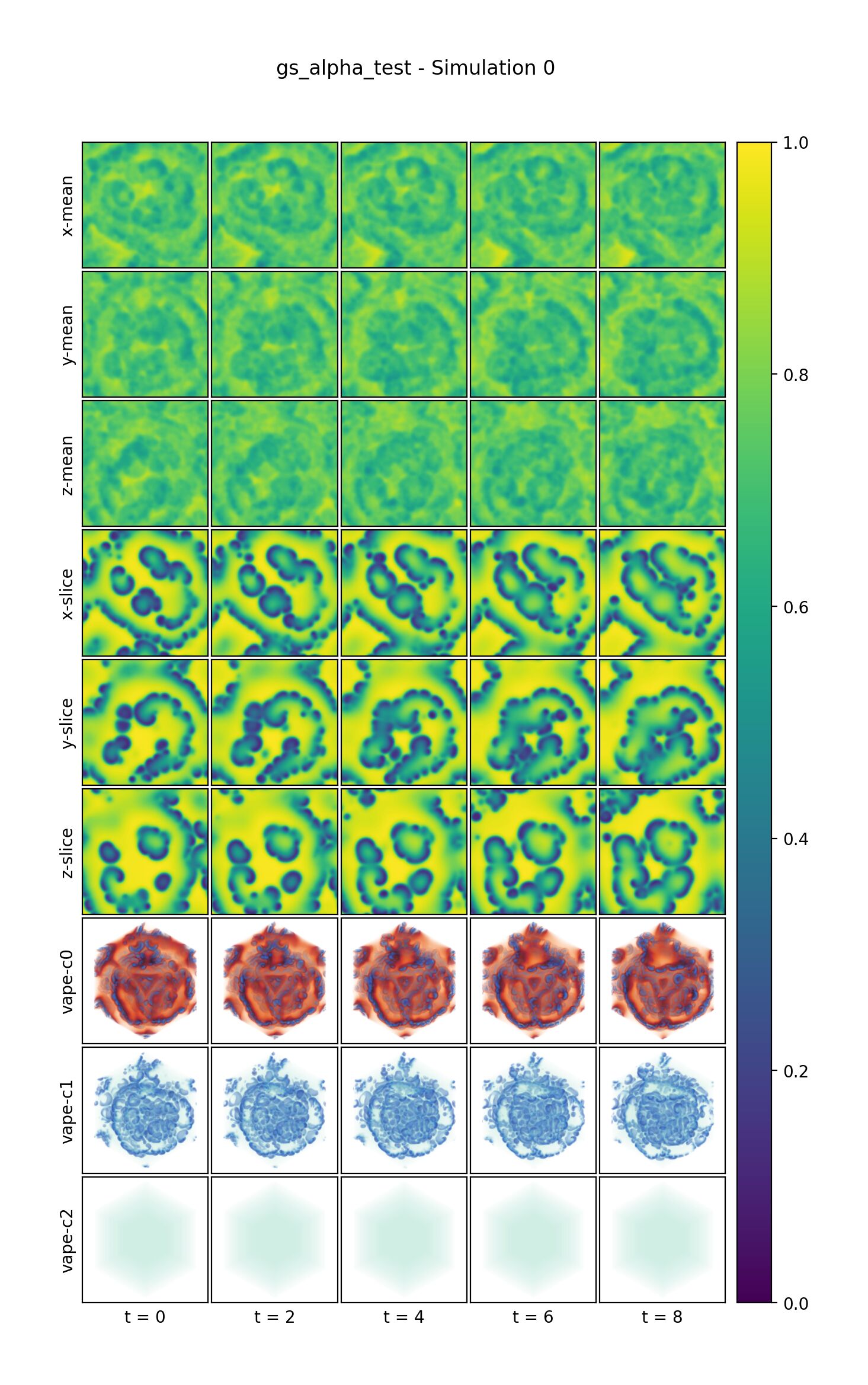}
    \includegraphics[width=0.49\linewidth]{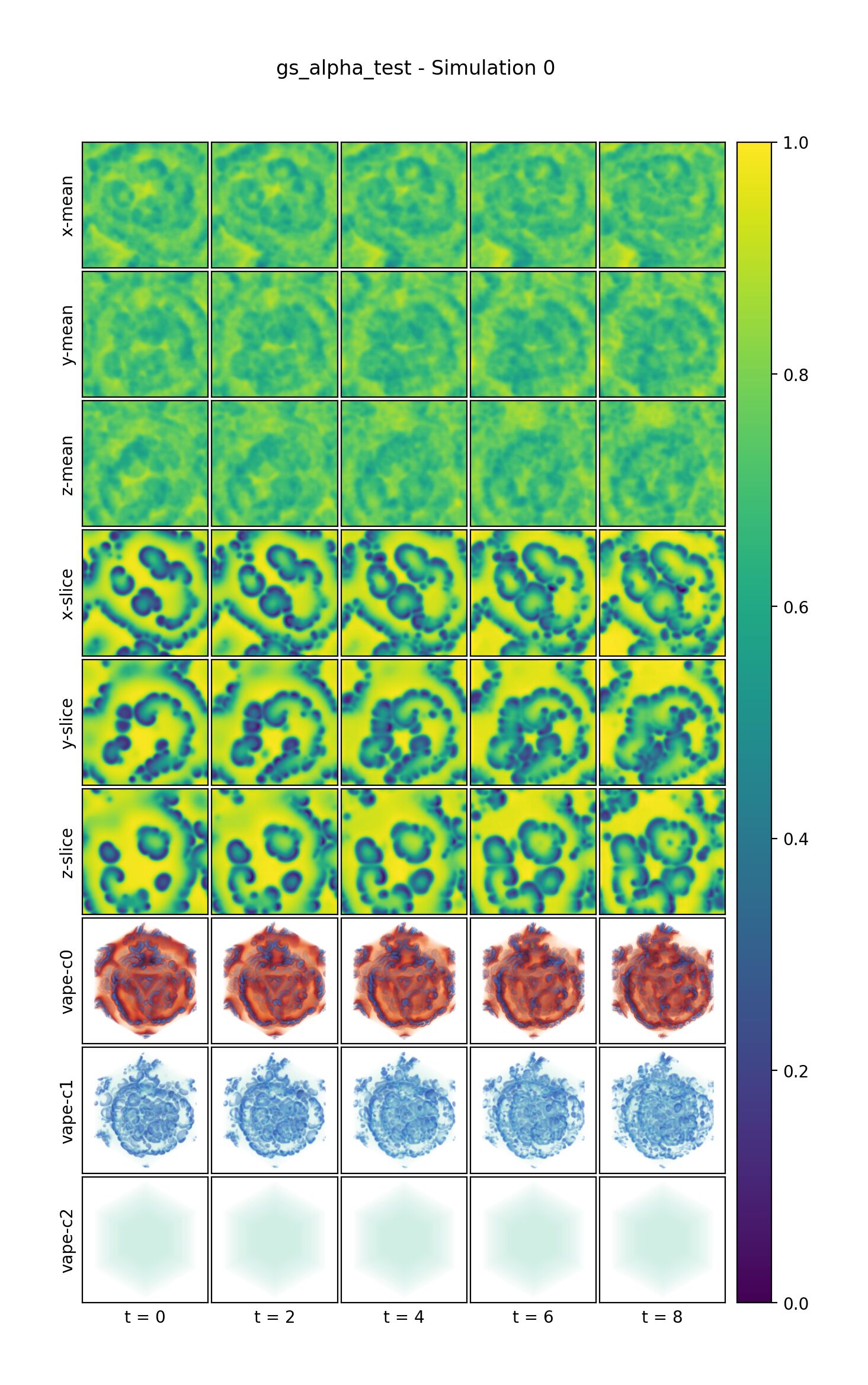}
    \caption{\dGSalpha{}. Reference (left) and autoregressive prediction for $t=8$ steps  with P3D-S <128|320> (right) on the test set at resolution $320^3$.}
    \label{fig:gs alpha prediction}
\end{figure}

\begin{figure}
    \centering
    \includegraphics[width=0.49\linewidth]{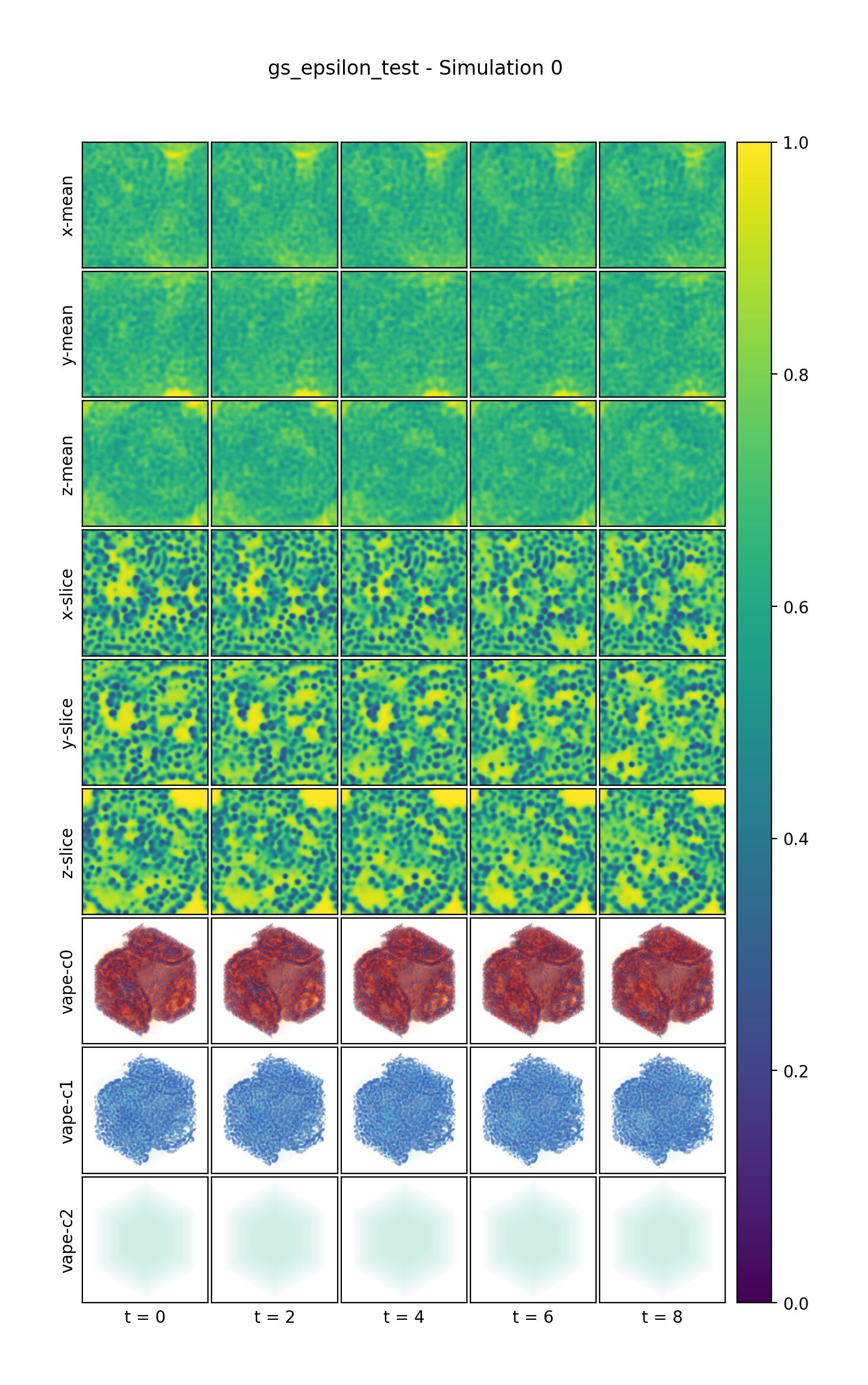}
    \includegraphics[width=0.49\linewidth]{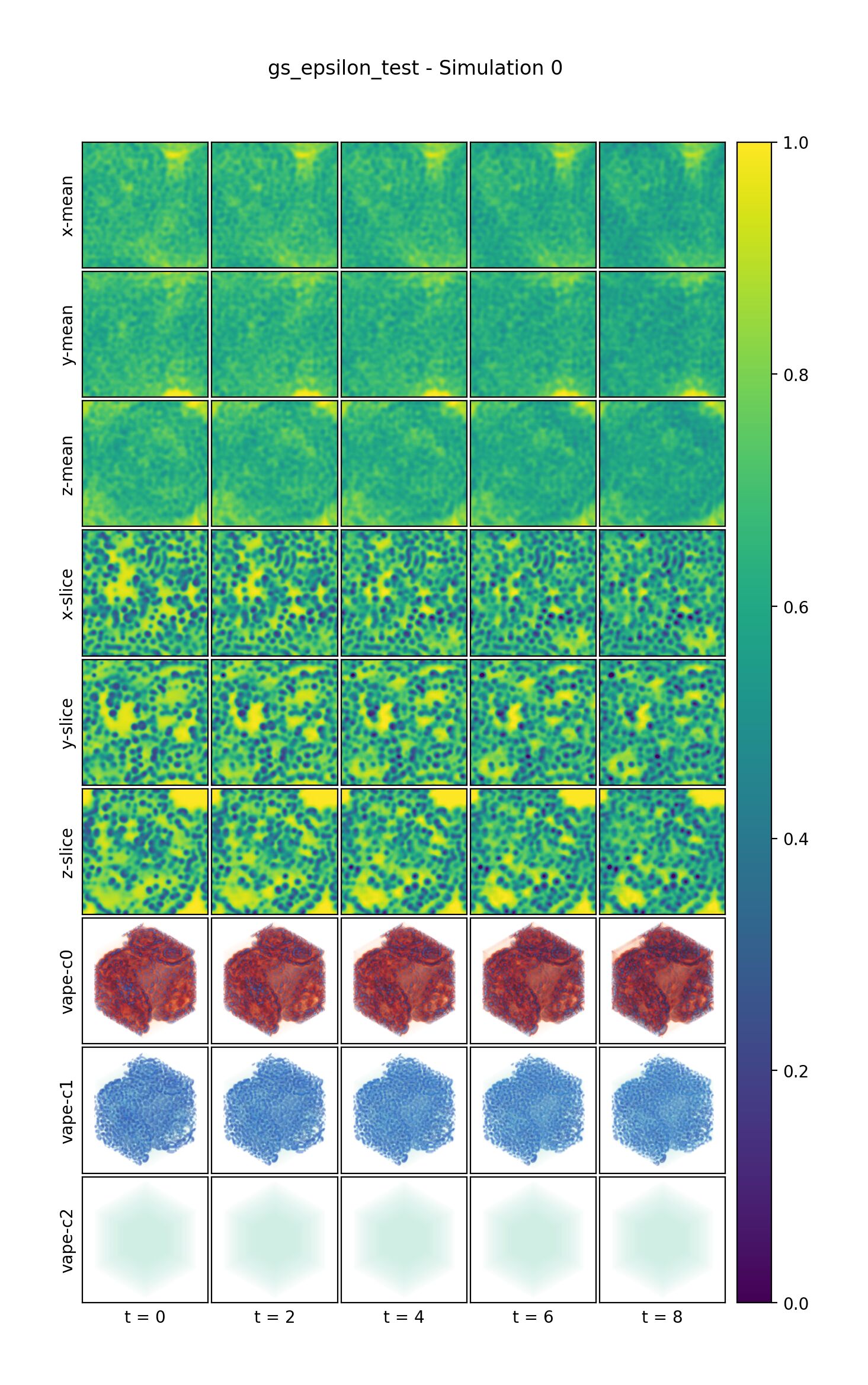}
    \caption{\dGSepsilon{}. Reference (left) and autoregressive prediction for $t=8$ steps  with P3D-S <128|320> (right) on the test set at resolution $320^3$.}
    \label{fig:gs epsilon prediction}
\end{figure}

\begin{figure}[h]
    \centering
    \includegraphics[width=0.49\linewidth]{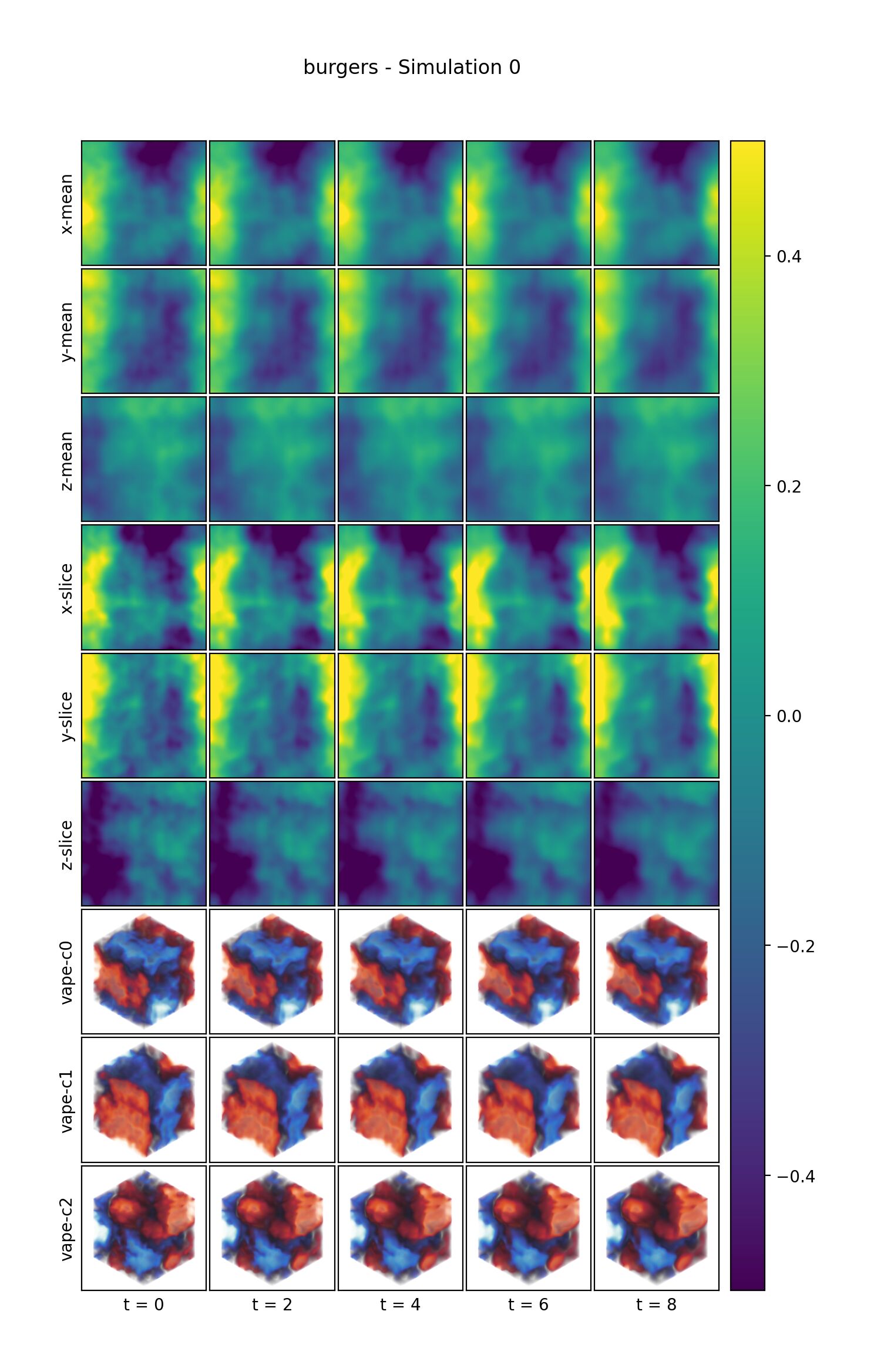}
    \includegraphics[width=0.49\linewidth]{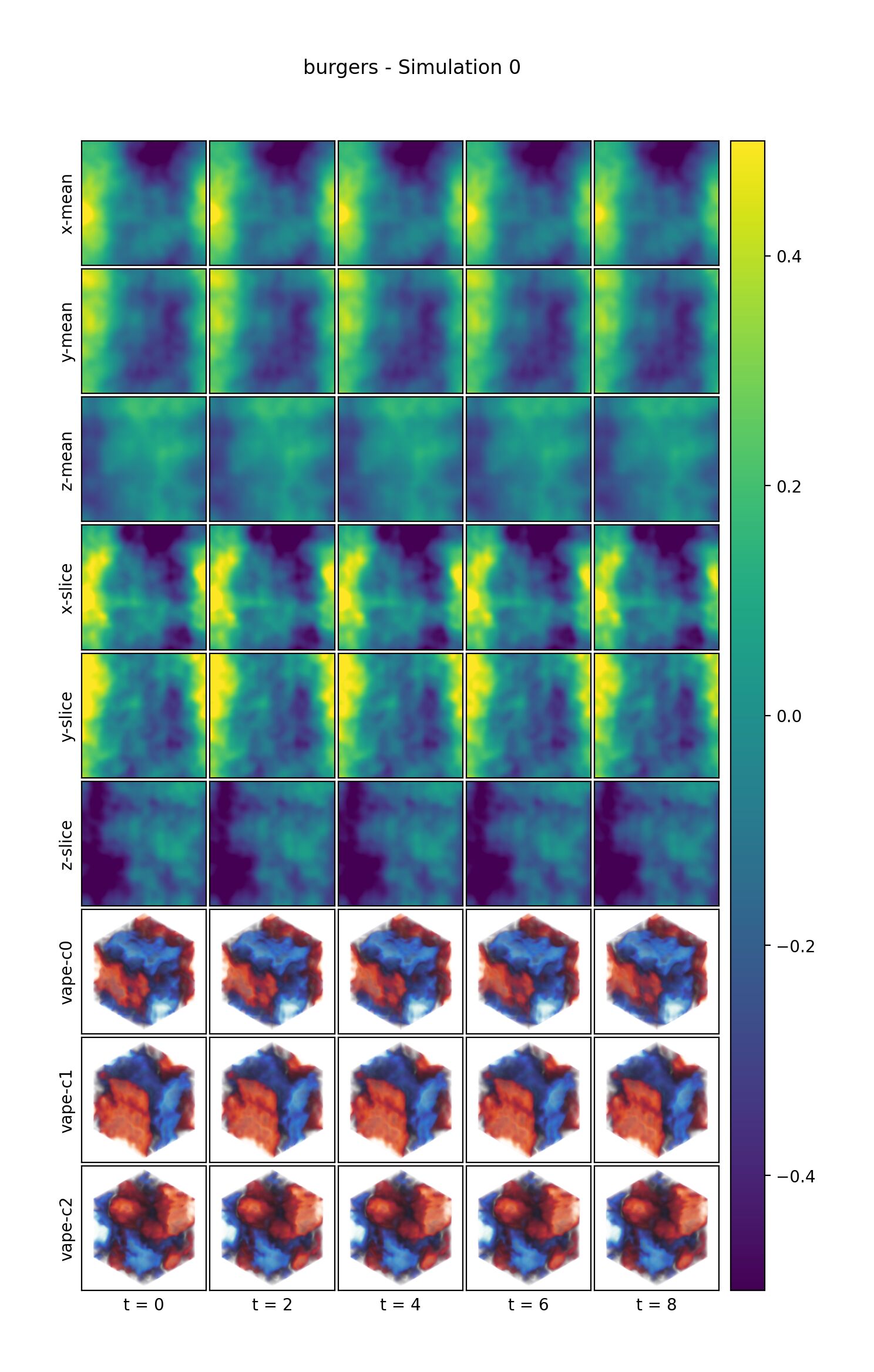}
    \caption{\dBurgers{}. Reference (left) and autoregressive prediction for $t=8$ steps  with P3D-S <128|320> (right) on the test set at resolution $320^3$.}
    \label{fig:burgers prediction}
\end{figure}

\begin{figure}
    \centering
    \includegraphics[width=0.49\linewidth]{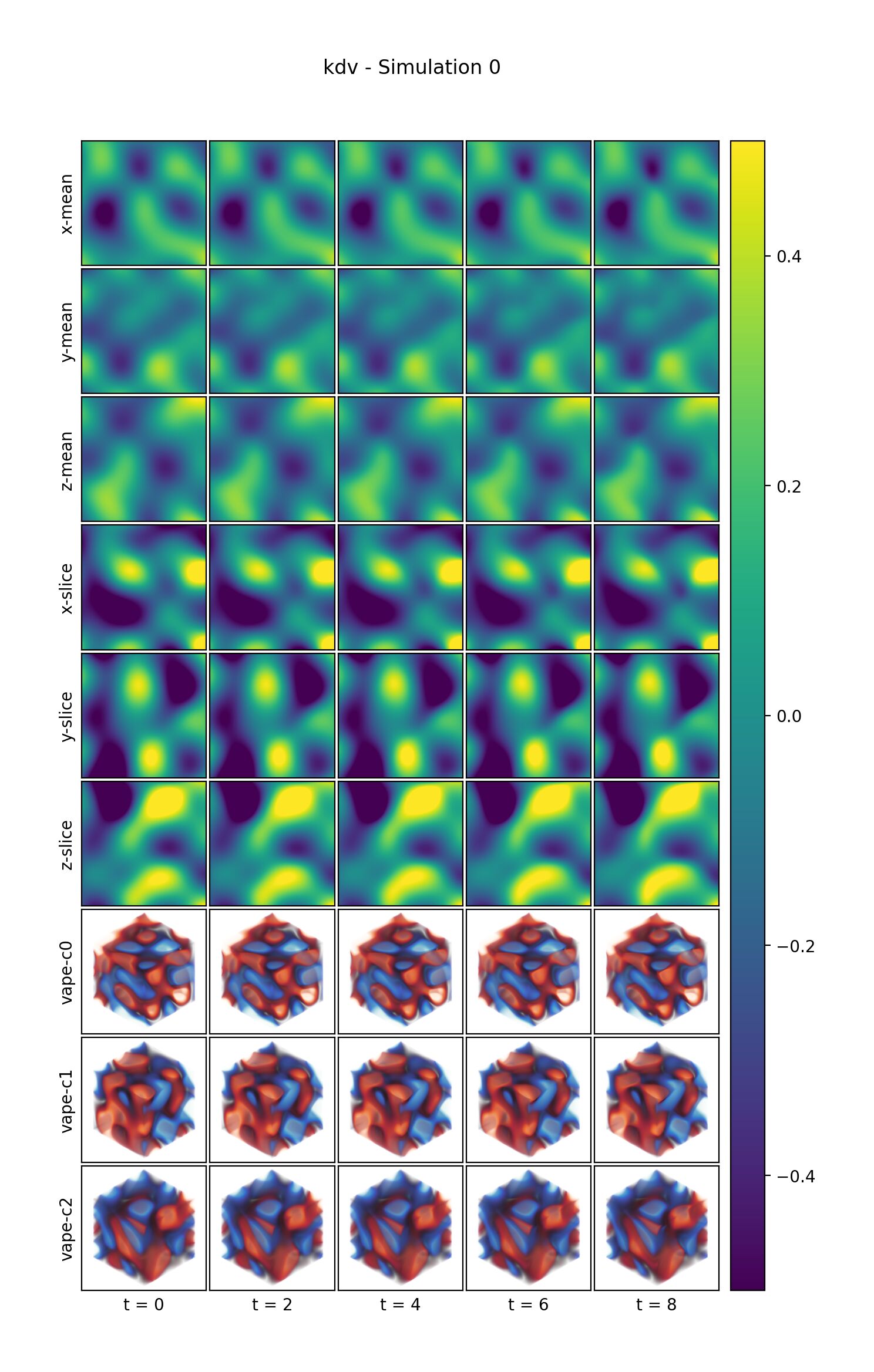}
    \includegraphics[width=0.49\linewidth]{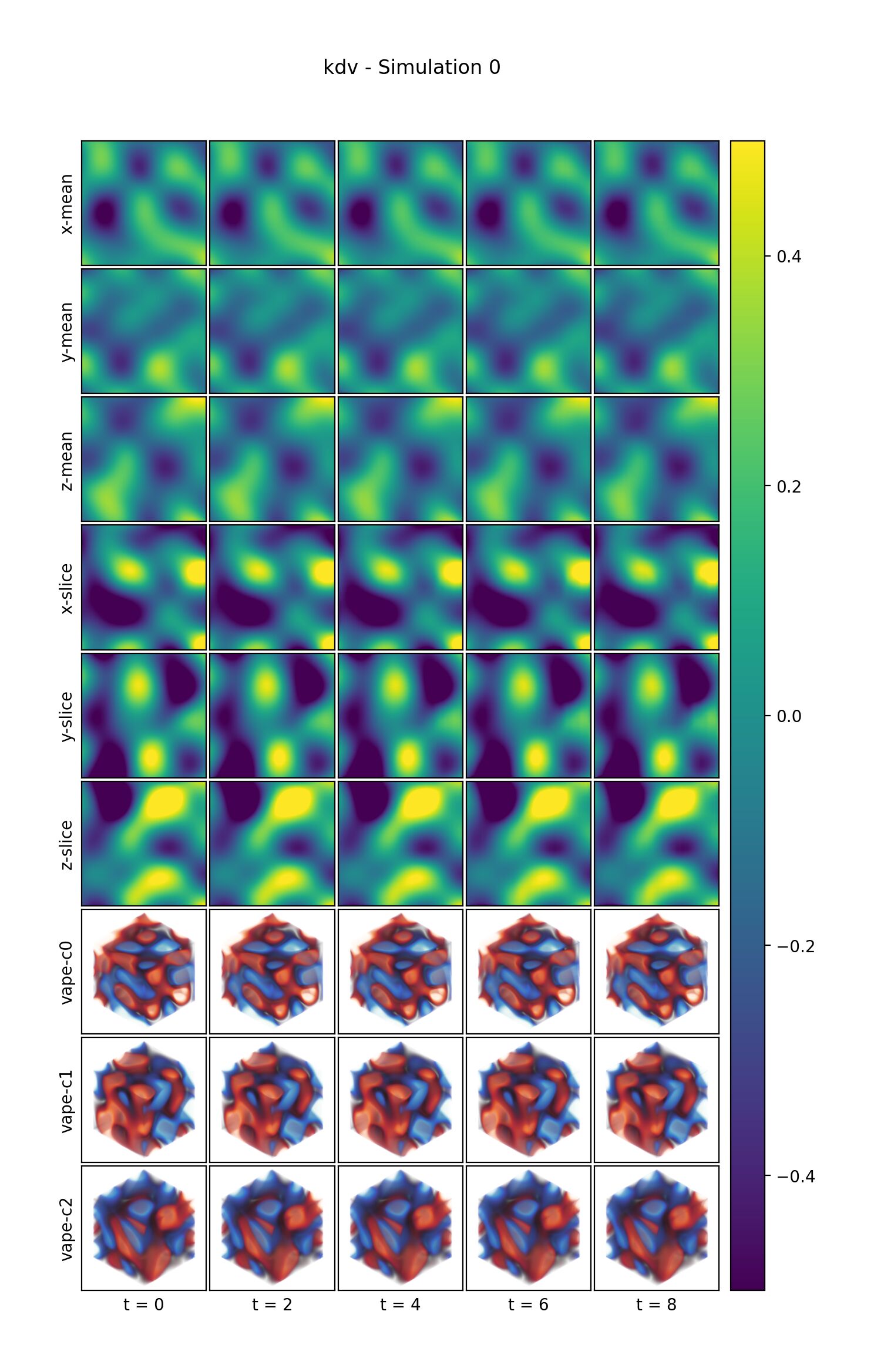}
    \caption{\dKDV{}. Reference (left) and autoregressive prediction for $t=8$ steps  with P3D-S <128|320> (right) on the test set at resolution $320^3$.}
    \label{fig:kdv prediction}
\end{figure}

\begin{figure}
    \centering
    \includegraphics[width=0.49\linewidth]{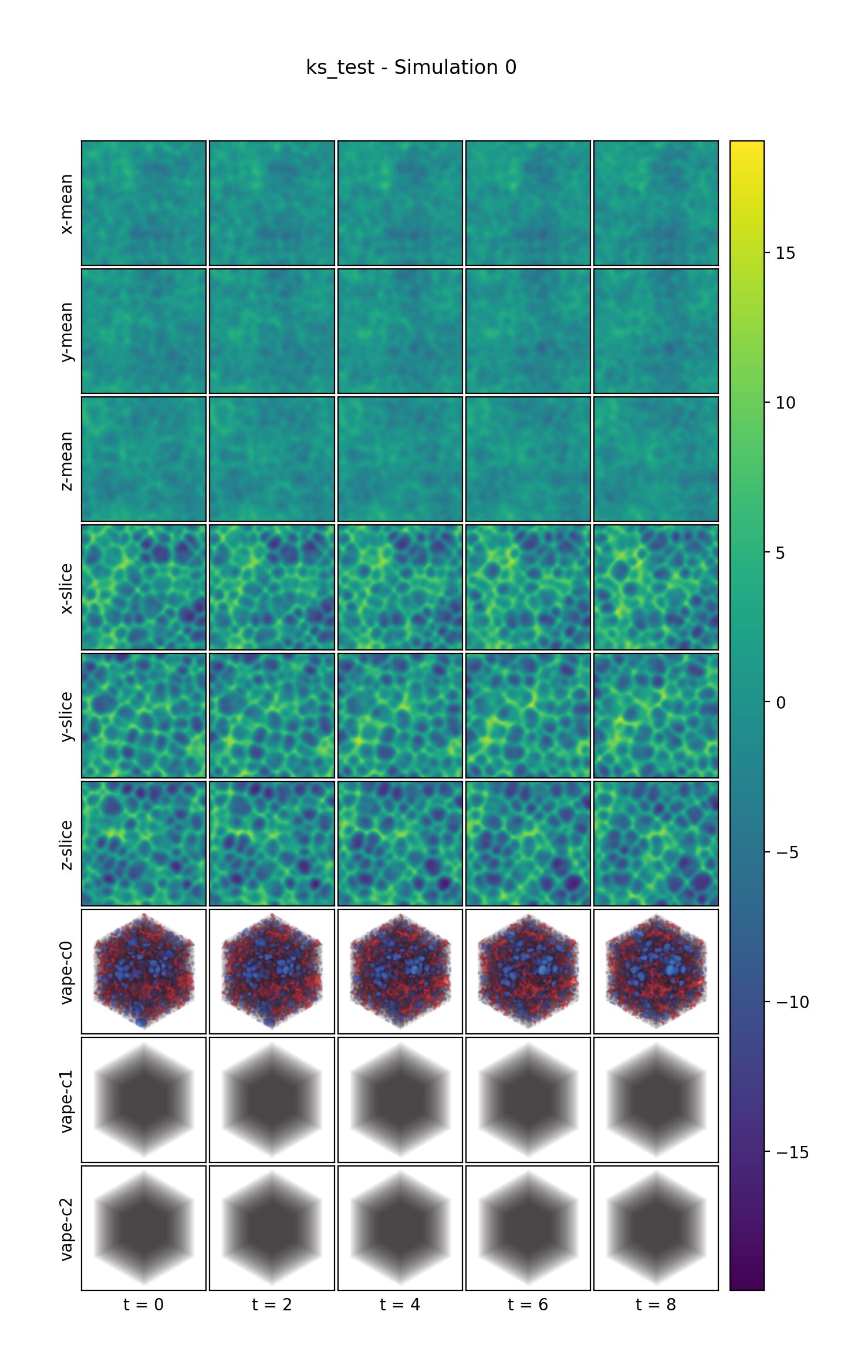}
    \includegraphics[width=0.49\linewidth]{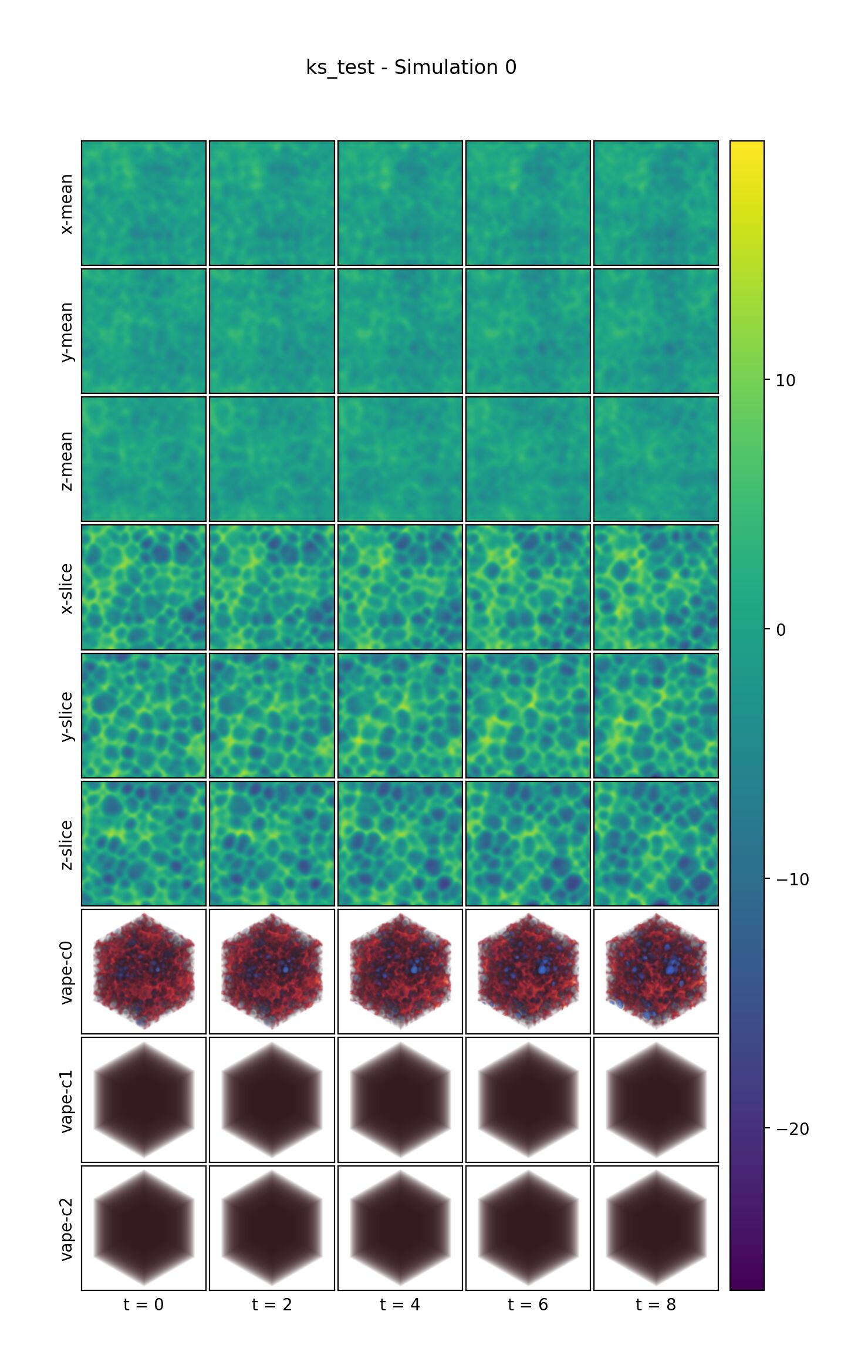}
    \caption{\dKS{}. Reference (left) and autoregressive prediction for $t=8$ steps  with P3D-S <128|320> (right) on the test set at resolution $320^3$.}
    \label{fig:ks prediction}
\end{figure}

Below, we visualize several example predictions on the test datasets from the P3D-S network trained on crop size $128$. See \cref{fig:disp prediction,fig:hyp prediction,fig:fisher prediction,fig:sh prediction,fig:gs alpha prediction,fig:gs epsilon prediction,fig:burgers prediction,fig:kdv prediction,fig:ks prediction}. During inference, we apply the network to larger crops of domain size $320^3$, which is significantly larger than what the network was orginially trained on. We focus on smaller autoregressive rollouts of up to $t=8$ steps. The network has to estimate the average behaviour outside the given input domain, which significantly affects the dynamics for longer time horizons.  

\clearpage

 \begin{table}[h!] \centering \caption{Normalized RMSE ($\times 10^{-3}$) for crop size 128.}  \label{tab:nrmse_128_performance_metrics} \resizebox{\textwidth}{!}{ \begin{tabular}{lccccccccccccccc} \toprule Model Name & Fisher & GS $\alpha$ & GS $\beta$ & GS $\delta$ & GS $\epsilon$ & GS $\gamma$ & GS $\iota$ & GS $\kappa$ & GS $\theta$ & Hyp & KDV & KS & SH & Burgers & Average \\ \midrule 

 P3D-$S$ & \num{8.3} & \num{15.7} & \num{29.3} & \num{14.1} & \num{13.4} & \num{47.4} & \num{12.9} & \num{16.0} & \num{6.3} & \num{73.4} & \num{92.0} & \num{18.9} & \num{91.4} & \num{27.2} & \num{33.3} \\ P3D-$B$ & \num{7.1} & \num{9.0} & \num{14.9} & \num{8.1} & \num{7.7} & \num{19.9} & \num{9.1} & \num{10.6} & \num{4.7} & \num{75.3} & \num{75.3} & \num{19.0} & \num{75.6} & \num{16.8} & \num{25.2} \\ P3D-$L$ & \textbf{\num{6.9}} & \textbf{\num{5.7}} & \textbf{\num{8.9}} & \textbf{\num{5.7}} & \textbf{\num{5.4}} & \textbf{\num{16.0}} & \textbf{\num{6.0}} & \textbf{\num{7.1}} & \textbf{\num{2.3}} & \textbf{\num{69.1}} & \textbf{\num{65.8}} & \textbf{\num{18.4}} & \textbf{\num{62.7}} & \textbf{\num{11.1}} & \textbf{\num{20.8}} \\

 \midrule
 
 AFNO &  \num{13.0} &  \num{20.7} & \num{30.4} & \num{18.8} & \num{23.3} & \num{43.3} & \num{17.1} & \num{20.0} & \num{6.1} & \num{146.5} & \num{81.5} & \num{32.4} & \num{183.5} & \num{34.6} & \num{47.9} \\ AVIT & \num{124.6} & \num{49.7} & \num{60.8} & \num{51.7} & \num{62.7} & \num{101.0} & \num{47.6} & \num{48.1} & \num{56.7} & \num{619.2} & \num{191.2} & \num{65.1} & \num{470.7} & \num{177.0} & \num{151.9} \\  Swin3D  & \num{11.7} &  \num{24.3} & \num{30.7} & \num{22.0} & \num{23.0} & \num{50.0} & \num{22.8} & \num{17.5} & \num{5.2} & \num{153.6} & \num{90.1} & \num{27.0} & \num{190.8} & \num{36.7} & \num{50.4} \\ 
 UNet$_\mathrm{GenCFD}$ & \num{17.3} & \num{35.8} & \num{19.5} & \num{6.4} & \num{24.9} & \num{50.7} & \num{6.4} & \num{6.4} & \num{6.4} & \num{214.2} & \num{107.7} & \num{43.2} & \num{573.1} & \num{45.8} & \num{82.7} \\
 \bottomrule \end{tabular} } \end{table}

 \begin{table}[h!] \centering \caption{Normalized RMSE ($\times 10^{-3}$) for crop size 64.} \label{tab:nrmse_64_performance_metrics} \resizebox{\textwidth}{!}{ 
 \begin{tabular}{lccccccccccccccc} 
 \toprule Model Name & Fisher & GS $\alpha$ & GS $\beta$ & GS $\delta$ & GS $\epsilon$ & GS $\gamma$ & GS $\iota$ & GS $\kappa$ & GS $\theta$ & Hyp & KDV & KS & SH & Burgers & Average \\ \midrule 

 P3D-$S$  & \num{10.4} & \num{21.0} & \num{25.4} & \num{5.2} & \num{18.1} & \num{43.3} & \num{4.6} & \num{4.6} & \num{4.6} & \num{92.2} & \num{100.4} & \num{50.5} & \num{111.8} & \num{34.4} & \num{37.6} \\ P3D-$B$ & \textbf{\num{8.3}} & \bf \num{11.7} & \textbf{\num{10.9}} & \textbf{\num{4.2}} & \textbf{\num{10.1}} & \textbf{\num{26.1}} & \textbf{\num{3.8}} & \textbf{\num{3.8}} & \textbf{\num{3.8}} & \textbf{\num{83.6}} & \textbf{\num{92.2}} & \textbf{\num{42.1}} & \textbf{\num{98.7}} & \textbf{\num{26.2}} & \textbf{\num{30.4}} \\
 P3D-$L$ & \num{6.8} & \num{7.3} & \num{10.1} & \num{7.1} & \num{6.6} & \num{17.5} & \num{8.8} & \num{8.3} & \num{1.4} & \num{69.6} & \num{73.7} & \num{28.5} & \num{88.1} & \num{15.0} & \num{24.9} \\ 

 \midrule
 
 AFNO & \num{19.7} & \num{20.0} & \num{22.7} & \num{3.1} & \num{21.0} & \num{38.0} & \num{2.3} & \num{2.3} & \num{2.3} & \num{189.4} & \num{90.8} & \num{39.4} & \num{210.7} & \num{36.3} & \num{49.9} \\ 
 AVIT & \num{175.6} & \num{81.9} & \num{53.6} & \num{19.5} & \num{88.3} & \num{106.0} & \num{19.0} & \num{19.0} & \num{19.0} & \num{2004.7} & \num{206.6} & \num{92.7} & \num{450.8} & \num{213.9} & \num{253.6} \\
 Swin3D & \num{20.7} & \num{67.4} & \num{56.9} & \num{5.0} & \num{59.7} & \num{98.4} & \num{4.5} & \num{4.5} & \num{4.5} & \num{155.6} & \num{115.9} & \num{61.2} & \num{285.3} & \num{46.2} & \num{70.4} \\ 
 FactFormer & \num{19.06} & \num{18.01} & \num{13.94} & \num{3.75} & \num{16.10} & \num{30.76} & \num{3.75} & \num{3.75} & \num{3.75} & \num{182.78} & \num{95.13} & \num{31.53} & \num{184.20} & \num{40.48} & \num{46.2} \\
 UNet$_\mathrm{GenCFD}$ & \num{36.6} & \num{35.1} & \num{72.7} & \num{14.0} & \num{42.5} & \num{65.5} & \num{13.7} & \num{13.6} & \num{13.7} & \num{272.4} & \num{128.4} & \num{37.0} & \num{324.4} & \num{56.2} & \num{80.4} \\  
UNet$_\mathrm{ConvNeXt}$ & \num{18.6} & \num{40.3} & \num{49.8} & \num{6.8} & \num{36.2} & \num{66.8} & \num{6.5} & \num{6.5} & \num{6.5} & \num{190.8} & \num{108.8} & \num{42.1} & \num{364.5} & \num{49.0} & \num{70.9} \\
 TFNO & \num{23.2} & \num{111.6} & \num{56.2} & \num{6.3} & \num{49.5} & \num{96.7} & \num{6.0} & \num{6.0} & \num{6.0} & \num{175.5} & \num{113.8} & \num{222.1} & \num{263.1} & \num{36.2} & \num{83.7} \\ \bottomrule \end{tabular} } \end{table}

\begin{table}[h!] \centering \caption{Normalized RMSE ($\times 10^{-3}$) for crop size 32.} \label{tab:nrmse_32_performance_metrics} \resizebox{\textwidth}{!}{ 
\begin{tabular}{lccccccccccccccc} \toprule Model Name & Fisher & GS $\alpha$ & GS $\beta$ & GS $\delta$ & GS $\epsilon$ & GS $\gamma$ & GS $\iota$ & GS $\kappa$ & GS $\theta$ & Hyp & KDV & KS & SH & Burgers & Average \\ \midrule 
P3D-$S$  & \num{9.2} & \num{49.7} & \num{56.3} & \num{6.3} & \num{33.7} & \num{57.2} & \num{6.3} & \num{6.3} & \num{6.3} & \num{146.5} & \num{127.3} & \num{158.4} & \num{176.2} & \num{38.8} & \num{62.7} \\ 
P3D-$B$  & \num{8.3}  & \num{24.5} & \num{14.8} & \num{2.5} & \num{15.9} & \num{40.1} & \num{2.5} & \num{2.5} & \num{2.5} & \num{131.1} & \num{126.4} & \num{117.5} & \num{131.7} & \num{37.9} & \num{46.9} \\ 
P3D-$L$ & \textbf{\num{7.9}} & \num{18.5} & \textbf{\num{11.4}} & \textbf{\num{1.9}} & \num{14.2} & \num{32.4} & \textbf{\num{1.9}} & \textbf{\num{1.9}} & \textbf{\num{1.9}} & \num{109.3} & \num{110.8} & \num{110.9} & \num{129.4} & \num{27.3} & \num{41.4} \\

\midrule

AFNO & \num{21.7}  & \num{43.8} & \num{22.0} & \num{5.8} & \num{33.4} & \num{65.9} & \num{5.8} & \num{5.8} & \num{5.8} & \num{188.7} & \num{118.3} & \num{74.0} & \num{619.5} & \num{43.1} & \num{89.5} \\
AVIT & \num{237.1} & \num{126.7} & \num{50.7} & \num{13.6} & \num{107.6} & \num{122.7} & \num{13.6} & \num{13.6} & \num{13.6} & \num{1089.4} & \num{256.7} & \num{177.6} & \num{476.0} & \num{230.1} & \num{209.2} \\ 
Swin3D & \num{16.6} & \num{154.5} & \num{38.6} & \num{3.8} & \num{56.2} & \num{92.6} & \num{3.8} & \num{3.8} & \num{3.8} & \num{189.1} & \num{143.6} & \num{83.4} & \num{263.5} & \num{55.7} & \num{79.2} \\ 
FactFormer & \num{20.1} & \num{43.1} & \num{23.1} & \num{5.8} & \num{32.0} & \num{43.9} & \num{5.8} & \num{5.8} & \num{5.8} & \num{222.9} & \num{105.0} & \num{54.4} & \num{263.7} & \num{43.3} & \num{62.5} \\ 
UNet$_\mathrm{GenCFD}$ & \num{18.5} & \num{38.6} & \num{22.1} & \num{3.5} & \num{20.9} & \num{53.6} & \num{3.5} & \num{3.5} & \num{3.6} & \num{196.3} & \num{105.8} & \num{37.0} & \num{517.2} & \num{41.9} & \num{76.1} \\ 
UNet$_\mathrm{ConvNeXt}$ & \num{20.3} & \num{111.6} & \num{38.3} & \num{7.4} & \num{49.2} & \num{89.9} & \num{7.4} & \num{7.4} & \num{7.4} & \num{166.3} & \num{128.5} & \num{77.4} & \num{442.5} & \num{48.9} & \num{85.9} \\ 
TFNO & \num{21.9} & \num{176.8} & \num{41.2} & \num{5.6} & \num{46.4} & \num{84.4} & \num{5.6} & \num{5.6} & \num{5.6} & \num{175.3} & \num{121.8} & \num{204.1} & \num{252.4} & \num{38.3} & \num{84.7} \\ \bottomrule \end{tabular} } \end{table}

\clearpage

\section{Experiment 2: Isotropic Turbulence} \label{app:isotropic_turbulence}

For the second experiment, we make use of
the Johns Hopkins Turbulence Database (\textit{JHTDB}). It contains data from various direct numerical simulations of homogeneous and wall-bounded turbulent flows \citep{perlman2007_JHTDB}. The simulations are stored with space-time history and allow for arbitrary spatiotemporal query points. 

\paragraph{Isotropic Turbulence (\dIsoTurb{})} a direct numerical simulation of the Navier-Stokes equations at Reynolds number around 433 simulated on a $1024^3$ periodic grid via a pseudo-spectral parallel code. It contains homogeneous isotropic turbulence, i.e., an idealized version of realistic turbulence with statistical properties that are invariant to translations and rotations of the coordinate axes. The following overview summarizes key characteristics of the dataset \citep[for further details see][]{perlman2007_JHTDB}:

\begin{itemize}[itemsep=0pt]
    \item Dimensionality: $s=1$, $t=500$, $f=4$, $x=512$, $y=512$, $z=512$
    \item Boundary conditions: periodic
    \item Time step of stored data: 0.002
    \item Spatial domain size: $[0, 2\pi] \times [0, 2\pi] \times [0, 2\pi]$
    \item Fields: velocity X/Y/Z, pressure
    \item Validation set: random $15\%$ split of all timesteps from $t \in [0,420]$
    \item Test set: all sequences from $t \in [420,500]$ 
\end{itemize}

\begin{figure}
    \centering
    \includegraphics[width=\linewidth]{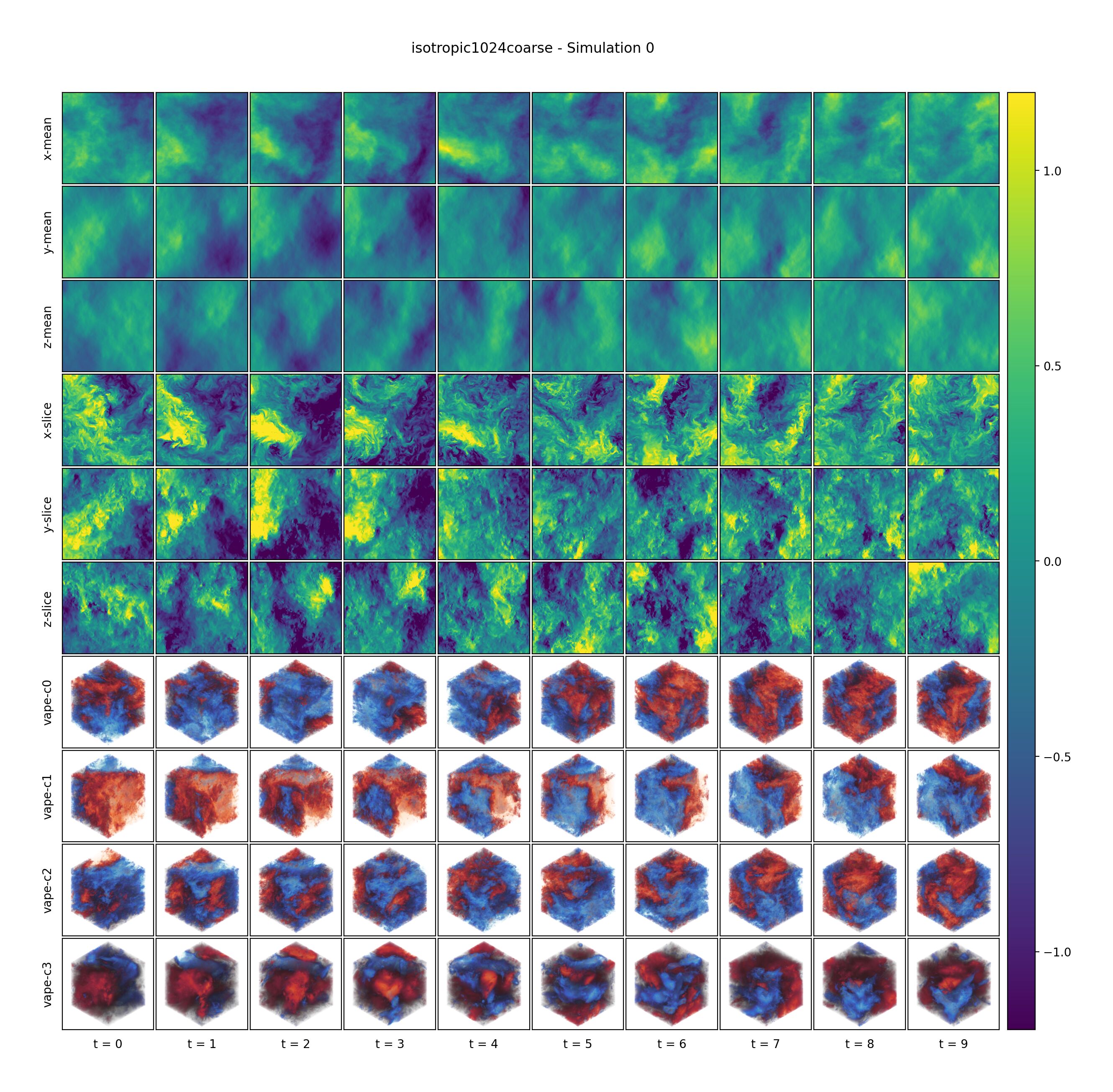}
    \caption{Isotropic Turbulence. Training dataset visualization at resolution $512^3$ showing the velocity X/Y/Z and pressure from $t=0$ until $t=420$.}
    \label{fig:isotropic dataset}
\end{figure}

\clearpage

\paragraph{Enstrophy graph}

See \cite[Appendix C.2]{chen2024probabilistic} for reference. The vorticity is not part of the data channels, only the velocity channels in X/Y/Z are there. 
Therefore, we compute the vorticity via finite differences from the velocity. The data on the cropped domain of size $128^3$ is not periodic, thus there are artifacts at the boundary of the crop. This would lead to problems when calculating the spectral coefficients via the Fourier transform.

\begin{figure}[h]
    \centering
    \includegraphics[width=0.9\linewidth]{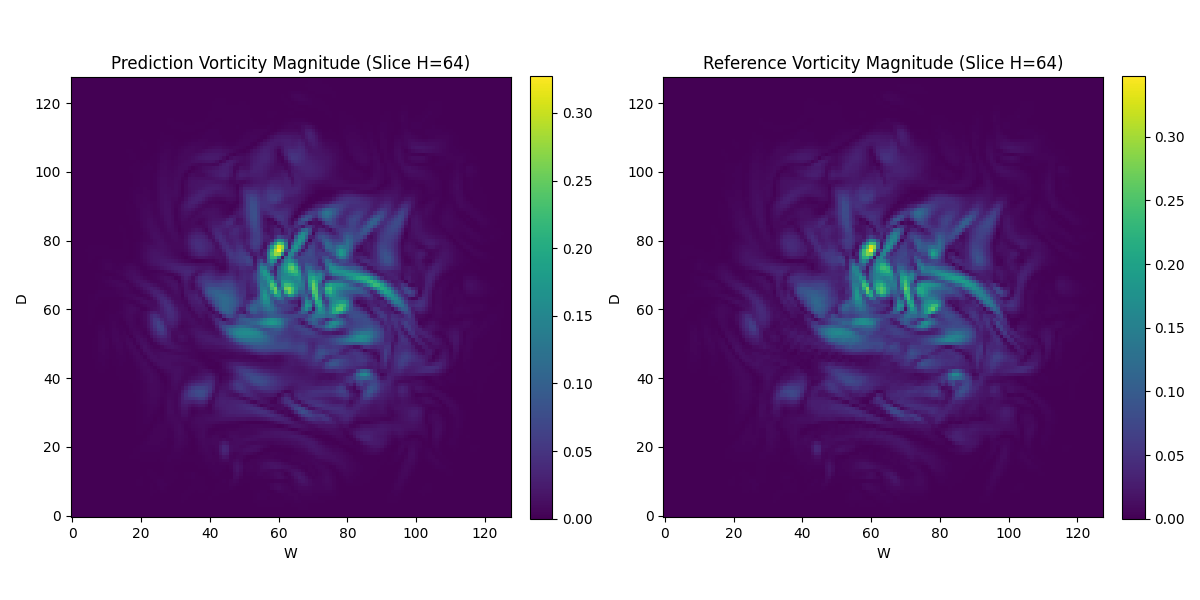}
    \caption{Slice of predicted vorticity of P3D-$B$ vs. reference after 1 step and applying the Hanning filter.}
\label{fig:hanning_filter_vorticity_pred}
\end{figure}

We therefore smoothen the vorticity field towards the boundary using a filter using the Hanning filter. Below is a visualization in \cref{fig:hanning_filter_vorticity_pred} is a visualization of the vorticity of P3D after applying the Hanning filter.

\begin{figure}
    \centering
     \includegraphics[width=0.45\linewidth]{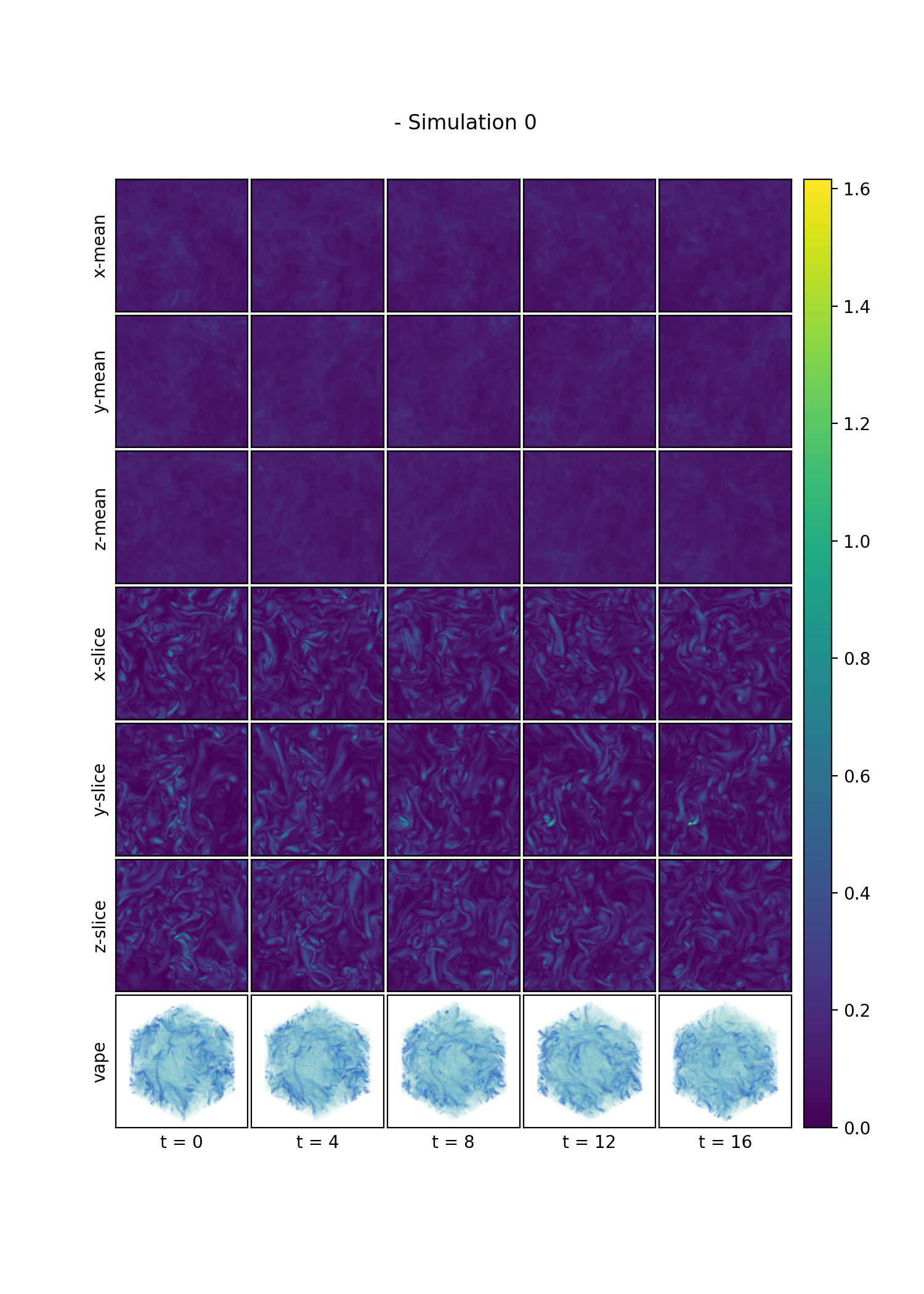}\includegraphics[width=0.45\linewidth]{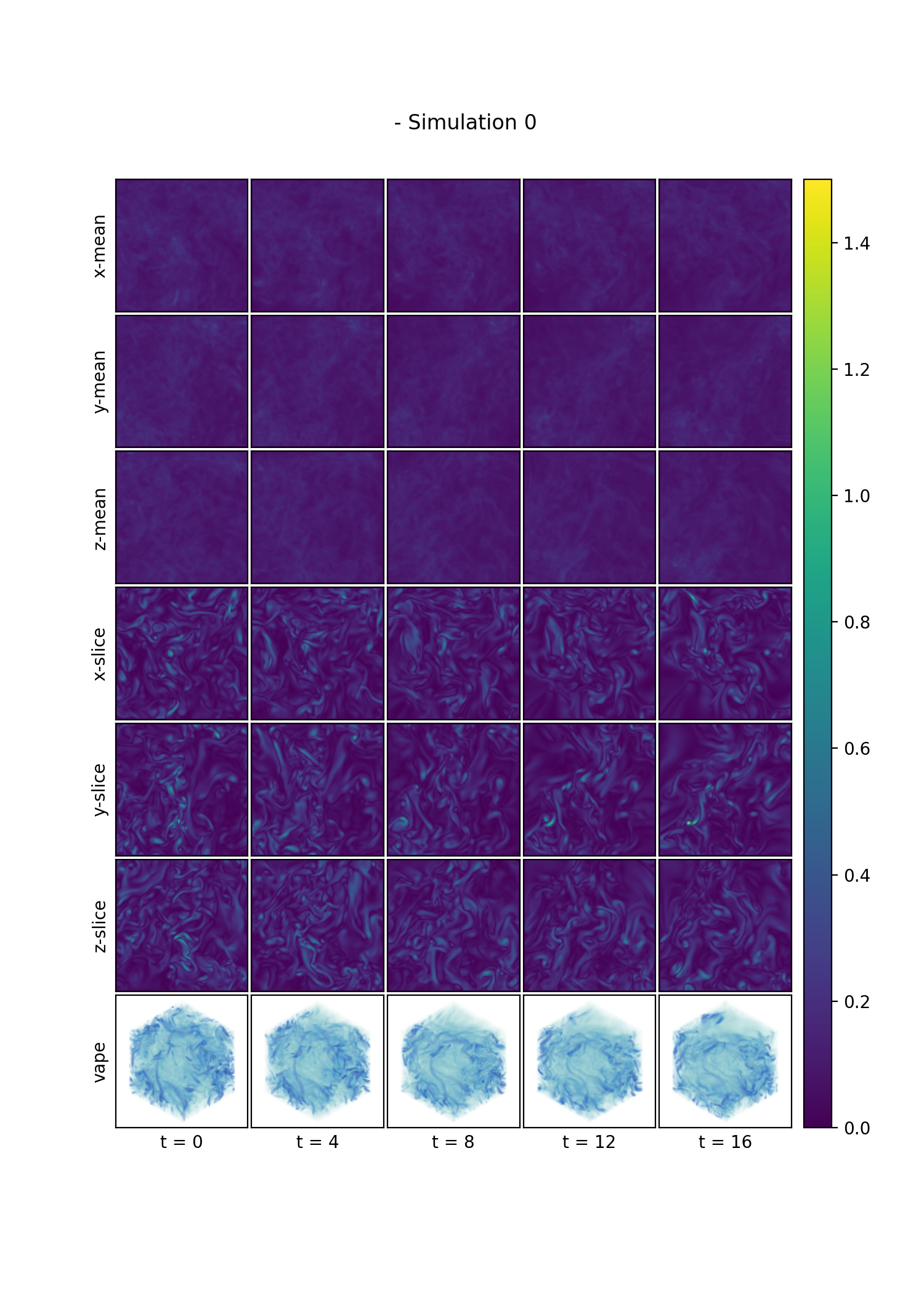}
    \caption{Vorticity. Reference (left) and autoregressive prediction with P3D-$B$ for $t=16$ steps (right) on the test set at resolution $128^3$.}
    \label{fig:vorticity_autoregressive}
\end{figure}

See \cref{fig:vorticity_autoregressive} for a comparison between the vorticity predicted by P3D and the reference without the Hanning filter. 

Based on the vorticity, we compute the enstrophy graph, which we we average over all region crops of the reference at the corresponding simulation time.

\begin{figure}
    \centering
    \includegraphics[width=0.45\linewidth]{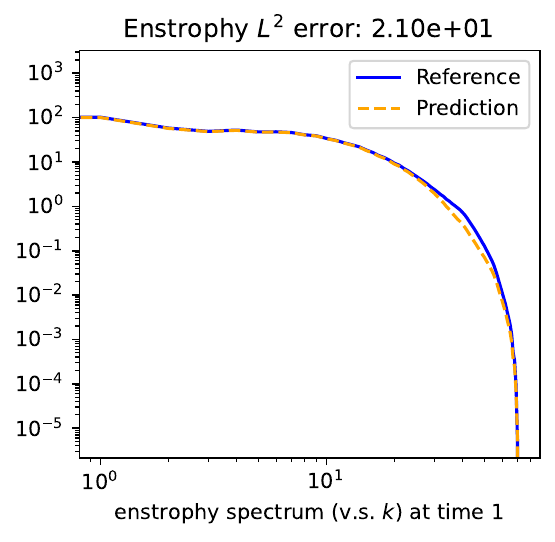}
    \includegraphics[width=0.45\linewidth]{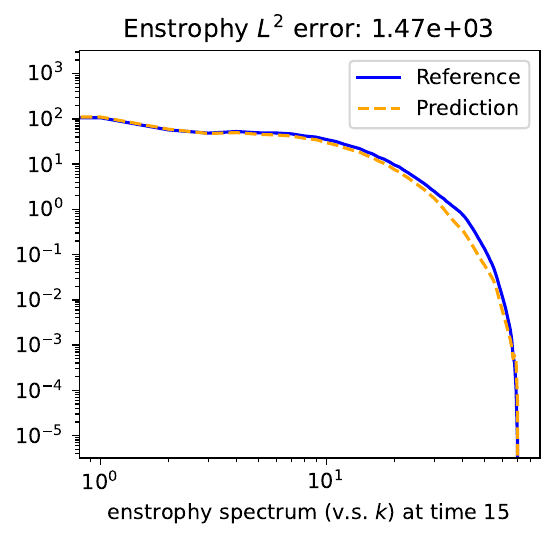}
    \caption{Enstrophy graphs of P3D-$B$ at resolution $128^3$ and the reference simulation at autoregressive prediction steps $t=1$ and $t=15$.}
    \label{fig:enstrophy_graph}
\end{figure}

See \cref{fig:enstrophy_graph} for the enstrophy graph of the predictions of P3D-$B$ and the reference at autoregressive unrolling steps $t=1$ and $t=15$ for the test set. We compute and report the L2 distance between the enstrophy graph of the predicted samples and the reference. 

\clearpage

\begin{figure}
    \centering
    \includegraphics[width=\linewidth]{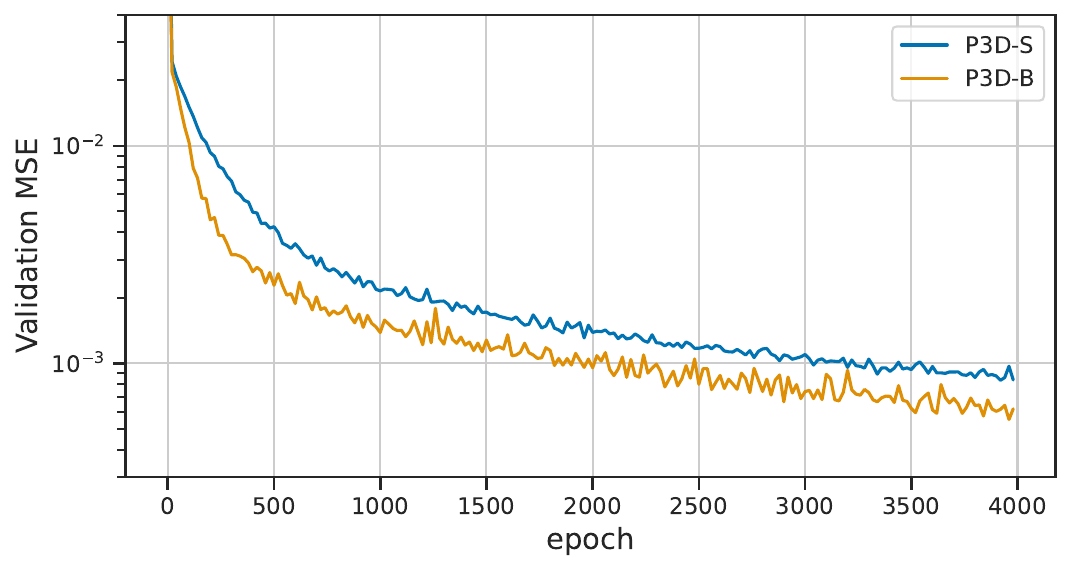}
    \caption{Isotropic Turbulence. Validation MSE for PDE-$S$ and PDE-$B$ during training.}
    \label{fig:isotropic training}
\end{figure}

\begin{figure}
    \centering
    \includegraphics[width=0.49\linewidth]{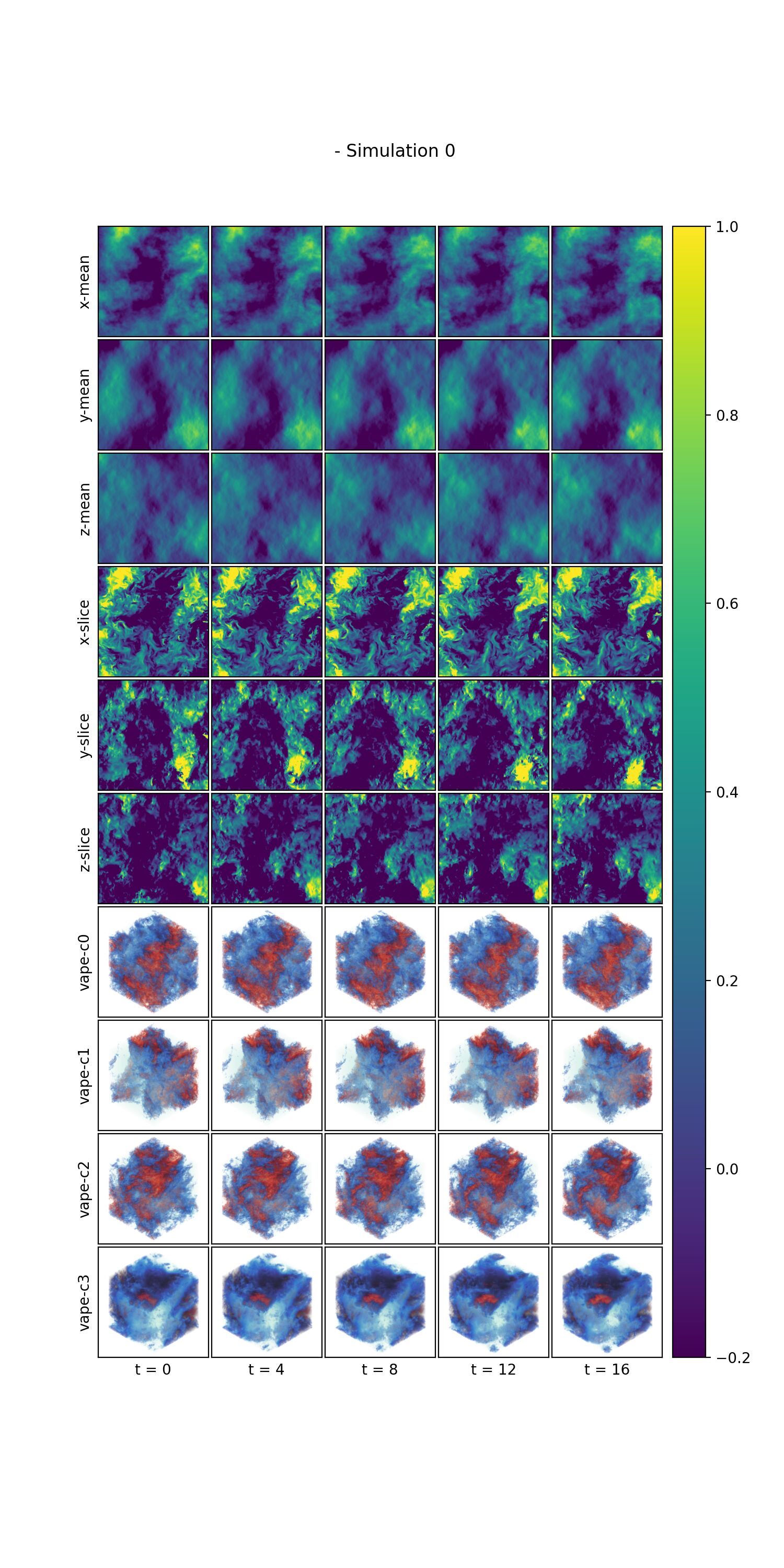}
    \includegraphics[width=0.49\linewidth]{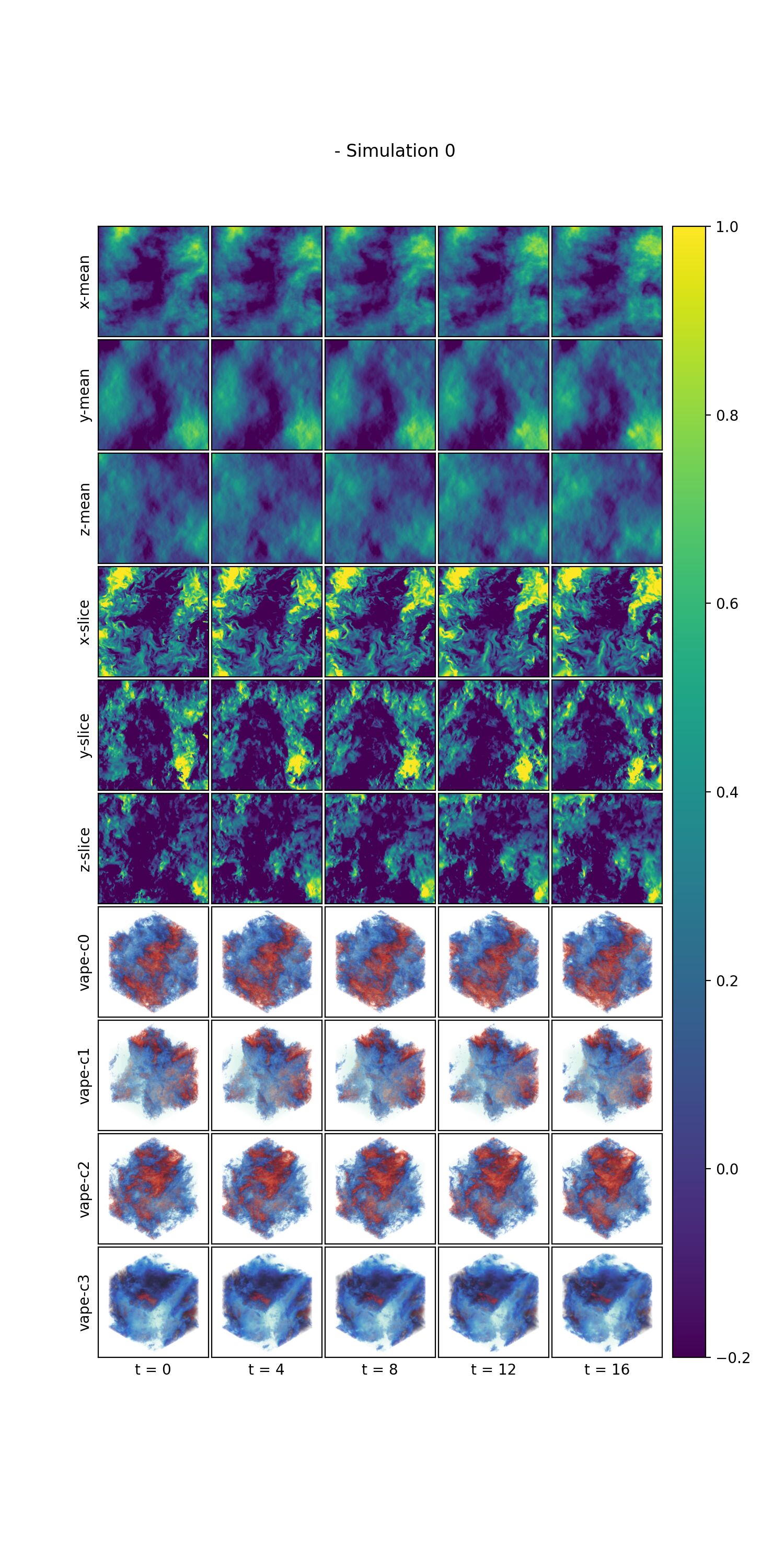}
    \caption{Isotropic Turbulence. Reference (left) and autoregressive prediction for $t=16$ steps  with P3D-S (right) on the test set at resolution $512^3$.}
    \label{fig:isotropic prediction}
\end{figure}

\paragraph{Training and Evaluation} We train both the $S$ and $B$ configurations of P3D for $4000$ epochs on crop size $128^3$, see \Cref{fig:isotropic training}. Training P3D-$S$ took 11h 48m and training P3D-$B$ took 20h 25m on 4 A100 GPUs. In \Cref{fig:isotropic prediction} we show a comparison between the reference and P3D-$S$ for a rollout until $t=16$ on the test set at resolution $512^3$.

\clearpage

\section{Experiment 3: Turbulent Channel Flow} \label{app:channel_flow}

\begin{figure*}[ht]
    \centering
    \includegraphics[width=0.99\textwidth]{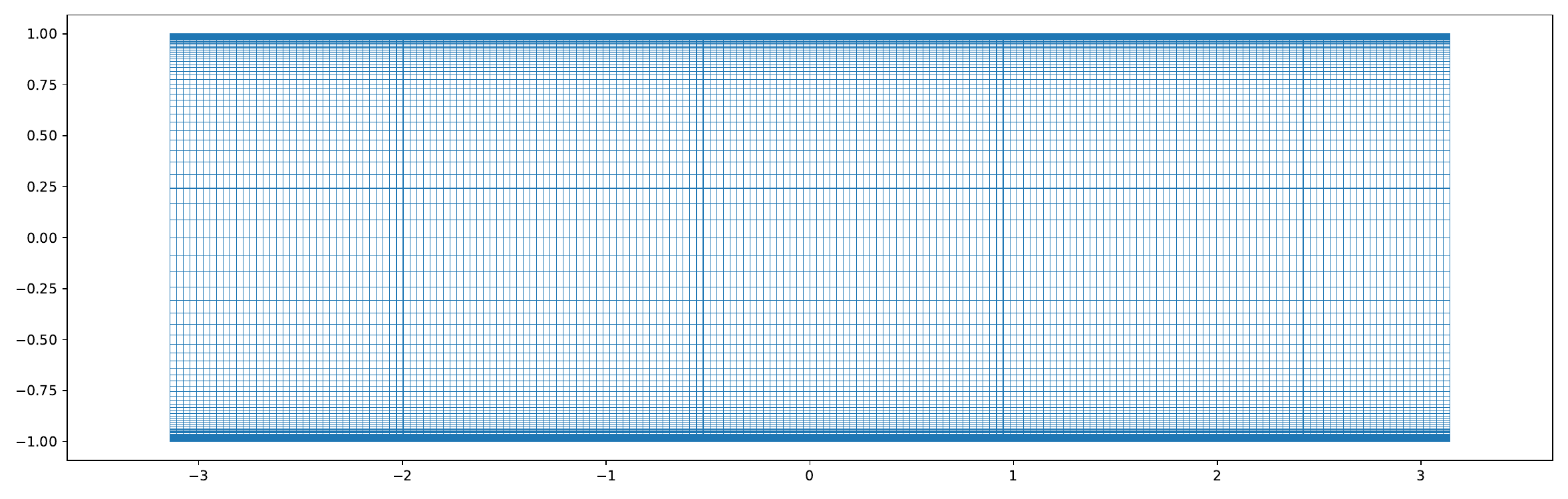}
    \caption{Computational grid of the turbulent channel flow simulation. The spatial discretization is refined in the near-wall region to  resolve the boundary layer.}
    \label{fig-app: tcf grid}
\end{figure*}

\begin{figure*}[ht]
    \centering
    \includegraphics[width=0.99\textwidth]{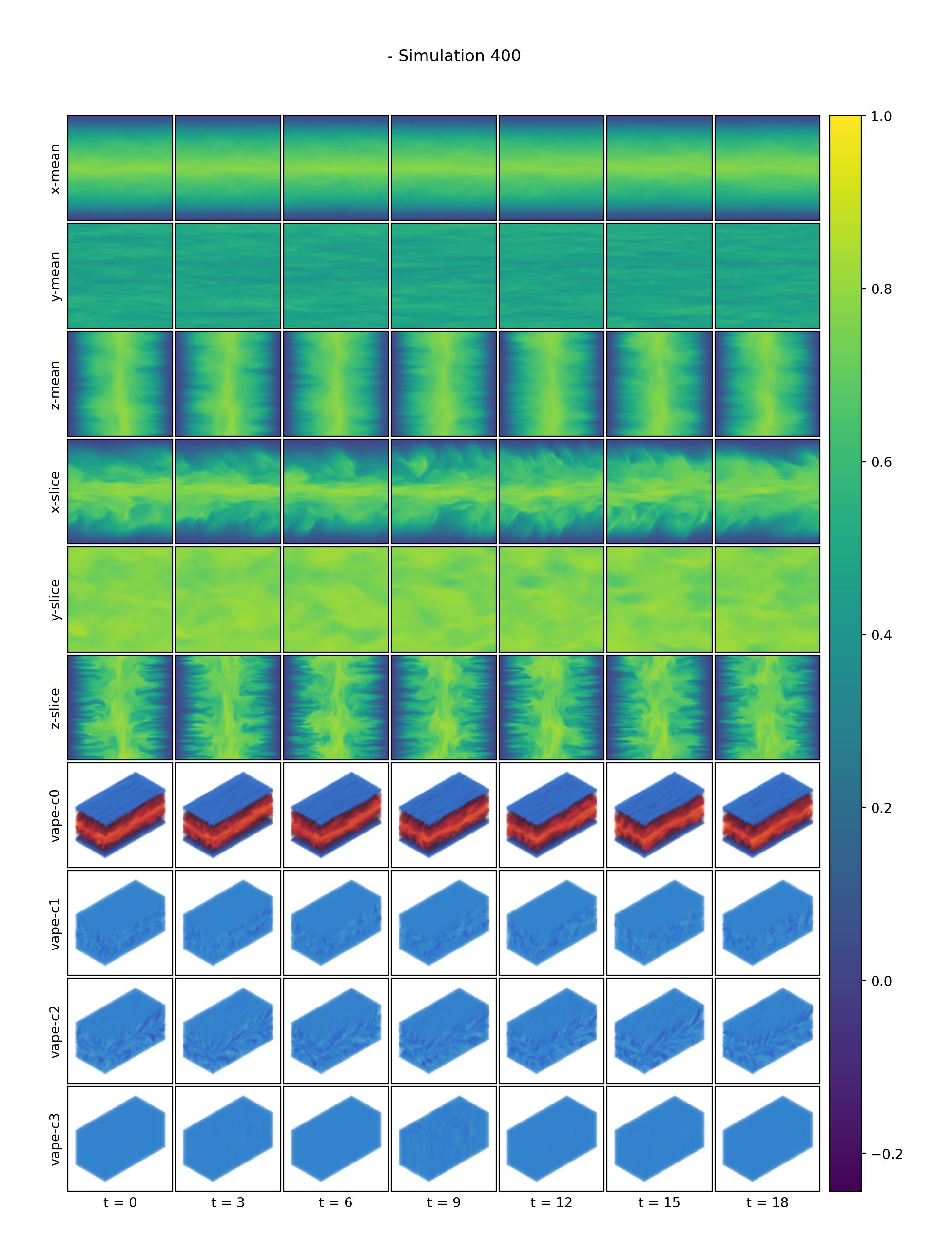}
    \caption{Turbulent channel flow with Reynolds number 400.}
    \label{fig-app: tcf RE 400}
\end{figure*}

\begin{figure*}[ht]
    \centering
    \includegraphics[width=0.99\textwidth]{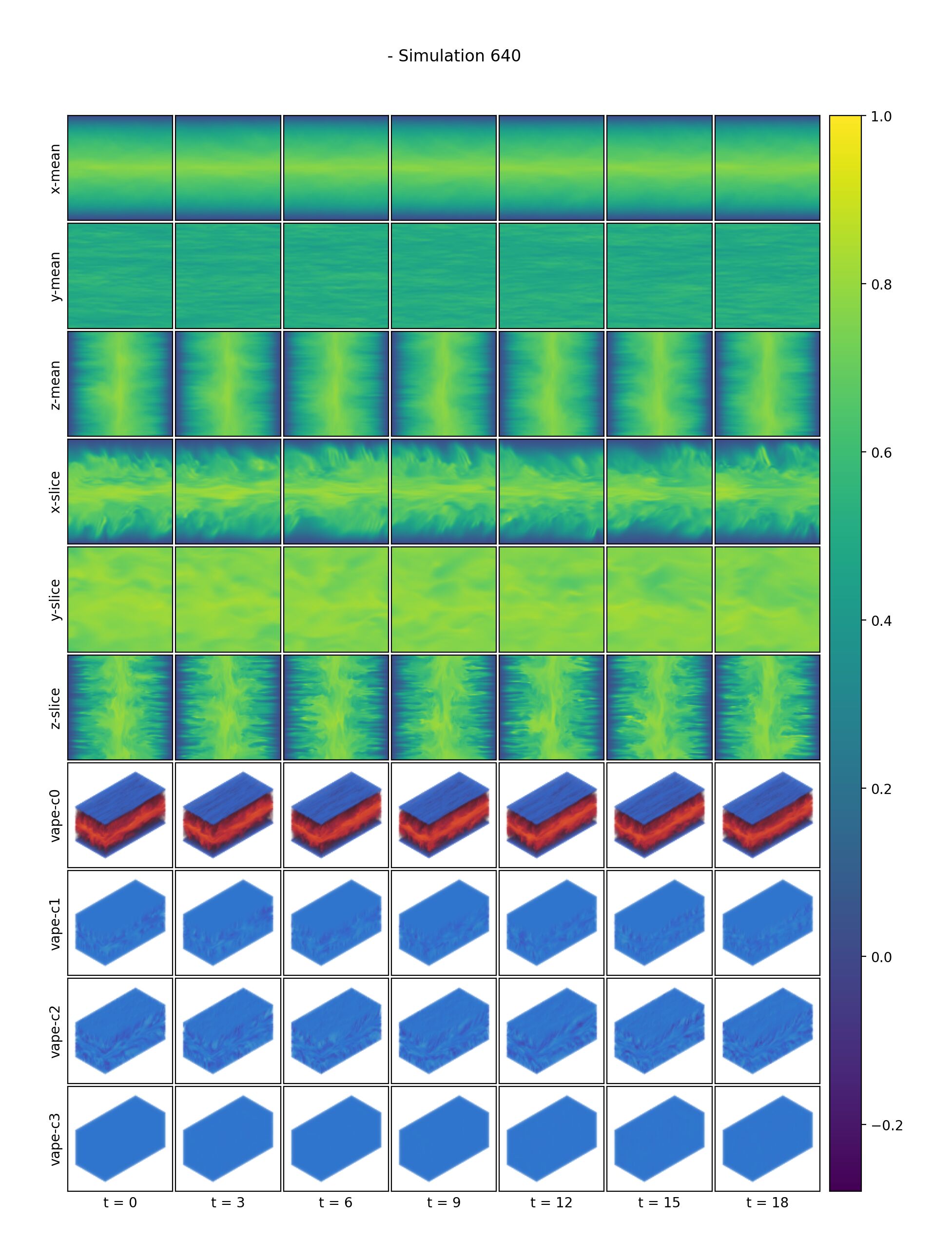}
    \caption{Turbulent channel flow with Reynolds number 640.}
    \label{fig-app: tcf RE 640}
\end{figure*}

The dataset for the turbulent channel flow (TCF) represents periodic channel with no-slip boundaries at $\pm$y that is driven by a dynamic forcing to re-inject energy lost due to wall friction, and prevent the flow from slowing down. This results in a continuous production of vortex structures at the walls, which have a very characteristic and well-studied, spatially-varying distribution~\cite{hoyas2008}.
Due to the complexity of the flow, these flows require very long transient phases to develop the characteristic structures, 
We target this scenario by learning with a generative model, in the context of which the TCF
problem represents a probabilistic learning problem to infer turbulent stats from the equilibrium phase, bypassing the costly transient warm-up phase.
\\
\\
\paragraph{Dataset} We generate a dataset comprising 20 simulations with Reynolds numbers within the interval $[400, 800]$ spaced equidistantly. After the initial-warmup phase, we simulate $ETT=20$ eddy turnover cycles, which we save in $200$ snapshots with $\Delta t = 0.1$. The computational grid comprises $96 \times 96 \times 192$ spatially adaptive cells with a finer discretization near the wall. The data contains channels for the velocity in X/Y/Z direction as well as pressure. We train P3D directly with computational grid data, which is shown in \Cref{fig-app: tcf grid}.
In \cref{fig-app: tcf RE 400,fig-app: tcf RE 640} we show visualizations of the turbulent channel flow for Reynold numbers $\text{Re}=400$ and $\text{Re}=640$ respectively.

\begin{itemize}[itemsep=0pt]
    \item Dimensionality: $s=20$, $t=200$, $f=4$, $x=96$, $y=96$, $z=192$
    \item Initial conditions: noise
    \item Boundary conditions: periodic (x), wall (y,z)
    \item Time step of stored data: 0.1 
    \item Number of warmup steps (discarded, in time step of data storage): 200
    \item Spatial domain: $[-1,1] \times [-1,1] \times [-\pi, \pi]$
    \item Fields: velocity X/Y/Z, pressure
    \item Varied parameters: Reynolds number $\in [400, 800]$
    \item Validation set: random $15\%$ split of Reynolds number
\end{itemize}

\clearpage

\begin{figure*}[ht]
    \centering
    \includegraphics[width=0.99\textwidth]{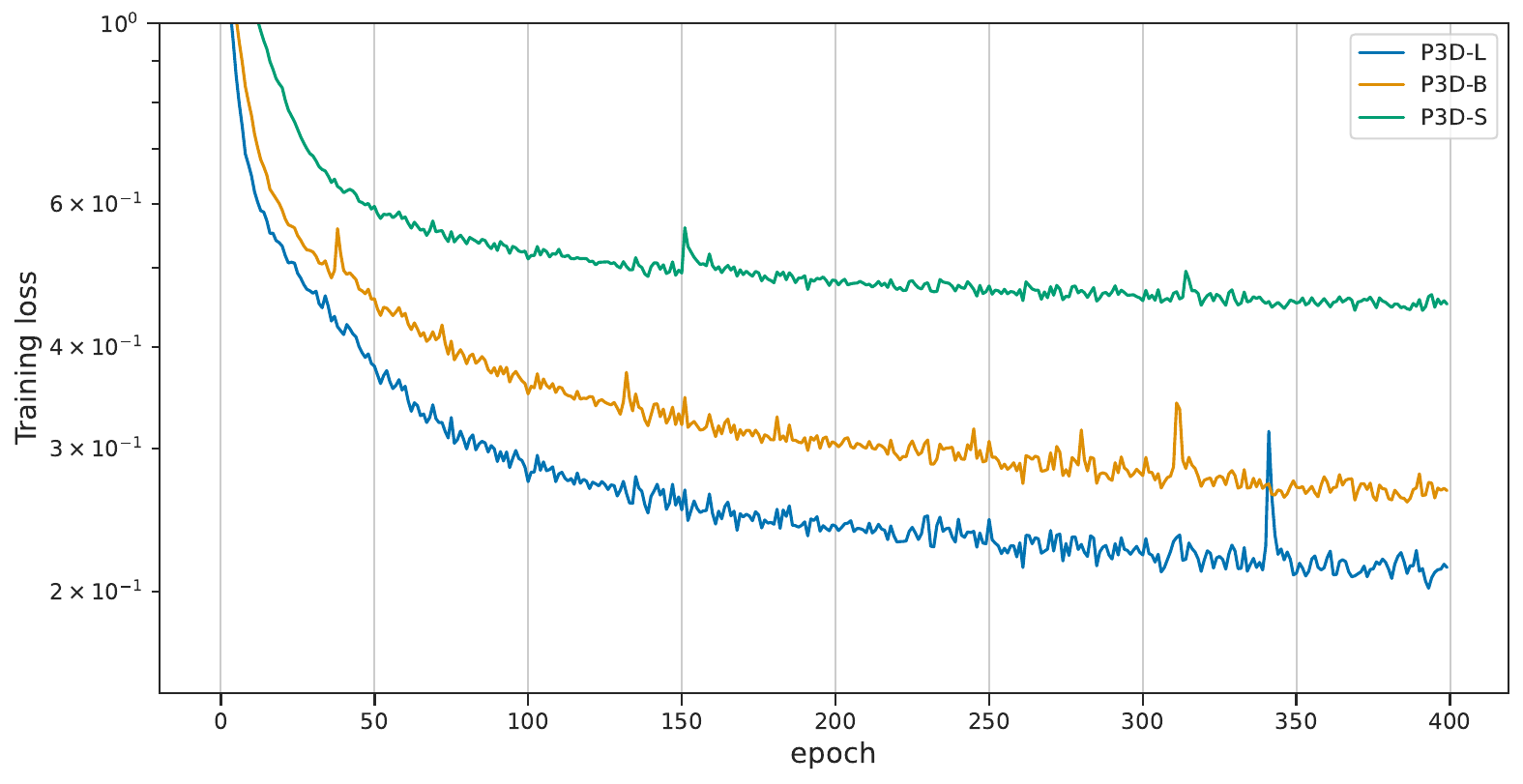}
    \caption{Turbulent channel flow. Training curve for different configs of P3D.}
    \label{fig-app: tcf config scaling}
\end{figure*}

\paragraph{Training S, B and L configs} We train P3D with the configurations $S$, $B$ and $L$ for 400 epochs on the full domain of size $96^2 \times 192$. The training loss is shown in \Cref{fig-app: tcf config scaling}. It is important to choose large architectures in our generative modeling setup based on flow matching. The network size, specifically the embedding dimension is critical for this task with the $L$ config reaching significantly lower loss values compared to the $S$ config trained with the same number of epochs. 
All models were trained on 4 A100 GPUs with 80GB VRAM. Training took 11h 4m, 14h 43m and 27h 55m for the $S$, $B$ and $L$ configs respectively. 

\clearpage

\begin{figure*}[ht]
    \centering
    \includegraphics[width=0.99\textwidth]{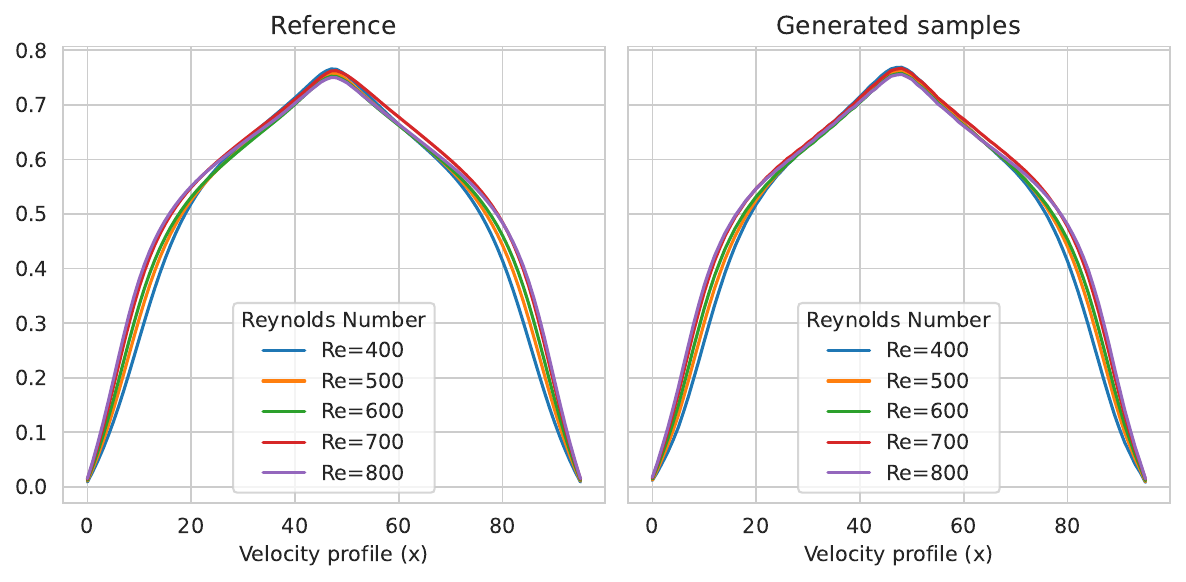}
    \caption{Comparison of the mean channel flow of the reference and of of generated samples from P3D-$L$ trained on the full domain.}
    \label{fig-app: tcf re comparison}
\end{figure*}

\begin{figure*}[ht]
    \centering
    \includegraphics[width=0.99\textwidth]{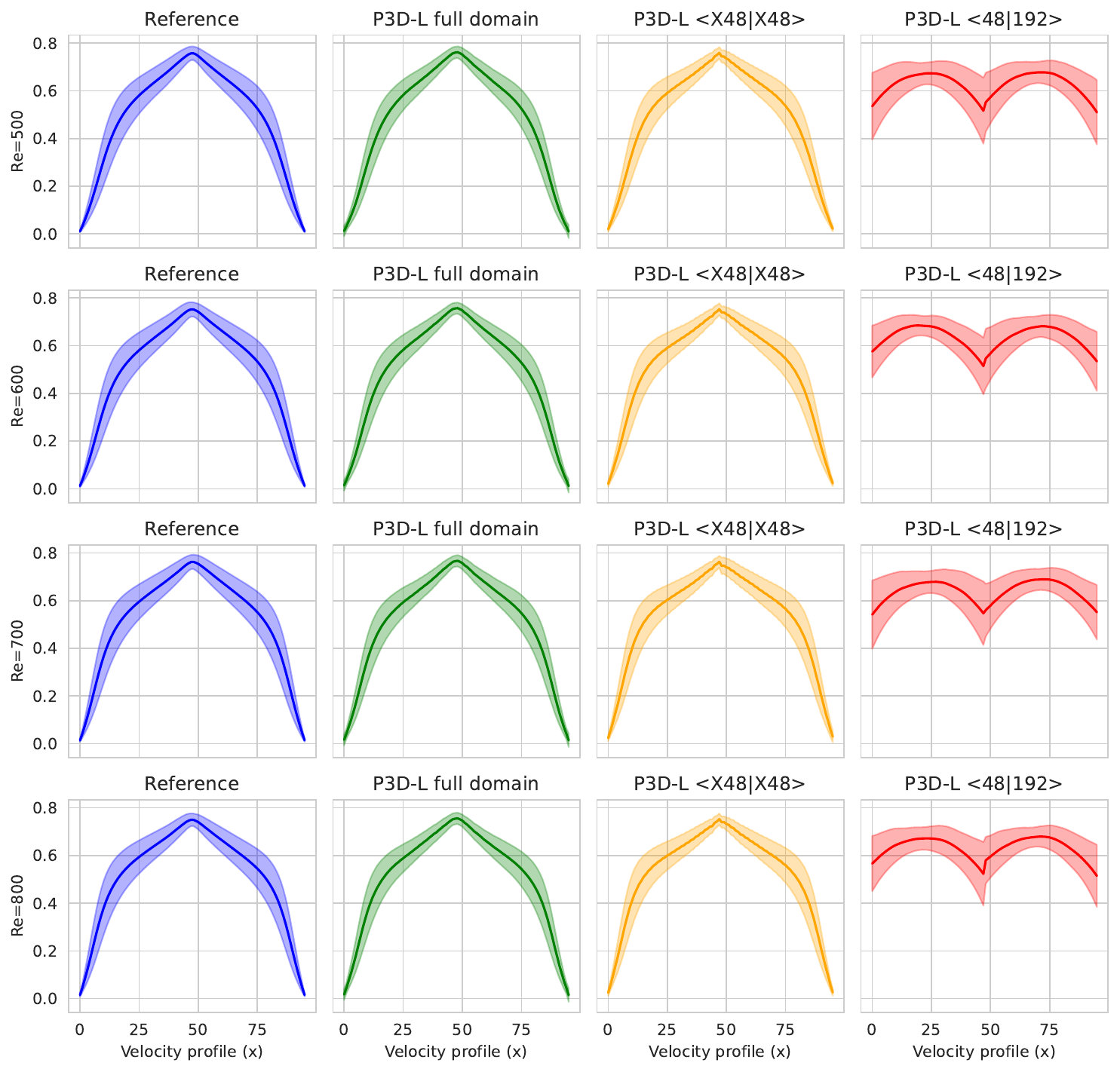}
    \caption{Comparison between the first two moments of the velocity profile of the reference simulations and generated samples for P3D-$L$ with different training and inference variants.}
    \label{fig-app: tcf methods comparison}
\end{figure*}

\begin{figure*}[ht]
    \centering
    \includegraphics[width=0.99\textwidth]{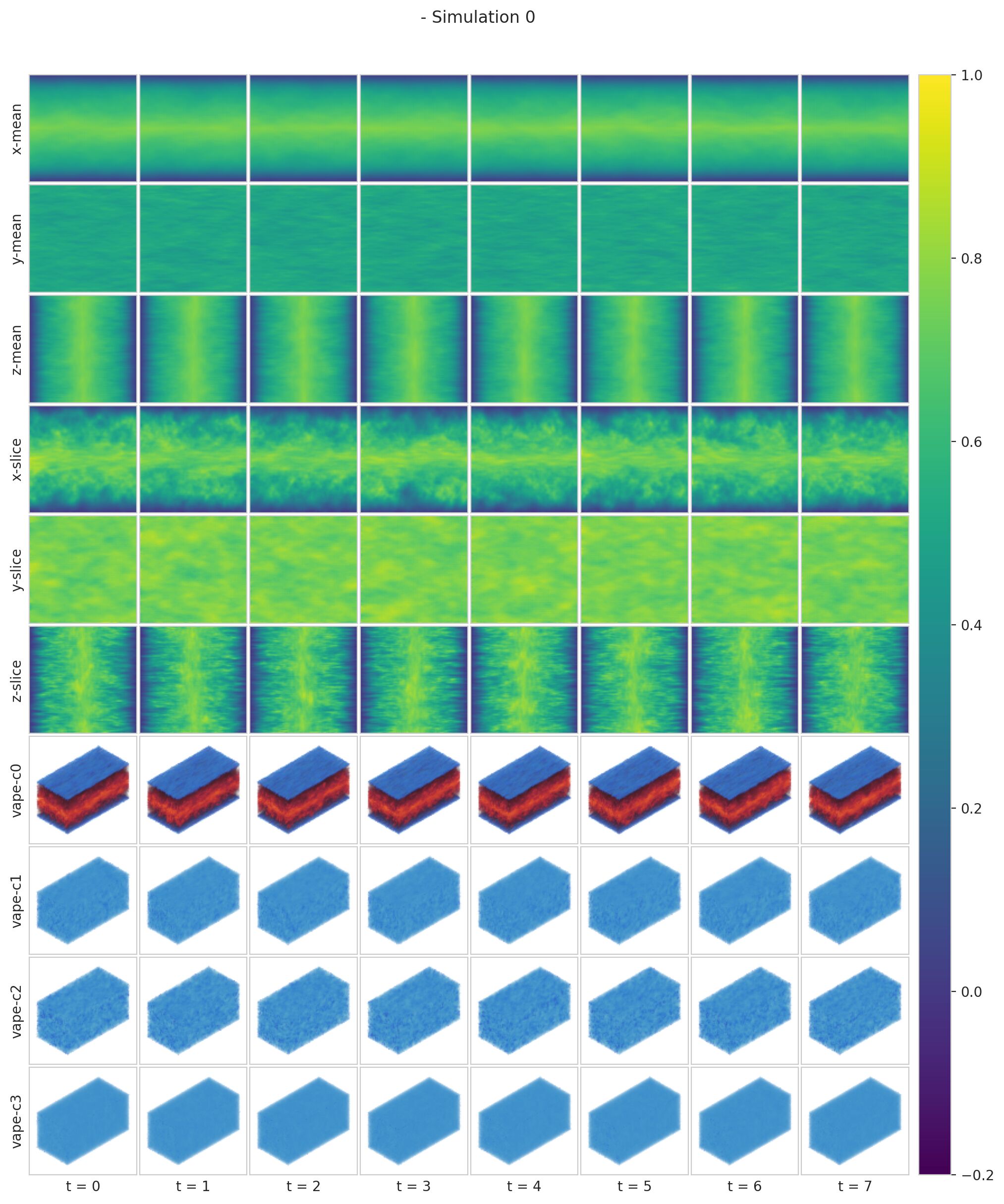}
    \caption{Samples from P3D-$L$ trained on the full domain at $\text{Re}=800$.}
    \label{fig-app: tcf samples full domain}
\end{figure*}

\begin{figure*}[ht]
    \centering
    \includegraphics[width=0.99\textwidth]{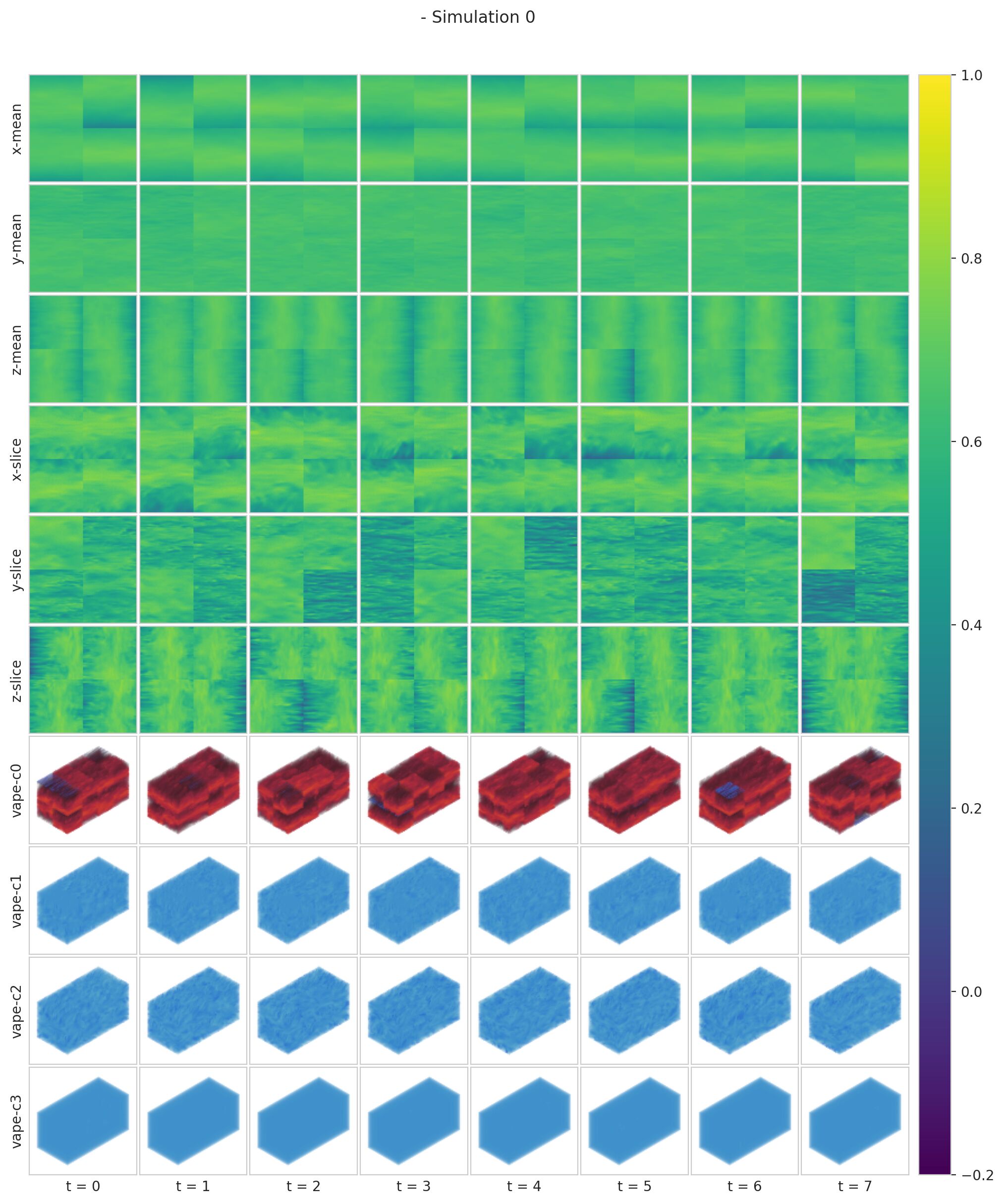}
    \caption{Samples from P3D-$L$ <48|192> pretrained on crops of size $48^3$. Inference on the full domain at $\text{Re}=800$ produces incorrect samples, as information of relative positions between crops is not available.}
    \label{fig-app: tcf samples cropped upscaled}
\end{figure*}

\begin{figure*}[ht]
    \centering
    \includegraphics[width=0.99\textwidth]{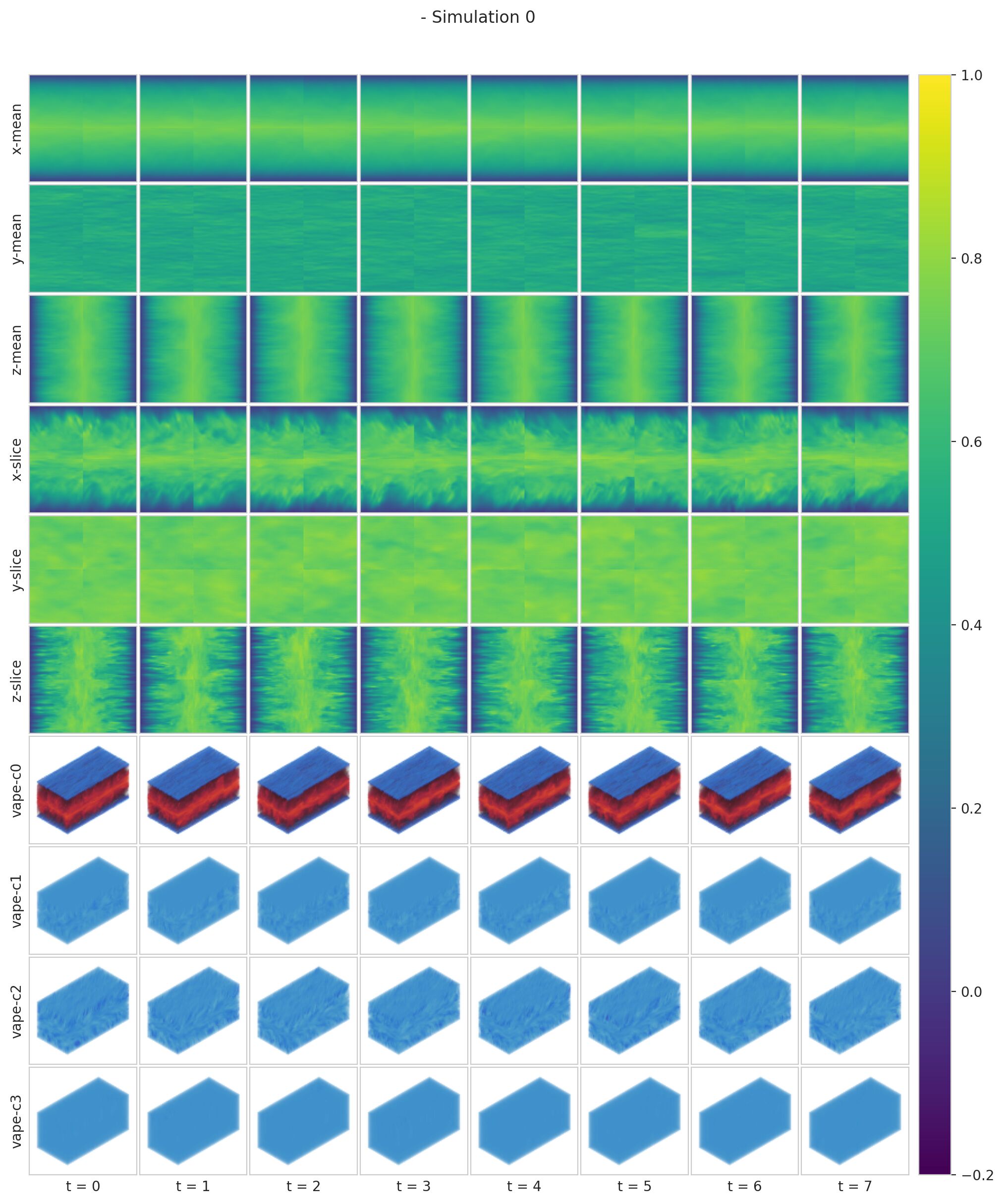}
    \caption{Samples from P3D-$L$ <X48|X48> pretrained on crops of size $48^3$ and finetuned with global context. Inference on the full domain at $\text{Re}=800$ produces samples that exhibit the correct flow statistics.}
    \label{fig-app: tcf samples context}
\end{figure*}

\paragraph{Velocity profile comparison} See \Cref{fig-app: tcf re comparison} for a comparison between the mean flow of the reference simulation for different Reynolds number and the mean flow from the generative model P3D-$L$ trained on the full domain.
We show additional comparisons between the different training and inference strategies in \Cref{fig-app: tcf methods comparison}. Scaling the P3D-$L$ network trained on the small crops of size $48^3$ to the full domain does not work well and results in incorrect velocity profiles. By finetuning with the context network, region crops can coordinate and obtain information about their relative position to each other as well as to the wall. As a results, the flow statistics improve significantly, more closely matching the reference and samples from P3D trained on the full domain.
\\
\\
We show samples from P3D-L trained on the full domain in \Cref{fig-app: tcf samples full domain}, when applying P3D-L pretrained on the full domain without any finetuning in \Cref{fig-app: tcf samples cropped upscaled} and with finetuning via the context network and learned region-dependent conditioning in \Cref{fig-app: tcf samples context} respectively. 

\clearpage

\printbibliography[heading=subbibliography]

\end{refsection}

\end{document}